\documentclass{vldb}
\usepackage{graphicx}
\usepackage{balance}  % for  \balance command ON LAST PAGE  (only there!)

\usepackage{cleveref}

\usepackage{url}
\usepackage{framed}
\usepackage{wrapfig}
\usepackage{enumitem}
\usepackage{tabularx}

\newtheorem{thm}{Theorem}

\newtheorem{assum}{Assumption}
\crefname{thm}{theorem}{theorems}
\crefname{thm}{Theorem}{Theorems}
\crefname{assum}{assumption}{assumptions}
\Crefname{assum}{Assumption}{Assumptions}

\usepackage{soul}	% For \st{} strikethrough text
\usepackage{todonotes}

\usepackage{color}
\newif\ifcomment
\commenttrue % <-- uncomment to add comments

\ifcomment
\newcommand{\qirong}[1]{{\color{blue}{\bf\sf [Qirong: #1]}}} %Qirong's comment
\newcommand{\qirongw}[1]{{\color{cyan}{#1}}} %Qirong's edited text
\else
\newcommand{\qirong}[1]{{\color{blue}{}}}
\newcommand{\qirongw}[1]{{\color{cyan}{}}}
\fi

\title{Distributed Machine Learning via Sufficient Factor Broadcasting}

\usepackage{amsmath}
\usepackage{amsfonts}
\usepackage{graphicx}

\newcommand{\mb}{\mathbf}

\usepackage{dsfont}
\newcommand{\RR}{\mathds{R}}

\usepackage{xspace}
\makeatletter
\DeclareRobustCommand\onedot{\futurelet\@let@token\@onedot}
\def\@onedot{\ifx\@let@token.\else.\null\fi\xspace}
\def\eg{\emph{e.g}\onedot} 
\def\ie{\emph{i.e}\onedot}

\begin{document}

% ****************** TITLE ****************************************

\title{Distributed Machine Learning via Sufficient Factor Broadcasting}

% possible, but not really needed or used for PVLDB:
%\subtitle{[Extended Abstract]
%\titlenote{A full version of this paper is available as\textit{Author's Guide to Preparing ACM SIG Proceedings Using \LaTeX$2_\epsilon$\ and BibTeX} at \texttt{www.acm.org/eaddress.htm}}}

% ****************** AUTHORS **************************************


\numberofauthors{7} %  in this sample file, there are a *total*

\author{
% 1st. author
\alignauthor
Pengtao Xie\\
       \affaddr{Machine Learning Department}\\
       \affaddr{Carnegie Mellon University}\\
       \affaddr{5000 Forbes Ave.}\\
       \affaddr{Pittsburgh, PA 15213}\\
       \email{pengtaox@cs.cmu.edu}
     \alignauthor  
       Jin Kyu Kim\\
       \affaddr{Computer Science Dept.}\\
       \affaddr{Carnegie Mellon University}\\
       \affaddr{5000 Forbes Ave.}\\
       \affaddr{Pittsburgh, PA 15213}\\
       \email{jinkyuk@andrew.cmu.edu}
       \alignauthor
       Yi Zhou\\
       \affaddr{Department of EECS}\\
       \affaddr{Syracuse University }\\
       \affaddr{900 South Crouse Ave}\\
       \affaddr{Syracuse, NY 13244}\\
       \email{yzhou35@syr.edu}
 \and
\alignauthor  Qirong Ho\\
       \affaddr{Institute for Infocomm Research, A*STAR}\\
       \affaddr{1 Fusionopolis Way }\\
       \affaddr{Singapore 138632}\\
       \email{hoqirong@gmail.com}
       \alignauthor   Abhimanu  Kumar\\
       \affaddr{Groupon Inc.}\\
      \affaddr{600 W Chicago Ave }\\
       \affaddr{Chicago, IL 60654-2801}\\
       \email{abhimanyu.kumar@gmail.com}
%\alignauthor
%Yaoliang Yu\\
%       \affaddr{Machine Learning Department}\\
%       \affaddr{Carnegie Mellon University}\\
%       \affaddr{5000 Forbes Ave.}\\
%       \affaddr{Pittsburgh, PA 15213}\\
%       \email{yaoliang@cs.cmu.edu}  
%              \and
       \alignauthor
       Eric Xing\\
       \affaddr{Machine Learning Department}\\
       \affaddr{Carnegie Mellon University}\\
       \affaddr{5000 Forbes Ave.}\\
       \affaddr{Pittsburgh, PA 15213}\\
       \email{epxing@cs.cmu.edu}      
}
%% There's nothing stopping you putting the seventh, eighth, etc.
% author on the opening page (as the 'third row') but we ask,
% for aesthetic reasons that you place these 'additional authors'
% in the \additional authors block, viz.
\additionalauthors{Additional authors: Yaoliang Yu (Machine Learning Department, Carnegie Mellon University {\texttt{yaoliang@cs.cmu.edu}})}
% Just remember to make sure that the TOTAL number of authors
% is the number that will appear on the first page PLUS the
% number that will appear in the \additionalauthors section.

\maketitle

%Matrix-parametrized models, including multiclass logistic regression and sparse coding, are used in machine learning (ML) applications ranging from computer vision to computational biology. When these models are applied to large-scale ML problems starting at millions of samples and tens of thousands of classes, their parameter matrix can grow at an unexpected rate, resulting in high parameter synchronization costs that greatly slow down distributed learning. To address this issue, we propose a Sufficient Factor Broadcasting (SFB) computation model for efficient distributed learning of a large family of matrix-parameterized models, which share the following property: the parameter update computed on each data sample is a rank-1 matrix, i.e., the outer product of two "sufficient factors" (SFs). By broadcasting the SFs among worker machines and reconstructing the update matrices locally at each worker, SFB improves communication efficiency --- communication costs are linear in the parameter matrix's dimensions, rather than quadratic --- without affecting computational correctness. We present a theoretical convergence analysis of SFB, and empirically corroborate its efficiency on four different matrix-parametrized ML models.

\begin{abstract}
Matrix-parametrized models, including multiclass logistic regression and sparse coding, are used in machine learning (ML) applications ranging from computer vision to computational biology. When these models are applied to large-scale ML problems starting at millions of samples and tens of thousands of classes, their parameter matrix can grow at an unexpected rate, resulting in high parameter synchronization costs that greatly slow down distributed learning. To address this issue, we propose a Sufficient Factor Broadcasting (SFB) computation model for efficient distributed learning of a large family of matrix-parameterized models, which share the following property: the parameter update computed on each data sample is a rank-1 matrix, \ie the outer product of two ``sufficient factors" (SFs). By broadcasting the SFs among worker machines and reconstructing the update matrices
locally at each worker, SFB improves communication efficiency --- communication costs are linear in the parameter matrix's dimensions, rather than quadratic --- without affecting computational correctness. We present a theoretical convergence analysis of SFB, and empirically corroborate its efficiency on four different matrix-parametrized ML models.
\end{abstract}

\section{Introduction}

For many popular machine learning (ML) models, such as multiclass logistic regression (MLR), neural networks (NN) \cite{chilimbi2014project}, distance metric learning (DML) \cite{xing2002distance} and sparse coding \cite{olshausen1997sparse}, their parameters can be represented by a matrix $\mb{W}$. For example, in MLR, rows of $\mb{W}$ represent the classification coefficient vectors corresponding to different classes; whereas in SC rows of $\mb{W}$ correspond to the basis vectors used for reconstructing the observed data. A learning algorithm, such as stochastic gradient descent (SGD), would iteratively compute an update $\Delta \mb{W}$ from data, to be aggregated with the current version of $\mb{W}$. We call such models {\it matrix-parameterized models} (MPMs). 
%\ericx{So, this is the tone and the kind of information-serving mentality I'd like to see from the writing of the entire paper. Now please continue to make the case of what happens when data is big and $\mb{W}$ is big.} 
%Matrix-parameterized machine learning (ML) models, including multiclass logistic regression (MLR) and sparse coding \cite{olshausen1997sparse}, are a broad class of methods with important applications in computer vision, web mining and computational biology (to name a few).
%\qirong{No need to explain what different models do in the intro. Giving names and examples is enough.}
%For example, in multiclass logistic regression (MLR) \cite{bishop2006pattern}, a weight coefficient matrix is learned to effectively distinguish different classes. In sparse coding \cite{olshausen1997sparse}, the goal is to learn a dictionary matrix to best reconstruct the input data.

Learning MPMs in large scale ML problems is challenging: ML application scales have risen dramatically, a good example being the ImageNet \cite{deng2009imagenet} compendium with millions of images grouped into tens of thousands of classes. To ensure fast running times when scaling up MPMs to such large problems, it is desirable to turn to distributed computation; however, a unique challenge to MPMs is that the parameter matrix grows rapidly with problem size,
%(unlike non-matrix applications),
causing straightforward parallelization strategies to perform less ideally. Consider a data-parallel algorithm, in which every worker uses a subset of the data to update the parameters ---
a common paradigm is to synchronize the full parameter matrix and update matrices amongst all workers
\cite{dean2008mapreduce,dean2012large,li2015malt,chilimbi2014project,sindhwani2012large,gopal2013distributed}.
However, this synchronization can quickly become a bottleneck: take MLR for example, in which the parameter matrix $\mb{W}$ is of size $J\times D$, where $J$ is the number of classes and $D$ is the feature dimensionality. In one application of MLR to Wikipedia~\cite{partalas2015lshtc}, $J=325$k and $D>10,000$, thus $\mb{W}$ contains several billion entries (tens of GBs of memory).
%when stored with single precision floating numbers.
Because typical computer cluster networks can only transfer a few GBs per second at the most, inter-machine synchronization of $\mb{W}$ can dominate and bottleneck the actual algorithmic computation.
%Apparently, learning parameter matrices of such large size is difficult to be harnessed by a single machine. One would resort to distributed machine learning systems for rescue.
%At the same time, 
In recent years, many distributed frameworks have been developed for large scale machine learning, including Bulk Synchronous Parallel (BSP) systems such as Hadoop
%\footnote{https://hadoop.apache.org/}
\cite{dean2008mapreduce} and Spark \cite{zaharia2012resilient}, graph computation frameworks such as Pregel \cite{malewicz2010pregel}, GraphLab \cite{gonzalez2012powergraph}, and bounded-asynchronous key-value stores such as Yahoo LDA\cite{ahmed2012scalable}, DistBelief\cite{dean2012large}, Petuum-PS \cite{ho2013more}, Project Adam \cite{chilimbi2014project} and \cite{li2014scaling}.
%\pengtao{I feel the following sentence is not a strong message.}
%When using these systems, inter-machine communication is a significant bottleneck to large scale ML;
%%\cite{malewicz2010pregel,ahmed2012scalable,dean2012large,gonzalez2012powergraph, zaharia2012resilient,ho2013more,chilimbi2014project,li2014scaling,li2015malt,xing2015petuum}.
%synchronizing a large volume of parameters (such as in matrix-parameterized applications) incurs delays that slow down algorithm convergence.
When using these systems to learn MPMs, it is common to transmit the full parameter matrices $\mb{W}$ and/or matrix updates $\Delta \mb{W}$ between machines, usually in a server-client style \cite{dean2008mapreduce,dean2012large,sindhwani2012large,gopal2013distributed,chilimbi2014project,li2015malt}. As the matrices become larger due to increasing problem sizes, so do communication costs and synchronization delays --- hence,
%Communication cost could be especially high when the systems are learning matrix parameterized models where the data communicated among machines are usually gigantic parameter matrices and update matrices \cite{dean2008mapreduce,dean2012large,zaharia2012resilient,ho2013more,chilimbi2014project,li2015malt}.
reducing such costs is a key priority when using these frameworks. %\cite{malewicz2010pregel,ahmed2012scalable,dean2012large, gonzalez2012powergraph,tsianos2012communication,zaharia2012resilient, ho2013more,shamir2013communication,yang2013trading,chilimbi2014project, li2014scaling,jaggi2014communication,li2015malt,xing2015petuum}. 

%\ericx{At this point, or after reading the paragraph bellow, one will naturally wonder, why people would do the obviously expensive server-client communication of the full matrix update, rather than the obviously less expensive low rank pre-update. Is it that just because people are stupid and we are smarter? You need to discuss what is the benefit of the former despite of its high expense; and what is the problem of the later (broadcast traffic, sensitive to delay, etc?), and then say it is because we have addressed these previously unknown or unsolved problem, SFB becomes feasible and correct.} \qirong{Made a first pass in cyan below.}

%\st{Motivated by this challenge,} We investigate the structure of matrix-parameterized models, in order to design efficient communication strategies that can be realized in distributed ML frameworks.
%matrices to design communication efficient systems for matrix parameterized models.
%While designing such a system that is suitable for all models parametrized by matrices is probably impossible,
%More formally, we consider ML models whose parameter $\mb{W}$ is a matrix,
%which is updated using stochastic gradient descent (SGD) \cite{dean2012large,ho2013more,chilimbi2014project} or stochastic dual coordinate ascent (SDCA) \cite{hsieh2008dual,shalev2013stochastic,yang2013trading,jaggi2014communication, hsieh2015comm}.
%and whose updates $\bigtriangleup \mb{W}$ are low-rank --- for example, an outer product of two vectors $\mb{u}$ and $\mb{v}$: $\bigtriangleup \mb{W}=\mb{u}\mb{v}^\top$. 
In this paper, we investigate the structure of matrix parameterized models, in order to design efficient communication strategies that can be realized in distributed ML frameworks. We focus on models with a common property: when the parameter matrix $\mb{W}$ of these models is optimized with stochastic gradient descent (SGD) \cite{dean2012large,ho2013more,chilimbi2014project} or stochastic dual coordinate ascent (SDCA) \cite{hsieh2008dual,shalev2013stochastic,yang2013trading,jaggi2014communication,
hsieh2015comm}, the update $\bigtriangleup \mb{W}$ computed over one (or a few) data sample(s) is of low-rank, \eg it
can be written as the outer product of two vectors $\mb{u}$ and $\mb{v}$: $\bigtriangleup \mb{W}=\mb{u}\mb{v}^\top$. 
The vectors $\mb{u}$ and $\mb{v}$ are \textit{sufficient factors} (SF, meaning that they are sufficient
%for carrying out the required computations. 
to reconstruct the update matrix $\bigtriangleup \mb{W}$).
A rich set of models \cite{olshausen1997sparse,lee1999learning,xing2002distance,
yuan2006model,chilimbi2014project} fall into this family: for instance, when solving an MLR problem using SGD, the stochastic gradient is $\bigtriangleup \mb{W}=\mb{u}\mb{v}^\top$, where $\mathbf{u}$ is the prediction probability vector and $\mathbf{v}$ is the feature vector. Similarly, when solving an $\ell_2$ regularized MLR problem using SDCA, the update matrix $\bigtriangleup \mb{W}$ also admits such as a structure, where $\mathbf{u}$ is the update vector of a dual variable and $\mathbf{v}$ is the feature vector. 
%In neural network \cite{rumelhart1988learning,chilimbi2014project}, $\mathbf{u}$ is the backward error vector and $\mathbf{v}$ is the forward activation vector. 
Other models include neural networks \cite{chilimbi2014project}, distance metric learning \cite{xing2002distance}, sparse coding \cite{olshausen1997sparse}, non-negative matrix factorization \cite{lee1999learning}, principal component analysis, and group Lasso \cite{yuan2006model}. 

Leveraging this property, we propose a computation model called Sufficient Factor Broadcasting (SFB), and evaluate its effectiveness in a peer-to-peer implementation (while noting that SFB can also be used in other distributed frameworks).
%which we demonstrate using a decentralized, peer-to-peer implementation (while noting that SFB can also be used in other distributed frameworks).
SFB efficiently learns parameter matrices using the SGD or SDCA algorithms, which are widely-used in distributed ML \cite{hsieh2008dual,dean2012large,ho2013more,shalev2013stochastic,yang2013trading,jaggi2014communication,chilimbi2014project,hsieh2015comm,li2015malt}.
The basic idea is as follows: 
%since the required computations (\eg assembling $\bigtriangleup \mb{W}$ or computing the matrix-vector product $\bigtriangleup \mb{W}\bs a$) can be exactly carried out using the sufficient factors, 
since $\bigtriangleup\mb{W}$ can be exactly constructed from the sufficient factors,
rather than communicating the full (update) matrix between workers, we can instead broadcast only the sufficient factors and have workers reconstruct the updates.
%among workers and use the sufficient factors to reconstruct the update matrices on each receiver's side.
SFB is thus highly communication-efficient; transmission costs are linear in the dimensions of the parameter matrix, and the resulting faster communication greatly reduces waiting time in synchronous systems (e.g. Hadoop and Spark), or improves parameter freshness in (bounded) asynchronous systems (e.g. GraphLab, Petuum-PS and \cite{li2014scaling}). SFs have been used to speed up some (but not all) network communication in deep learning \cite{chilimbi2014project}; our work differs primarily in that we always transmit SFs, never full matrices.

SFB does not impose strong requirements on the distributed system --- it can be used with synchronous~\cite{dean2008mapreduce,malewicz2010pregel,zaharia2012resilient}, asynchronous~\cite{gonzalez2012powergraph,ahmed2012scalable,dean2012large}, and bounded-asynchronous consistency models~\cite{BertsekasTsitsiklis89,ho2013more,terry2013replicated},
%, including Bulk Synchronous Parallel \cite{dean2008mapreduce,malewicz2010pregel,zaharia2012resilient}, Asynchronous Parallel \cite{gonzalez2012powergraph,ahmed2012scalable,dean2012large} and Staleness Synchronous Parallel \cite{BertsekasTsitsiklis89,ho2013more,terry2013replicated}. These consistency models can be used
in order to trade off between system efficiency and algorithmic accuracy.
%To further reduce communication cost, we utilize the Halton sequence broadcast \cite{li2015malt} strategy, where each worker broadcasts messages to a subset of machines rather than all other machines, with the pursuit of eventual consistency over a period of time.
We provide theoretical analysis of SFB under synchronous and bounded-async consistency, and demonstrate that SFB learning of matrix-parametrized models significantly outperforms strategies that communicate the full parameter/update matrix,
%implemented on SFB versus implementations on other platforms for
on a variety of applications including distance metric learning \cite{xing2002distance}, sparse coding \cite{olshausen1997sparse} and unregularized/$\ell_2$-regularized multiclass logistic regression.
%\qirong{I removed the ``preachy" statement about wide applicability; I feel it is not so tasteful. Let our breadth of applications do the talking.}
%Considering that SGD and SDCA are among the most widely used algorithms in distributed ML \cite{dean2012large,ho2013more,yang2013trading,chilimbi2014project, jaggi2014communication,li2015malt} and a rich class of matrix parameterized models can fit into the SFB computation model, our system can potentially find wide applicabilities in many large scale ML tasks.
Using our own C++ implementations of each application, our experiments show that, for parameter matrices with 5-10 billion entries, replacing full-matrix communication with SFB improves convergence times by 3-4 fold. Notably, our SFB implementation of $\ell_2$-MLR is approximately 9 times faster than the Spark v1.3.1 implementation. We expect the performance benefit of SFB (versus full matrix communication) to improve with even larger matrix sizes.

The major contributions of this paper are summarized as follows:
\begin{itemize}
\item We identify the \textit{sufficient factor property} of a large family of matrix-parametrized models when solved with two popular algorithms: stochastic gradient descent and stochastic dual coordinate ascent.
\item In light of the sufficient factor property, we propose a \textit{sufficient factor broadcasting} model of computation, which can greatly reduce the communication complexity without losing computational correctness.
\item We provide an efficient implementation of SFB, with flexible consistency models and easy-to-use programming interface.
\item We analyze the communication and computation costs of SFB and provide a convergence guarantee of SFB based minibatch SGD algorithm.
\item We perform extensive evaluation of SFB on four popular models and corroborate the efficiency and low communication complexity of SFB. 
\end{itemize}

The rest of the paper is organized as follows. In Section 2 and 3, we introduce the sufficient factor property of matrix-parametrized models and propose the sufficient factor broadcasting computation model, respectively. Section 4 presents SFBroadcaster, an implementation of SFB. Section 5 analyzes the costs and convergence behavior of SFB. Section 6 gives experimental results. Section 7 reviews related works and Section 8 concludes the paper.

\section{Sufficient Factor Property of Matrix-Parametrized Models}

%\qirong{TODO: shorten this section (no need so many technical details and examples); introduce DCA}

%In this section, we introduce the SFB computation model.
% and its extensions including Proj-SFB and Partial-SFB.

% \vspace{-0.1in}
% \subsection{Sufficient Factor Update Pattern}
% \vspace{-0.1in}
The core goal of Sufficient Factor Broadcasting (SFB) is to reduce network communication costs for matrix-parametrized models; specifically, those that follow an optimization formulation
%We consider the following optimization problem:
\begin{equation}
\textbf{(P)}\quad
\begin{matrix}
 \underset{\mb{W}}{\textrm{min}}& \frac{1}{N}\sum\limits_{i=1}^{N}f_i(\mb{W}\mb{a}_{i})+h(\mb{W})\\
\end{matrix}
\end{equation} 
where the model is parametrized by a matrix $\mathbf{W}\in \RR^{J\times D}$. The loss function $f_i(\cdot)$ is typically defined over a set of training samples $\{(\mb{a}_i,\mb{b}_i)\}_{i=1}^{N}$, with the dependence on $\mb{b}_i$ being suppressed. 
We allow $f_i(\cdot)$ to be either convex or nonconvex, smooth or nonsmooth (with subgradient everywhere); examples include
$\ell_2$ loss and multiclass logistic loss, amongst others. The regularizer $h(\mb{W})$ is assumed to admit an efficient proximal operator $\textrm{prox}_{h}(\cdot)$ \cite{beck2009fast}. For example, $h(\cdot)$ could be an indicator function of convex constraints, $\ell_1$-, $\ell_2$-, trace-norm, to name a few.
%If there is no regularization, $h(\cdot)$ is simply 0.
The vectors $\mb{a}_i$ and $\mb{b}_i$ can represent observed features, supervised information (e.g., class labels in classification, response values in regression), or even unobserved auxiliary information (such as sparse codes in sparse coding \cite{olshausen1997sparse}) associated with data sample $i$. The key property we exploit below ranges from the matrix-vector multiplication $\mb{W}\mb{a}_{i}$.
This optimization problem \textbf{(P)} can be used to represent a rich set of ML models \cite{olshausen1997sparse,lee1999learning,xing2002distance,
yuan2006model,chilimbi2014project}, such as the following:

%\noindent\textbf{Multiclass logistic regression (MLR)} is used in classification problems with tens of thousands of classes, such as on web data collections like Wikipedia. In MLR, $\mb{W}$ is the weight coefficient matrix, $\mb{a}_{i}$ is the feature vector, $\mb{b}_{i}$ is the 1-of-$K$ coding representation of the class label and $f_i(\cdot)$ is composed of a cross-entropy error function and a softmax mapping of $\mb{W}\mb{a}_{i}$.
%More generally\todoy[]{Can delete for space}, by restricting $\mb{a}_i$ to input features and $\mb{b}_i$ to supervision, $\mathbf{(P)}$ is reduced to the family of multiclass generalized linear models \cite{agarwal2013least}.

\noindent\textbf{Distance metric learning (DML)} \cite{xing2002distance} improves the performance of other ML algorithms, by learning a new distance function that correctly represents similar and dissimilar pairs of data samples; this distance function is a matrix $\mb{W}$ that can have billions of parameters or more, depending on the data sample dimensionality. The vector $\mb{a}_{i}$ is the difference of the feature vectors in the $i$th data pair and $f_i(\cdot)$ can be either a quadratic function or a hinge loss function, depending on the similarity/dissimilarity label $\mb{b}_{i}$ of the data pair. In both cases, $h(\cdot)$ can be an $\ell_1$-, $\ell_2$-, trace-norm regularizer or simply $h(\cdot)=0$ (no regularization).

\noindent\textbf{Sparse coding (SC)} \cite{olshausen1997sparse} learns a dictionary of basis from data, so that the data can be re-represented sparsely (and thus efficiently) in terms of the dictionary.
In SC, $\mb{W}$ is the dictionary matrix, $\mb{a}_{i}$ are the sparse codes, $\mb{b}_{i}$ is the input feature vector and $f_i(\cdot)$ is a quadratic function \cite{olshausen1997sparse}. To prevent the entries in $\mathbf{W}$ from becoming too large, each column $\mathbf{W}_{k}$ must satisfy $\|\mathbf{W}_{k}\|_{2}\leq 1$. In this case, $h(\mb{W})$ is an indicator function which equals 0 if $\mb{W}$ satisfies the constraints and equals $\infty$ otherwise.
%In NMF, $\mb{W}$ is the (pseudo) basis matrix, $\mb{a}_{i}$ is the latent representation, $\mb{b}_{i}$ is the input feature vector and $f$ is a quadratic function. The entries in $\mathbf{W}$ are required to be nonnegative.

\subsection{Optimization via proximal SGD and SDCA}

To solve the optimization problem \textbf{(P)}, it is common to employ either (proximal) stochastic gradient descent (SGD)
%which is one of the premier optimization techniques utilized in large scale distributed ML 
\cite{dean2012large,ho2013more,chilimbi2014project,li2015malt} or stochastic dual coordinate ascent (SDCA) %which has aroused much attention recently in parallel optimization
\cite{hsieh2008dual,shalev2013stochastic,yang2013trading,jaggi2014communication,hsieh2015comm}, both of which are popular and well-established parallel optimization techniques.

\noindent\textbf{Proximal SGD:} In proximal SGD, a stochastic estimate of the gradient, $\bigtriangleup\mb{W}$, is first computed over one data sample (or a mini-batch of samples), in order to update $\mb{W}$ via $\mb{W} \leftarrow \mb{W}-\eta\bigtriangleup\mb{W}$ (where $\eta$ is the learning rate). Following this, the proximal operator $\textrm{prox}_{\eta h}(\cdot)$ is applied to $\mb{W}$.
Notably, the stochastic gradient $\bigtriangleup\mb{W}$ in \textbf{(P)} can be written as the outer product of two vectors $\bigtriangleup\mb{W}=\mb{u}\mb{v}^\top$, where $\mb{u}=\frac{\partial f(\mb{W}\mb{a}_{i},\mb{b}_i)}{\partial (\mb{W}\mb{a}_{i})}$, $\mb{v}=\mb{a}_{i}$, according to the chain rule. Later, we will show that this low rank structure of $\bigtriangleup\mb{W}$ can greatly reduce inter-worker communication.

\noindent\textbf{Stochastic DCA:} SDCA
%\cite{hsieh2008dual,shalev2013stochastic,yang2013trading,jaggi2014communication, hsieh2015comm},
applies to problems \textbf{(P)} where $f_i(\cdot)$ is convex and $h(\cdot)$ is strongly convex 
%w.r.t. $\|\cdot\|_2$ 
\cite{yang2013trading} (\eg when $h(\cdot)$ contains the squared $\ell_2$ norm%
%this includes the $\ell_2$  and elastic net regularizers \cite{zou2005regularization}
);
it solves the dual problem of \textbf{(P)}, via stochastic coordinate ascent on the dual variables. More specifically, introducing the dual matrix $\mb{U} = [\mb{u}_1, \ldots, \mb{u}_N] \in \RR^{J\times N}$ and the data matrix $\mb{A} = [\mb{a}_1, \ldots, \mb{a}_N] \in \RR^{D\times N}$, we can write the dual problem of \textbf{(P)} as
%we obtain the dual problem by vectorizing the parameter matrix $\mb{W}\in \mathrm{R}^{J\times D}$, and turning each vector $\mb{a}_i\in \mathrm{R}^{D}$ into a matrix $\mb{A}_i\in \mathrm{R}^{J\times JD}$ \footnote{We concatenate the row vectors in $\mb{W}$ (from top to bottom) into a long vector $\mb{w}\in \mathrm{R}^{JD}$. We also turn each $\mb{a}_i\in \mathrm{R}^{D}$ into a matrix $\mb{A}_i\in \mathrm{R}^{J\times JD}$ such that in the $j$th row of $\mb{A}_i$, the sub-vector starting from position $(j-1)D+1$ and ending at $jD$ is identical to $\mb{a}_i$, and all other entries are zero. It is easy to check $\mb{W}\mb{a}_{i}=\mb{A}_i\mb{w}$.}. Then the dual problem of \textbf{(P)} can be written as
\begin{equation}
\textbf{(D)}\quad
\begin{matrix}
 \underset{\mb{U}}{\textrm{min}}& \frac{1}{N}\sum\limits_{i=1}^{N}f_i^*(-\mb{u}_{i}) + h^*(\frac{1}{N} \mb{U} \mb{A}^\top)
\end{matrix}
\end{equation} 
where 
%$\mb{v}_{i}\in \mathrm{R}^{J}$ is the dual variable associated with data sample $i$ (and is also the $i$-th row of $\mb{V}$). 
$f_i^*(\cdot)$ and $h^*(\cdot)$ are the Fenchel conjugate functions of $f_i(\cdot)$ and $h(\cdot)$, respectively. The primal-dual matrices $\mb{W}$ and $\mb{U}$ are connected by\footnote{The strong convexity of $h$ is equivalent to the smoothness of the conjugate function $h^*$.} $\mb{W} = \nabla h^*(\mb{Z})$, where the auxiliary matrix $\mb{Z} := \tfrac{1}{N}\mb{U}\mb{A}^\top$.
%the following primal-dual relationship holds:
%$\mb{w}=\nabla h^*(\frac{1}{N}\sum\mb{A}_i^{\mathsf{T}}\mb{v}_{i})$, or equivalently, $\mb{W}=\nabla h^*(\frac{1}{N}\sum_{i=1}^{N}\mb{v}_{i}\mb{a}^{\mathsf{T}}_i)$.
Algorithmically, we need to update the dual matrix $\mb{U}$, the primal matrix $\mb{W}$, and the auxiliary matrix $\mb{Z}$: %$\mb{U}=\frac{1}{N}\sum_{i=1}^{N}\mb{v}_{i}\mb{a}^{\mathsf{T}}_i$: 
every iteration, we pick a random data sample $i$, and compute the stochastic update $\bigtriangleup \mb{u}_{i}$ by minimizing \textbf{(D)} while holding $\{\mb{u}_j\}_{j \ne i}$ fixed. The dual variable is updated via $\mb{u}_{i} \gets \mb{u}_{i} - \bigtriangleup \mb{u}_{i}$, 
%(where $\bigtriangleup \mb{u}_i$ is obtained by minimizing \wrt ), 
the auxiliary variable via $\mb{Z} \gets \mb{Z} - \bigtriangleup\mb{u}_{i}\mb{a}^\top_i$,
and the primal variable via $\mb{W} \gets \nabla h^*(\mb{Z})$. Similar to SGD, the update of the SDCA auxiliary variable $\mb{Z}$ is also the outer product of two vectors: $\bigtriangleup\mb{u}_{i}$ and $\mb{a}_i$, which can be exploited to reduce communication cost.
%The operator $\nabla h^*(\cdot)$ transforming $\mb{U}$ to $\mb{W}$ is analogous to the proximal operator in proximal-SGD.

\noindent\textbf{Sufficient Factor property in SGD and SDCA:}
In both SGD and SDCA, the parameter matrix update can be computed as the outer product of two vectors --- we call these sufficient factors (SFs). This property can be leveraged to improve the communication efficiency of distributed ML systems: instead of communicating parameter/update matrices among machines, we can communicate the SFs and reconstruct the update matrices locally at each machine.
%be able to carry out all required computations locally at each machine (\eg through reconstructing the update matrices). 
Because the SFs are much smaller in size, synchronization costs can be dramatically reduced. See \Cref{sec:theory} below for a detailed analysis.

\noindent\textbf{Low-rank Extensions:}
More generally, the update matrix $\bigtriangleup \mb{W}$ may not be exactly rank-1, but still of very low rank. 
%It is worth noting that ML problems which do not fit into problem \textbf{(P)} may still possess the SF property. 
For example, when each machine uses a mini-batch of size $K$, $\bigtriangleup \mb{W}$ is of rank at most $K$; in Restricted Boltzmann Machines \cite{smolensky1986information}, the update of the weight matrix is computed from four vectors $\mb{u}_1,\mb{v}_1,\mb{u}_2,\mb{v}_2$ as $\mb{u}_1\mb{v}_1^\top-\mb{u}_2\mb{v}_2^\top$, \ie rank-2; for the BFGS algorithm \cite{bertsekas1999nonlinear}, the update of the inverse Hessian is computed from two vectors $\mb{u},\mb{v}$ as $\alpha\mb{u}\mb{u}^\top-\beta(\mb{u}\mb{v}^\top+\mb{v}\mb{u}^\top)$, \ie rank-3.
%In this paper, we focus on models fitting into problem \textbf{(P)}.
Even when the update matrix $\bigtriangleup \mb{W}$ is not genuinely low-rank, to reduce communication cost, it might still make sense to send only a certain low-rank \emph{approximation}. 
We intend to investigate these possibilities in future work.

%We call $\mb{u}$ and $\mb{v}$ as \textit{sufficient factors} (SF) in the sense that they are sufficient to compute the stochastic gradient $\bigtriangleup\mb{W}$.

%Other than the standard SGD algorithm, SFB can also support proximal SGD \cite{beck2009fast}, which is widely utilized to solve problems of the following form:
%\begin{equation}
%\textbf{(P2)}\quad
%\begin{matrix}
% \underset{\mb{W}}{\textrm{min}}& \sum\limits_{i=1}^{N}f(\mb{W}\mb{a}_{i},\mb{b}_i)+h(\mb{W})\\
%\end{matrix}
%\vspace{-0in}
%\end{equation} 
%where $f(\cdot)$ is differentiable. $h(\cdot)$ is not necessarily differentiable, but has a proximal operator $\textrm{prox}_{h}(\cdot)$. Proximal SGD first updates the parameter matrix by stochastic gradient descent w.r.t $f(\cdot)$, then applies the proximal operator $\textrm{prox}_{h}(\cdot)$ to the updated parameter matrix. Similar to \textbf{(P1)}, the stochastic gradient $\bs\bigtriangleup\mb{W}$ in \textbf{(P2)} also possesses a rank-1 structure.
% Other examples are matrix parametrized optimization problems with constraints, including sparse coding (SC) \cite{olshausen1997sparse}, nonnegative matrix factorization (NMF) \cite{lee1999learning}, principal component analysis (PCA) and others. 

\section{Sufficient Factor Broadcasting}

%\qirong{TODO: further emphasize stochastic issues: minibatch size and delay/batching before sending updates.}

\begin{figure}
\centering
\includegraphics[width=0.8\columnwidth]{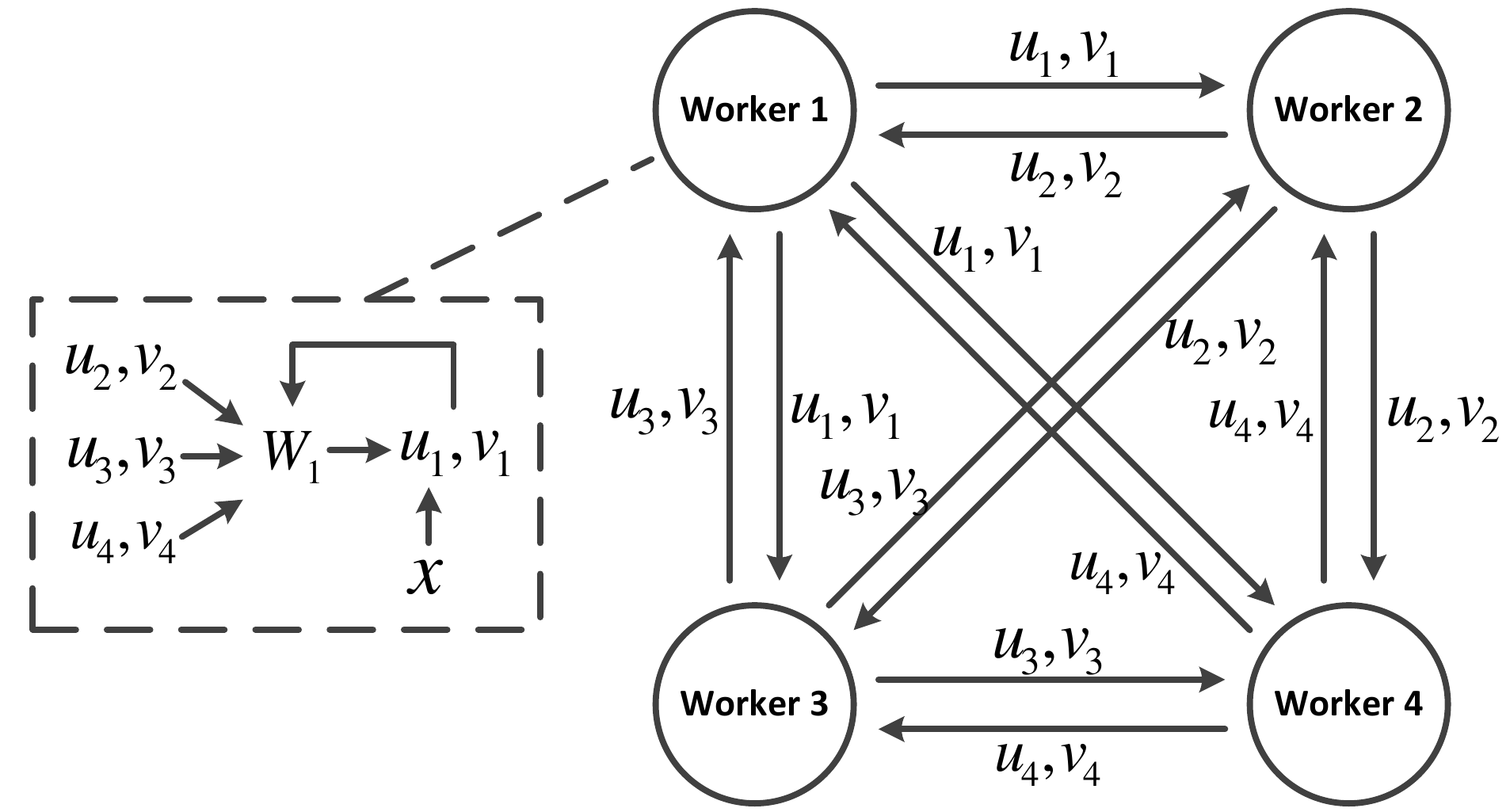}
\caption{Sufficient Factor Broadcasting (SFB).}
\label{fig:svs}
\vspace{-0.2in}
\end{figure}

Leveraging the SF property of the update matrix in problems \textbf{(P)} and \textbf{(D)}, we propose a \textit{Sufficient Factor Broadcasting} (SFB) model of computation, that supports efficient (low-communication) distributed learning of the parameter matrix $\mb{W}$.
%\qirong{I removed the ``decentralized" statement, as SFB can also be used with centralized architectures.}
%In SFB, the parameter matrix is learned with distributed SGD or SDCA in a decentralized fashion.
We assume a setting with $P$ workers, each of which holds a data shard and a copy of the parameter matrix\footnote{For simplicity, we assume each worker has enough memory to hold a full copy of the parameter matrix $\mb{W}$. If $\mb{W}$ is too large, one can either partition it across multiple machines \cite{dean2012large,li2014scaling,lee2014strads}, or use local disk storage (\ie out of core operation). We plan to investigate these strategies as future work.} $\mb{W}$. Stochastic updates to $\mb{W}$ are generated via proximal SGD or SDCA, and communicated between machines to ensure parameter consistency.
%SFB broadcasts the sufficient factors in order to synchronize the local parameter matrix copies at each worker.
In proximal SGD, on every iteration, each worker $p$ computes SFs $(\mathbf{u}_{p},\mathbf{v}_{p})$, based on one data sample $\mb{x}_i = (\mb{a}_i,\mb{b}_i)$ in the worker's data shard. The worker then broadcasts $(\mathbf{u}_{p},\mathbf{v}_{p})$
%which represent the stochastic update to $\mb{W}$ due to $(\mb{a}_i,\mb{b}_i)$,
to all other workers; once all $P$ workers have performed their broadcast (and have thus received all SFs),
%worker $p$ will have received all other workers' SF-pairs $\mathbf{u}_{q},\mathbf{v}_{q}$ (where $q\in \{1,\cdots,P\}/\{p\}$). Finally,
they re-construct the $P$ update matrices (one per data sample) from the $P$ SFs, and 
apply them to update their local copy of $\mb{W}$. Finally, each worker applies
%In the stochastic gradient descent step, workers synchronize parameters by broadcasting sufficient factors. 
%After the parameter copy is updated with the gradient (computed from the SFs),
the proximal operator $\textrm{prox}_{h}(\cdot)$.
When using SDCA, the above procedure is instead used to broadcast SFs for the auxiliary matrix $\mb{Z}$, which is then used to obtain the primal matrix $\mb{W} = \nabla h^*(\mb{Z})$.
%is applied to the parameter copy locally on each worker machine. 
%\todoy{Shall one mention how the system is considered converged, e.g., when to halt?} 
%\pengtao{I will put how to halt in the implementation details in supplements.}
\Cref{fig:svs} illustrates SFB operation: 4 workers compute their respective SFs $(\mb{u}_1,\mb{v}_1)$, $\dots$, $(\mb{u}_4,\mb{v}_4)$, which are then broadcast to the other 3 workers.
%with 4 workers, Worker 1 computes sufficient factors $(\mathbf{u}_{1},\mathbf{v}_{1})$ based on the local parameter copy $\mathbf{W}_{1}$ and a data point $\mb{x}$ randomly sampled from the data shard. Then $(\mathbf{u}_{1},\mathbf{v}_{1})$ are used to update $\mathbf{W}_{1}\gets \mathbf{W}_{1}-\mathbf{u}_{1}\mathbf{v}^{\mathsf{T}}_{1}$. Worker 1 broadcasts $(\mathbf{u}_{1},\mathbf{v}_{1})$ to worker 2, 3, 4 and receives the sufficient factors $(\mathbf{u}_{2},\mathbf{v}_{2})$, $(\mathbf{u}_{3},\mathbf{v}_{3})$, $(\mathbf{u}_{4},\mathbf{v}_{4})$ sent from worker 2, 3, 4.
Each worker $p$ uses all 4 SFs $(\mb{u}_1,\mb{v}_1), \dots, (\mb{u}_4,\mb{v}_4)$
%The received SFs $(\mathbf{u}_{2},\mathbf{v}_{2})$, $(\mathbf{u}_{3},\mathbf{v}_{3})$, $(\mathbf{u}_{4},\mathbf{v}_{4})$ are used
to exactly reconstruct the update matrices $\bigtriangleup \mb{W}_{p} = \mb{u}_{p} \mb{v}_{p}^\top$, and update their local copy of the parameter matrix: $\mb{W}_{p} \leftarrow \mb{W}_{p} - \sum_{q=1}^4 \mb{u}_{q} \mb{v}_{q}^\top$.  While the above description reflects synchronous execution, asynchronous and bounded-asynchronous extensions are also possible (Section \ref{sec:sfbcaster}).

\noindent\textbf{SFB vs client-server architectures:}
The SFB peer-to-peer topology can be contrasted with a ``full-matrix" client-server architecture for parameter synchronization, e.g. as used by Project Adam \cite{chilimbi2014project} to learn neural networks: there, a centralized server maintains the global parameter matrix, and each client keeps a local copy. Clients compute sufficient factors and send them to the server, which uses the SFs to update the global parameter matrix; the server then sends the full, updated parameter matrix back to clients. Although client-to-server costs are reduced (by sending SFs), server-to-client costs are still expensive because full parameter matrices need to be sent. In contrast, the peer-to-peer SFB topology never sends full matrices; only SFs are sent over the network.
%\pengtao{To answer Yaoliang's question regarding the reconstruction cost.}
We also note that under SFB, the update matrices are reconstructed at each of the $P$ machines, rather than once at a central server (for full-matrix architectures). %\cite{dean2008mapreduce,dean2012large,li2015malt,chilimbi2014project,sindhwani2012large,gopal2013distributed}.
Our experiments show that the time taken for update reconstruction is empirically negligible compared to communication and SF computation.

\noindent\textbf{Mini-batch proximal SGD/SDCA:}
SFB can also be used in mini-batch proximal SGD/SDCA; every iteration, each worker samples a mini-batch of $K$ data points, and computes $K$ pairs of sufficient factors $\{(\mb{u}_{i}, \mb{v}_{i})\}_{i=1}^{K}$. These $K$ pairs are broadcast to all other workers, which reconstruct the originating worker's update matrix as $\bigtriangleup\mb{W}=\frac{1}{K}\sum_{i=1}^{K}\mb{u}_{i}\mb{v}_{i}^{\mathsf{T}}$.
%While the $K$ SF pairs cannot be aggregated, and therefore must be sent individually (unlike full matrix updates which can be summed up beforehand), the benefit of not having to send a full $J\times D$ update matrix more than makes up, as our experiments will show.

%\noindent\textbf{Delayed or batched updates:}
%\qirong{Since SFB requires each SF to be individually transmitted, batching/delaying will not have a significant performance impact on SFB. However, it does improve full matrix transmission (because of aggregation). We need to show the tradeoff, and hopefully show that SFB is still preferable under reasonable minibatch sizes.}
%\qirong{Note that minibatch size $K$ also impacts convergence rate --- that's another potential axis to study for a full systems paper (but perhaps not needed in this paper).}

%\vspace{-0.1in}
%\subsection{Proximal Sufficient Factor Broadcasting}
%\vspace{-0.1in}

\section{Sufficient Factor Broadcaster: An Implementation}

\label{sec:sfbcaster}
In this section, we present Sufficient Factor Broadcaster (SFBcaster) --- an implementation of SFB --- including consistency models, programing interface and implementation details.
%The SFB computational model complements synchronous, asynchronous, and bounded-asynchronous models of communication, such as Bulk Synchronous Parallel (BSP) \cite{dean2008mapreduce,malewicz2010pregel,zaharia2012resilient}, Asynchronous Parallel (ASP) \cite{ahmed2012scalable,dean2012large,gonzalez2012powergraph} and Stale Synchronous Parallel (SSP) \cite{BertsekasTsitsiklis89,ho2013more,terry2013replicated}. In order to thoroughly investigate how SFB benefits ML program execution under these communication models, we have written our own implementation with a simple programming interface --- the Sufficient Factor Broadcaster (SFBcaster) --- that supports the aforementioned BSP, ASP and SSP communication. 
We stress that SFB does not require a special system; it can be implemented on top of existing distributed frameworks, using any suitable communication topology --- such as star\footnote{For example, each worker sends the SFs to a hub machine, which re-broadcasts them to all other workers.}, ring, tree, fully-connected and Halton-sequence \cite{li2015malt}. 
%As future work, we intend to investigate the effect of these topologies on performance.

% To investigate the SFB computational model, we have implemented it as a software package, the Sufficient Factor Broadcaster, that incorporates several ML-centric consistency models and a simple programming interface. The SFB idea can certainly be implemented on other distributed ML frameworks, though such implementations are out of the scope of this paper.
%In this section, we present an implementation of the SFB computation model, including consistency models and programming interface. 
%%We acknowledge that the logical computation model of SFB can be physically implemented on top of a broad range of existing systems with proper alteration, either on the ones with a centralized architecture \cite{ahmed2012scalable,dean2012large,ho2013more,li2014scaling}\footnote{Using a central coordinator to collect and dispatch all the sufficient factors.} or those with a peer-to-peer architecture \cite{bhaduri2008distributed,datta2009approximate,das2010local,ormandi2013gossip,li2015malt}.
%The implementation details and the discussion of fault tolerance are deferred to supplementary materials.

\subsection{Flexible Consistency Models}
\label{sec:consist}

Our SFBcaster implementation supports three consistency models: Bulk Synchronous Parallel (BSP-SFB), Asynchronous Parallel (ASP-SFB), and Stale Synchronous Parallel (SSP-SFB), and we provide theoretical convergence guarantees for BSP-SFB and SSP-SFB in the next section.

\noindent\textbf{BSP-SFB:}
Under BSP \cite{dean2008mapreduce,malewicz2010pregel,zaharia2012resilient},
%in each iteration, each worker computes one (or a minibatch of) SF pair(s), sends them to all others workers, receives SFs from all other workers and uses SFs to update local parameter copy.
an end-of-iteration global barrier ensures all workers have completed their work, and synchronized their parameter copies, before proceeding to the next iteration. BSP is a strong consistency model, that guarantees the same computational outcome (and thus algorithm convergence) each time.

\noindent\textbf{ASP-SFB:}
BSP can be sensitive to stragglers (slow workers) \cite{ho2013more,terry2013replicated}, 
%which are caused by transient effects such as network congestion or other users' jobs \cite{chilimbi2014project,li2014scaling};
%At a certain time interval, some workers (called stragglers) may run slower than others.
%when stragglers are present, BSP requires faster workers to wait for them
limiting the distributed system to the speed of the slowest worker.
%The faster workers have to spend a lot of idle time waiting for the stragglers and the overall pace of the system is slowed down by the stragglers.

The Asynchronous Parallel (ASP) \cite{gonzalez2012powergraph,ahmed2012scalable,dean2012large} communication model addresses this issue, by allowing workers to proceed without waiting for others.
%: under ASP-SFB, workers immediately broadcast SF pairs as they are computed, and resume computation using their current local parameter copy (without waiting for other workers). When a worker receives other workers' SF pairs, it updates the local parameter copy in the background. 
ASP is efficient in terms of iteration throughput, but carries the risk that worker parameter copies can end up greatly out of synchronization, which can lead to algorithm divergence \cite{ho2013more}. 
%rendering the parameter copy on the slow worker out of synchronization. Even worse, the out-of-date parameter copy would generate staled sufficient factors (gradients) that hurt the parameter copies of other workers. 

\noindent\textbf{SSP-SFB:}
Stale Synchronous Parallel (SSP) \cite{BertsekasTsitsiklis89,ho2013more,terry2013replicated} is a bounded-asynchronous consistency model that serves as a middle ground between BSP and ASP; it allows workers to advance at different rates, provided that the difference in iteration number between the slowest and fastest workers is no more than a user-provided staleness $s$. SSP alleviates the straggler issue while guaranteeing algorithm convergence \cite{ho2013more,terry2013replicated}. Under SSP-SFB, each worker $p$ tracks the number of SF pairs computed by itself, $t_p$, versus the number $\tau_p^{q}(t_p)$ of SF pairs received from each worker $q$. If there exists a worker $q$ such that $t_p-\tau_p^{q}(t_p)> s$ (\ie some worker $q$ is likely more than $s$ iterations behind worker $p$), then worker $p$ pauses until $q$ is no longer $s$ iterations or more behind.
%$\forall q\in\{1,\cdots,P\}/\{p\}$, $C_p-C_{q}\leq s$.
When $s=0$, SSP-SFB reduces to BSP-SFB \cite{dean2008mapreduce,zaharia2012resilient}, and when $s=\infty$, SSP-SFB becomes ASP-SFB.
%\todoy{Note the subtlety here: $t_p-\tau_p^{q}(t_p)$ is only an upper bound of the iteration difference between machine $p$ and $q$. (machine $q$ could have already catched up but its updates have not arrived at machine $p$)}
%Our experiments using SFBcaster will show that SFB improves the performance of matrix-parameterized ML programs running under all three communication models (compared to using full matrix updates).

\subsection{Programming Interface}

\begin{figure}
%\vspace{-0.3in}
\begin{framed}
\textcolor{blue}{sfb\_app} mlr ( \textcolor{blue}{int} J, \textcolor{blue}{int} D, \textcolor{blue}{int} staleness )\\*
\textcolor{orange}{//SF computation function}\\*
function compute\_sv ( \textcolor{blue}{sfb\_app} mlr ):\\*
 \hspace*{0.2cm} while ( ! converged ):\\
\hspace*{0.4cm} $X$ = sample\_minibatch ()\\
\hspace*{0.4cm} foreach $\mb{x}_i$ in $X$:\\
 \hspace*{0.6cm} \textcolor{orange}{//sufficient factor $\mb{u}_i$}\\
\hspace*{0.6cm} pred = predict ( mlr.para\_mat, $\mb{x}_i$ ) \\            
\hspace*{0.6cm} mlr.sv\_list[i].write\_u ( pred ) \\
  \hspace*{0.6cm} \textcolor{orange}{//sufficient factor $\mb{v}_i$}\\
  \hspace*{0.6cm} mlr.sv\_list[i].write\_v ( $\mb{x}_i$ ) \\
  \hspace*{0.4cm} commit() 
\end{framed}
\caption{Multiclass LR pseudocode.}
\vspace{-0.2in}
\label{fig:mlr}

\end{figure}
\begin{figure}

\begin{framed}
\textcolor{blue}{sfb\_app} sc ( \textcolor{blue}{int} D, \textcolor{blue}{int} J , \textcolor{blue}{int} staleness)\\*
\textcolor{orange}{//SF computation function}\\*
 function compute\_sf ( \textcolor{blue}{sfb\_app} sc ):\\*
 \hspace*{0.2cm} while ( ! converge ): \\
\hspace*{0.4cm} $X$=sample\_minibatch ()\\
\hspace*{0.4cm} foreach $\mb{x}_i$ in $X$:\\
 \hspace*{0.6cm} \textcolor{orange}{//compute sparse code}\\
\hspace*{0.6cm} a = compute\_sparse\_code ( sc.para\_mat, $\mb{x}_i$ )   \\   
\hspace*{0.6cm} \textcolor{orange}{//sufficient factor $\mb{u}_i$}\\         
\hspace*{0.6cm} sc.sf\_list[i].write\_u (  sc.para\_mat * a-$\mb{x}_i$ )\\
  \hspace*{0.6cm} \textcolor{orange}{//sufficient factor $\mb{v}_i$}\\
  \hspace*{0.6cm} sc.sf\_list[i].write\_v ( a )\\
  \hspace*{0.4cm} commit()\\
\textcolor{orange}{//Proximal operator function}\\*
 function prox ( \textcolor{blue}{sfb\_app} sc ):\\*
\hspace*{0.2cm} foreach column $\mb{b}_i$ in sc.para\_mat:\\
  \hspace*{0.4cm} if $\|\mb{b}_i\|_2>1$:\\
  \hspace*{0.6cm} $\mb{b}_i=\frac{\mb{b}_i}{\|\mb{b}_i\|_2}$
\end{framed}
\label{fig:sc}
\caption{Sample code of sparse coding in SFB}
\end{figure}

The SFBcaster programming interface is simple; users need to provide a SF computation function to specify how to compute the sufficient factors. %users read the parameter matrix via:
%\begin{itemize}[leftmargin=*,topsep=0pt,noitemsep]
%\vspace{-0.05in}
%\itemsep-0.3em
%\item $read\_para\_mat$ (int $i$, int $j$): read the parameter matrix element at row $i$ and column $j$.
%\end{itemize}
%\vspace{-0.05in}
To send out SF pairs $(\mb{u},\mb{v})$, the user adds them to a buffer object $sv\_list$, via:
%\begin{itemize}[leftmargin=*,topsep=0pt,noitemsep]
%\vspace{-0.05in}
%\itemsep-0.3em
%\item 
$write\_u(vec\_u)$, $write\_v(vec\_v)$, which set $i$-th SF $\mb{u}$ or $\mb{v}$ to $vec\_u$ or $vec\_v$.
%\item $sv\_list[i].write\_v(vec)$: set $i$-th SF $\mb{v}$ to $vec$.
%\vspace{-0.05in}
%\end{itemize}
All SF pairs 
%(indexed by $i$) inside $sv\_list$ 
are sent out at the end of an iteration, which is signaled by
%To signal the end of iteration and broadcast the SF pairs, the user invokes:
%\begin{itemize}[leftmargin=*,topsep=0pt,noitemsep]
%$\vspace{-0.05in}
%\item 
$commit()$.
%, which notifies SFBcaster that the current iteration/minibatch is finished.
%\end{itemize}
%\vspace{-0.05in}
Finally, in order to choose between BSP, ASP and SSP consistency, users simply set $staleness$ to an appropriate value (0 for BSP, $\infty$ for ASP, all other values for SSP).
SFBcaster automatically updates each worker's local parameter matrix using all SF pairs --- including both locally computed SF pairs added to $sv\_list$, as well as SF pairs received from other workers.
% The programmer declares an $svs\_app$ object, and and then write a function to iteratively perform the following three things: (1) randomly sample one (or a minibatch of) data point(s); (2) compute sufficient vectors for each data sample; (3) commit the computation.

Figure \ref{fig:mlr} shows SFBcaster pseudocode for multiclass logistic regression. For proximal SGD/SDCA algorithms, SFBcaster requires users to write an additional function, $prox(mat)$, which applies the proximal operator $\textrm{prox}_{h}(\cdot)$ (or the SDCA dual operator $h^*(\cdot)$) to the parameter matrix $mat$. Figure 3 shows the sample code of implementing sparse coding in SFB.  $D$ is the feature dimensionality of data and $J$ is the dictionary size. Users write a SF computation function to specify how to compute the sufficient factors: for each data sample $\mb{x}_{i}$, we first compute its sparse code $\mb{a}$ based on the dictionary $\mb{B}$ stored in the parameter matrix $sc.para\_mat$. Given $\mb{a}$, the sufficient factor $\mb{u}$ can be computed as $\mb{B}\mb{a}-\mb{x}_{i}$ and the sufficient factor $\mb{v}$ is simply $\mb{a}$. In addition, users provide a proximal operator function to specify how to project $\mb{B}$ to the $\ell_2$ ball constraint set.

\subsection{Halton Sequence Broadcast}

\begin{figure}
\begin{center}
\includegraphics[width=0.4\columnwidth]{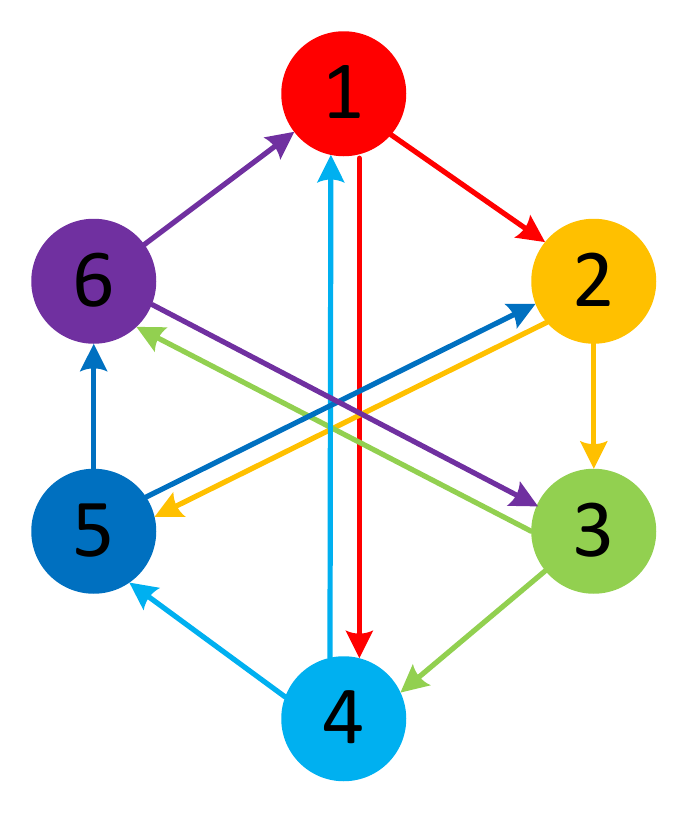}
\caption{An Example of the Halton Sequence Broadcasting.}
\label{fig:halton}
\end{center}
\vspace{-0.2in}
\end{figure}

In previous sections, we assume the sufficient factors generated by each worker are broadcasted to all other workers. Such a full broadcast pattern incurs a communication cost $O(P^2)$, which grows quadratically with the number of workers $P$. In data center scale clusters, $P$ can reach several thousand \cite{dean2012large,li2014scaling}, inwhere a full broadcast scheme would be too costly. To address this issue, we borrow the Halton-sequence idea proposed in \cite{li2015malt}, where each machine connects with and broadcasts messages to a subset of $Q$ machines rather than all other machines. $Q$ is in the scale of $\log P$, thereby, the communication cost can be greatly reduced. The basic idea of Halton sequence broadcast (HSB) \cite{li2015malt} works as follows: given a constructed Halton sequence\footnote{\url{http://en.wikipedia.org/wiki/Halton_sequence}} $\{P/2, P/4, 3P/4, P/8, 3P/8, \cdots\}$, each worker $p$ sends the sufficient vectors to $Q$ machines with IDs $\{$$(p+\lfloor P/2 \rfloor)\%P$, $(p+\lfloor P/4 \rfloor)\%P$, $(p+\lfloor 3P/4 \rfloor)\%P$, $(p+\lfloor P/8 \rfloor)\%P$, $(p+\lfloor 3P/8 \rfloor)\%P$, $\cdots\}$. Figure \ref{fig:halton} gives an illustration. In this example, we have 6 machines and set $Q=2\approx\log_2(6)$. According to the connection pattern rule, machine $p$ should broadcast messages to machine $(p+3)\%6$ and $(p+1)\%6$. For example, machine 1 broadcasts messages to machine 2 and 4; machine 5 broadcasts messages to machine 2 and 6. HSB loses per-iteration consistency of different parameter copies since each worker only broadcasts the sufficient vectors to part of peers. However, it pursues eventual consistency in the sense that over a period of time all the workers can see the effects of updates from every other worker directly or indirectly via an intermediate worker.

\subsection{Implementation Details}

\begin{figure}
\begin{center}
\includegraphics[width=0.8\columnwidth]{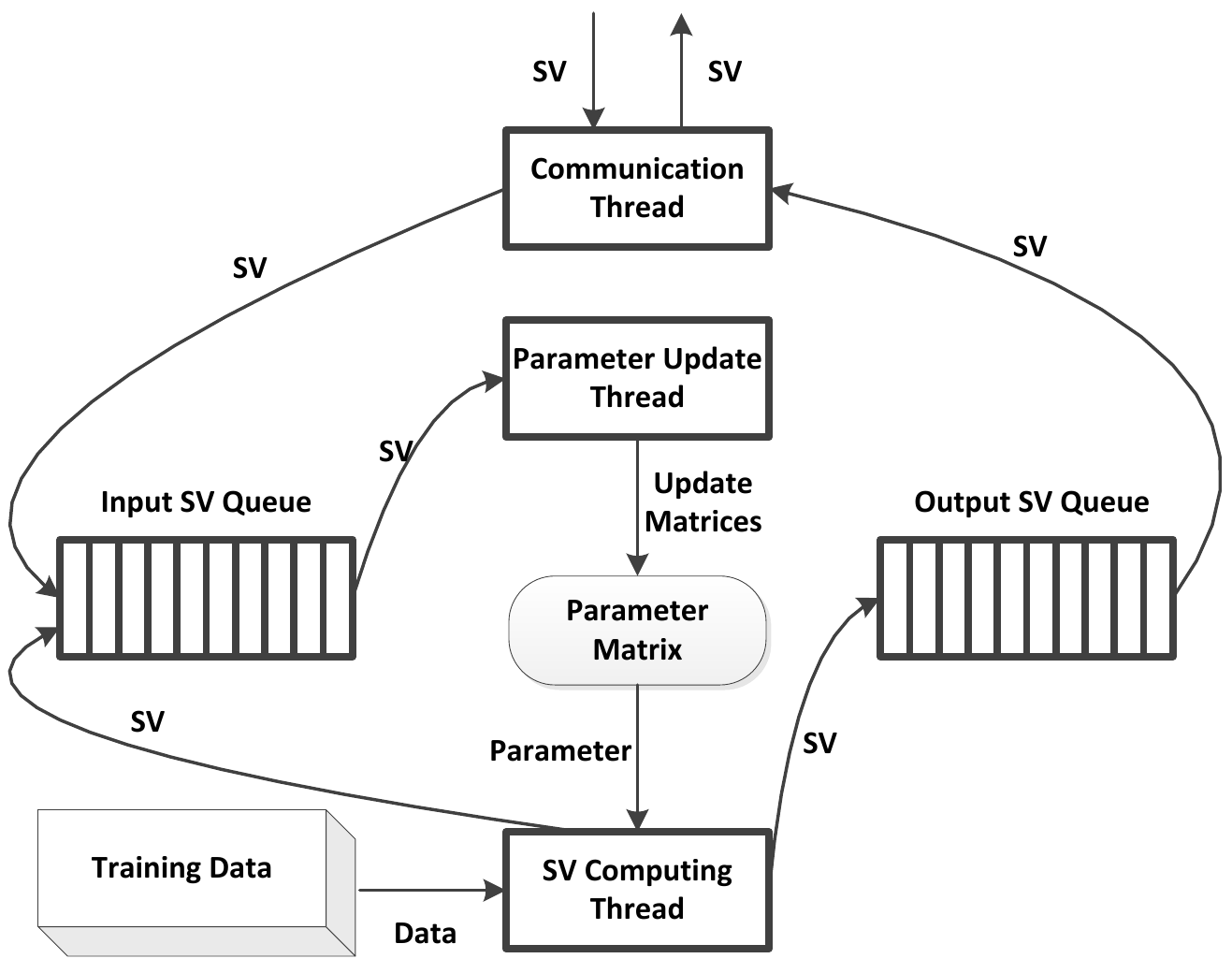}
\caption{Implementation details on each worker in SFBcaster.}
\label{fig:imple}
\end{center}
\vspace{-0.2in}
\end{figure}
Figure \ref{fig:imple} shows the implementation details on each worker in SFBcaster.
Each worker maintains three threads: SF computing thread, parameter update thread and communication thread. Each worker holds a local copy of the parameter matrix and a partition of the training data. It also maintains an input SF queue which stores the sufficient factors computed locally and received remotely and an output SF queue which stores SFs to be sent to other workers. In each iteration, the SF computing thread checks the consistency policy detailed in Section 4 in the main paper. If permitted, this thread randomly chooses a minibatch of samples from the training data, computes the SFs and pushes them to the input and output SF queue. The parameter update thread fetches SFs from the input SF queue and uses them to update the parameter matrix. In proximal-SGD/SDCA, the proximal/dual operator function (provided by the user) is automatically called by this thread as a function pointer. 
The communication thread receives SFs from other workers and pushes them into the input SF queue and sends SFs in the output SF queue to other workers, either in a full broadcasting scheme or a Halton sequence based partial broadcasting scheme. One worker is in charge of measuring the objective value. Once the algorithm converges, this worker notifies all other workers to terminate the job. We implemented SFBcaster in C++. OpenMPI was used for communication between workers and OpenMP was used for multicore parallelization within each machine.

%\section{Fault Tolerance and Elasticity}
The decentralized architecture of SFBcaster makes it robust to machine failures. If one worker fails, the rest of workers can continue to compute and broadcast the sufficient factors among themselves. In addition, SFBcaster possesses high elasticity \cite{li2014scaling}: new workers can be added and existing workers can be taken offline, without restarting the running framework.
A thorough study of fault tolerance and elasticity will be left for future work.

\section{Cost Analysis and Theory}
\label{sec:theory}

\begin{figure*}
%\vspace{-0.2in}
\small
\begin{tabularx}{\textwidth}{ X|X|X|X }
\hline
{\bf Computational Model} & Total comms, per iter & $\mb{W}$ storage per machine & $\mb{W}$ update time, per iter \\
\hline\hline
SFB (peer-to-peer, full broadcasting)  &  $O(P^2K(J+D))$ & $O(JD)$ & $O(P^2KJD)$ \\
\hline
SFB (peer-to-peer, Halton sequence broadcasting)  &  $O(P\log PK(J+D))$ & $O(JD)$ & $O(P\log PKJD)$ \\
\hline
FMS (client-server \cite{chilimbi2014project}) & $O(PJD)$ & $O(JD)$ & $O(PJD)$ at server, $O(PKJD)$ at clients \\
%\hline
%Single Machine & N/A & $O(JD)$ & $O(KPJD)$\\
\hline
\end{tabularx}

\caption{Cost of using SFB versus FMS, where $K$ is minibatch size, $J,D$ are dimensions of $\mb{W}$, and $P$ is the number of workers.
%\todoy{Second line last column, is it $O(KJD)$, instead of $O(JD)$ for each client? I added the single machine just for our own reference, maybe better NOT include it in the submission.}
%\qirong{Agree, it's $O(KJD)$. Changed.}
}
\label{tab:comms_cost}

\end{figure*}

We now examine the costs and convergence behavior of SFB under synchronous and bounded-async (e.g. SSP~\cite{BertsekasTsitsiklis89,ho2013more,dai2015high}) consistency, and show that SFB can be preferable to full-matrix synchronization/communication schemes.

%\noindent\textbf{Cost Analysis}:
\subsection{Cost Analysis}
Table \ref{tab:comms_cost} compares the communications, space and time (to apply updates to $\mb{W}$) costs of peer-to-peer SFB, against full matrix synchronization (FMS) under a client-server architecture \cite{chilimbi2014project}. For SFB with a full broadcasting scheme, in each minibatch, every worker broadcasts $K$ SF pairs $(\mb{u},\mb{v})$ to $P-1$ other workers, i.e. $O(P^2K(J+D))$ values are sent per iteration ---
%workers under the Halton sequence broadcasting scheme, where $Q$ is the number of outbound neighbors of each worker and is in the scale of $\log P$.
linear in matrix dimensions $J,D$, and quadratic in $P$. For SFB with a Halton sequence broadcasting scheme, every worker communicates SF pairs with $Q=O(P)$ peers, hence the communication cost is reduced to $O(P\log PK(J+D))$.
Because SF pairs cannot be aggregated before transmission, the cost has a dependency on $K$.
In contrast, the communication cost in FMS is $O(PJD)$, linear in $P$, quadratic in matrix dimensions, and independent of $K$. For both SFB and FMS, the cost of storing $\mb{W}$ is $O(JD)$ on every machine. As for the time taken to update $\mb{W}$ per iteration, FMS costs $O(PJD)$ at the server (to aggregate $P$ client update matrices) and $O(PKJD)$ at the $P$ clients (to aggregate $K$ updates into one update matrix for the server). By comparison, SFB bears a cost of $O(P^2KJD)$ under full broadcasting and $O(P\log PKJD)$ under Halton sequence broadcasting due to the additional overhead of reconstructing each update matrix $P$ times.

Compared with FMS, SFB achieves communication savings by paying an extra computation cost. 
%At first glance, it seems that SFB comes with both positive and negative tradeoffs --- however, 
In a number of practical scenarios, such a tradeoff is worthwhile. Consider large problem scales where $\min(J,D)\ge 10000$, and moderate minibatch sizes $1\le K\le 1000$ (as studied in this paper); when using a moderate number of machines (around 10-100), the $O(P^2K(J+D))$ communications cost of SFB is lower than the $O(PJD)$ cost for FMS, and the relative benefit of SFB improves as the dimensions $J,D$ of $\mb{W}$ grow. In data center scale computing environments with thousands of machines, we can adopt the Halton sequence broadcasting scheme under which the communication cost is linearithmic ($O(P\log P)$) in $P$. As for the time needed to apply updates to $\mb{W}$, it turns out that the additional cost of reconstructing each update matrix $P$ times in SFB is negligible in practice --- we have observed in our experiments that the time spent computing SFs, as well as communicating SFs over the network, greatly dominates the cost of reconstructing update matrices using SFs. Overall, the communication savings dominate the added computational overhead, which we validated in experiments (Section \ref{sec:exp}).

\subsection{Convergence Analysis}
We study the convergence of minibatch SGD under full broadcasting SFB (with extensions to proximal-SGD, SDCA and Halton sequence broadcasting being a topic for future study). Since SFB is a peer-to-peer decentralized computation model, we need to show the parameter copies on different workers converge to the same limiting point. This is different from the analyses of centralized parameter server systems \cite{ho2013more,dai2015high,agarwal2011distributed}, which show convergence of global parameters on the central server.

%\vspace{-0.1in}
%\subsection{Convergence Analysis}
%Given function $f_i: \mathcal{X}\subset \mathbb{R}^n \rightarrow \mathbb{R}$,
We wish to solve the optimization problem
$\min_{\mb{W}}~ \sum_{m=1}^{M}$
$ f_m(\mb{W})$,
where $M$ is the number of training data minibatches, and $f_m$ corresponds to the loss function on the $m$-th minibatch. Assume the training data minibatches $\{1,...,M\}$ are divided into $P$ disjoint subsets $\{S_1,...,S_P\}$ with $|S_p|$ denoting the number of minibatches in $S_p$. Denote $F =  \sum_{m=1}^{M} f_m$ as the total loss, and for $p=1,\ldots, P$, $F_p := \sum_{j\in S_p} f_j$ is the loss on $S_p$ (residing on the $p$-th machine). 

Consider a distributed system with $P$ machines. Each machine $p$ keeps a local variable $\mathbf{W}_p$ and the training data in $S_p$. At each iteration, machine $p$ draws one minibatch $I_p$ uniformly at random from partition $S_p$, and computes the partial gradient $\sum_{j\in I_p}\nabla f_j (\mathbf{W}_p)$. %In the discussion, we use capital letter $I, J$ to denote the index of minibatch sampled by machine $i,j$ from set $S_i, S_j$, respectively.  
Each machine updates its local variable by accumulating partial  updates from all machines. Denote $\eta_c$ as the learning rate at $c$-th iteration on every machine. The partial update generated by machine $p$ at its $c$-th iteration is %$u_{I_p^c}(\mathbf{W}_p^c) = -\eta_c |S_p| \nabla f_{I_p^c}(\mathbf{W}_p^c)$.  
denoted as $U_p(\mb{W}_p^c, I_p^c) = -\eta_c |S_p| \sum_{j\in I_p^c}\nabla f_{j}(\mathbf{W}_p^c)$. Note that $I_p^c$ is random and the factor $|S_p|$ is to restore unbiasedness in expectation.
Then the local update rule of machine $p$ is 
\begin{equation}
\label{eq:local}
\textstyle\mathbf{W}_p^c = \mathbf{W^0} + \sum_{q=1}^{P}\sum_{t=0}^{\tau_{p}^q(c)}  U_q(\mb{W}_q^t, I_q^t),\quad
 0\le (c-1)-\tau_{p}^q(c) \le s
\end{equation}
%\pengtao{should the subscript be q for $U_q(\mb{W}_q^t, I_q^t)$?}
where $\mb{W}^0$ is the common initializer for all $P$ machines, and $\tau_{p}^q(c)$ is the number of iterations machine $q$ has transmitted to machine $p$ when machine $p$ conducts its $c$-th iteration. Clearly, $\tau_p^p(c) = c$. Note that we also require $\tau_p^q(c) \leq c-1$, \ie, machine $p$ will not use any partial updates of machine $q$ that are too fast forward. This is to avoid correlation in the theoretical analysis.
Hence, machine $p$ (at its $c$-th iteration) accumulates updates generated by machine $q$ up to iteration $\tau_{p}^q(c)$, which is restricted to be at most $s$ iterations behind.
%In full synchronization, $\forall i, c_{i} = c-1$ so that all fresh updates are accumulated. While
%In SFB, we allow each machine accumulates updates up to a delayed clock, i.e., $c_{i}\le c-1$. We also require the clock difference $(c-1)-c_{i}$ to be $\le s$,
Such bounded-asynchronous communication addresses the slow-worker problem caused by bulk synchronous execution, while ensuring that the updates accumulated by each machine are not too outdated. 
%\todoy{Communication delay is not yet modelled.}
The following standard assumptions are needed for our theoretical analysis:
\begin{assum}\label{ass:model_3}
%The objective fuction satisfies:
(1) For all $j$, $f_j$ is continuously differentiable and $F$ is bounded from below; (2) $\nabla F$, $\nabla F_{p}$ are Lipschitz continuous with constants $L_F$ and $L_p$, respectively, and let $L = \sum_{p=1}^{P} L_p$; (3) There exists $B, \sigma^2$ such that for all $p$ and $c$, we have (almost surely) $\|\mathbf{W}_p^c\|\le B$ and $\mathbb{E}\| ~|S_p|\sum_{j\in I_p}\nabla f_{j}(\mathbf{W}) - \nabla F_{p}(\mathbf{W}) ~\|_2^2\le \sigma^2$.
\end{assum}

\begin{figure*}[t]
\begin{center}
\includegraphics[width=0.5\columnwidth]{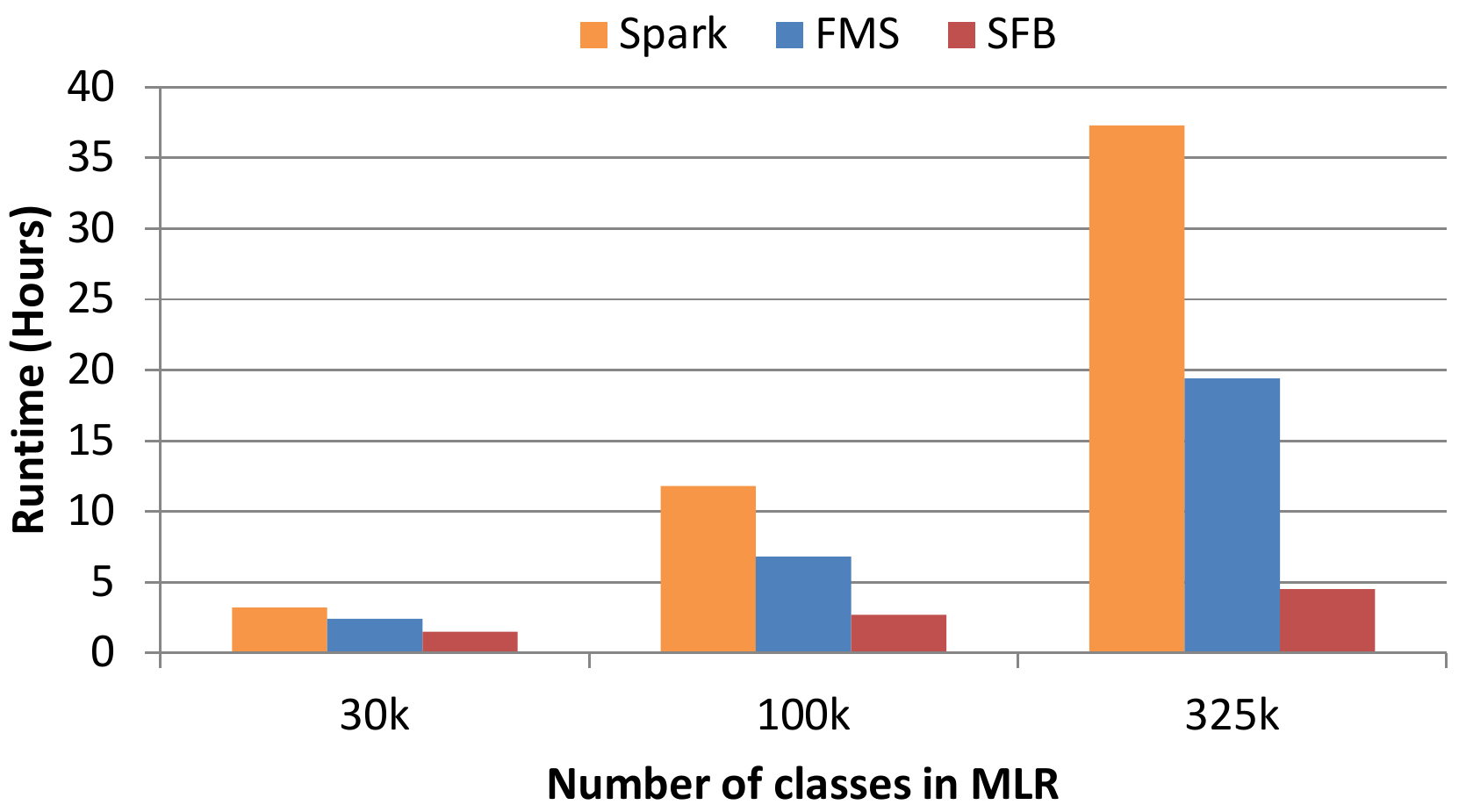}
\includegraphics[width=0.5\columnwidth]{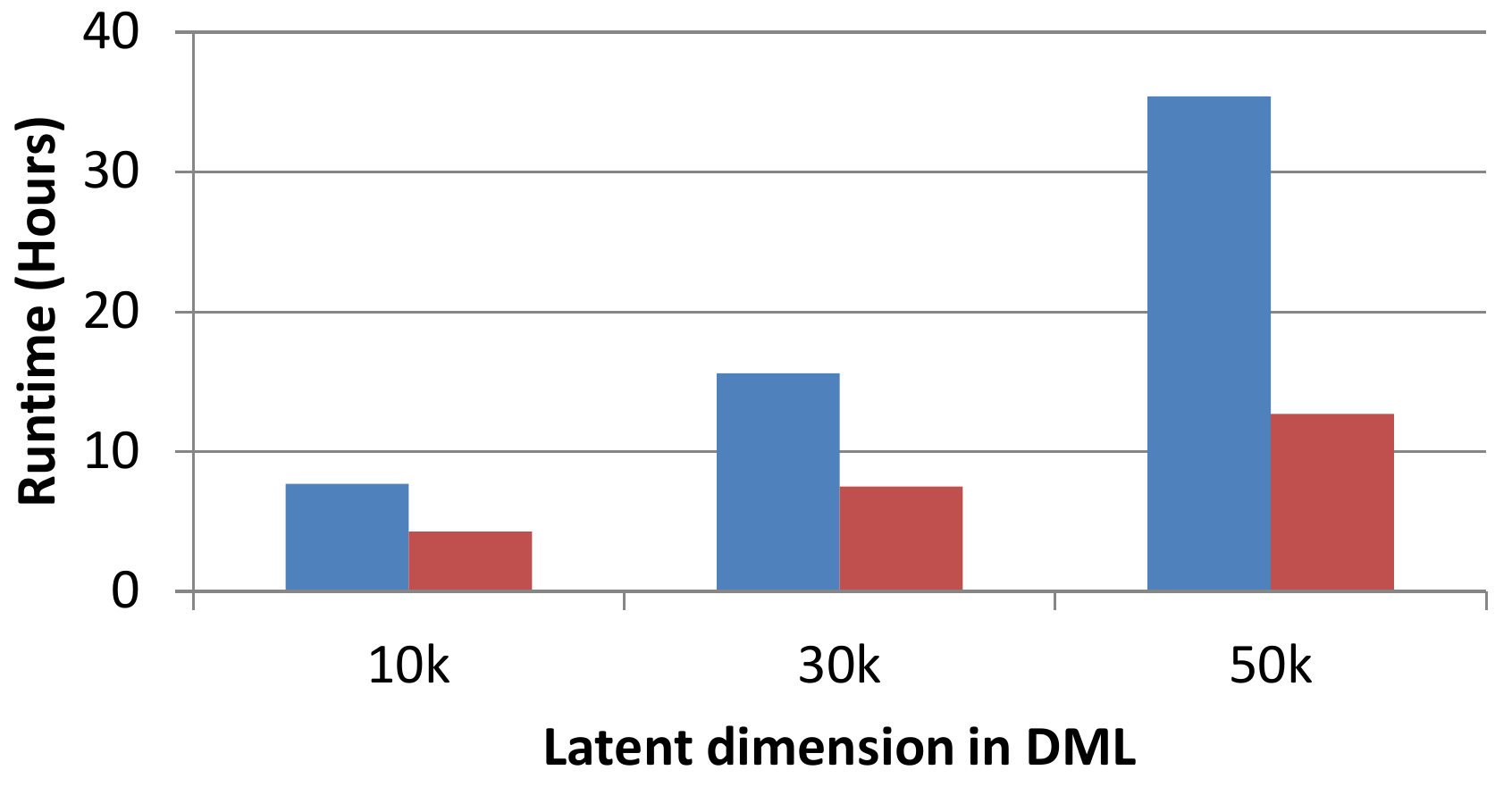}
\includegraphics[width=0.5\columnwidth]{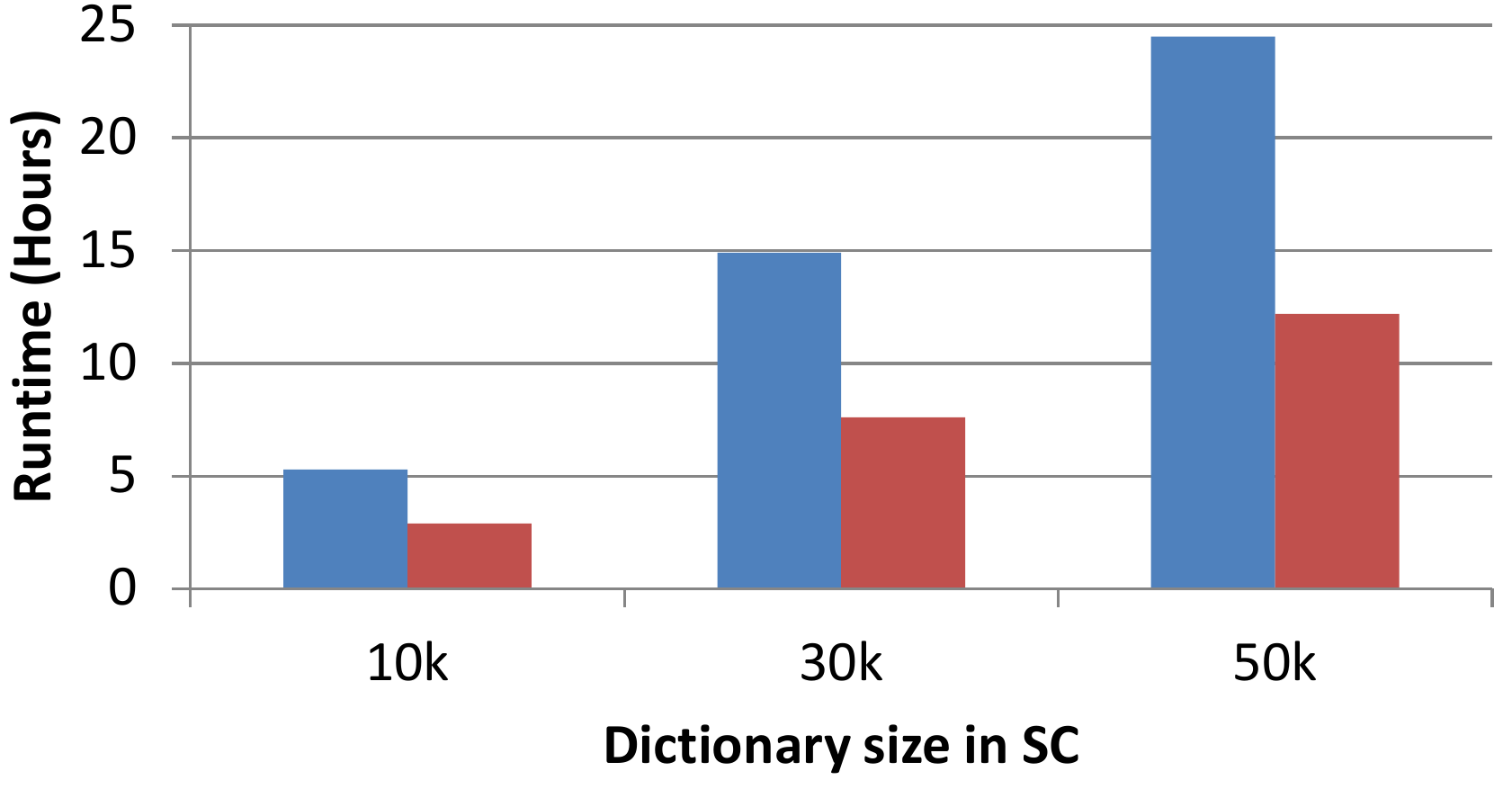}
\includegraphics[width=0.5\columnwidth]{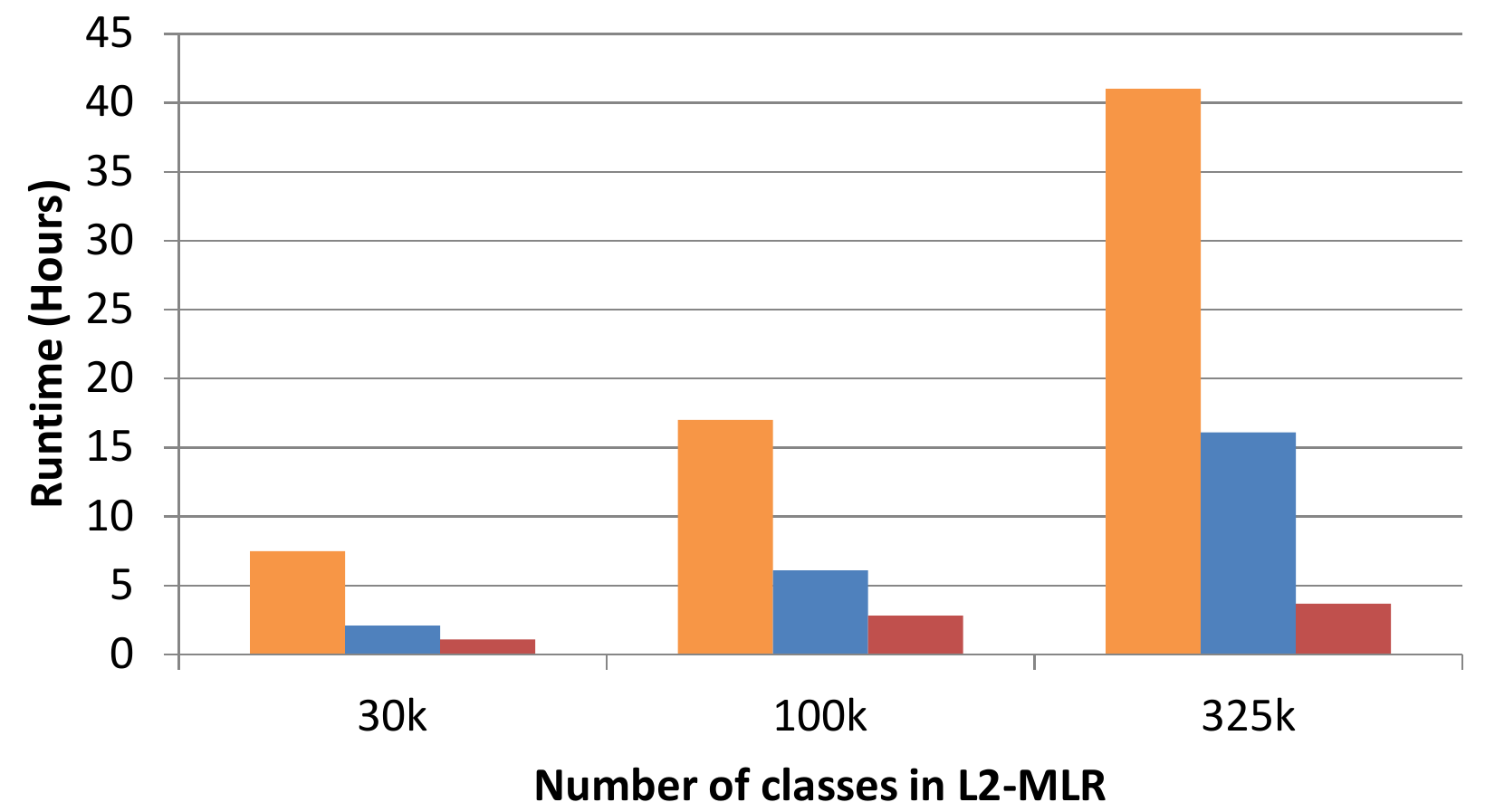}

\caption{Convergence time versus model size for MLR, DML, SC, L2-MLR (left to right), under BSP.}
\label{fig:exp_runtime}
\end{center}
\vspace{-0.2in}
\end{figure*}

\begin{figure*}[t]
\begin{center}
\includegraphics[width=0.5\columnwidth]{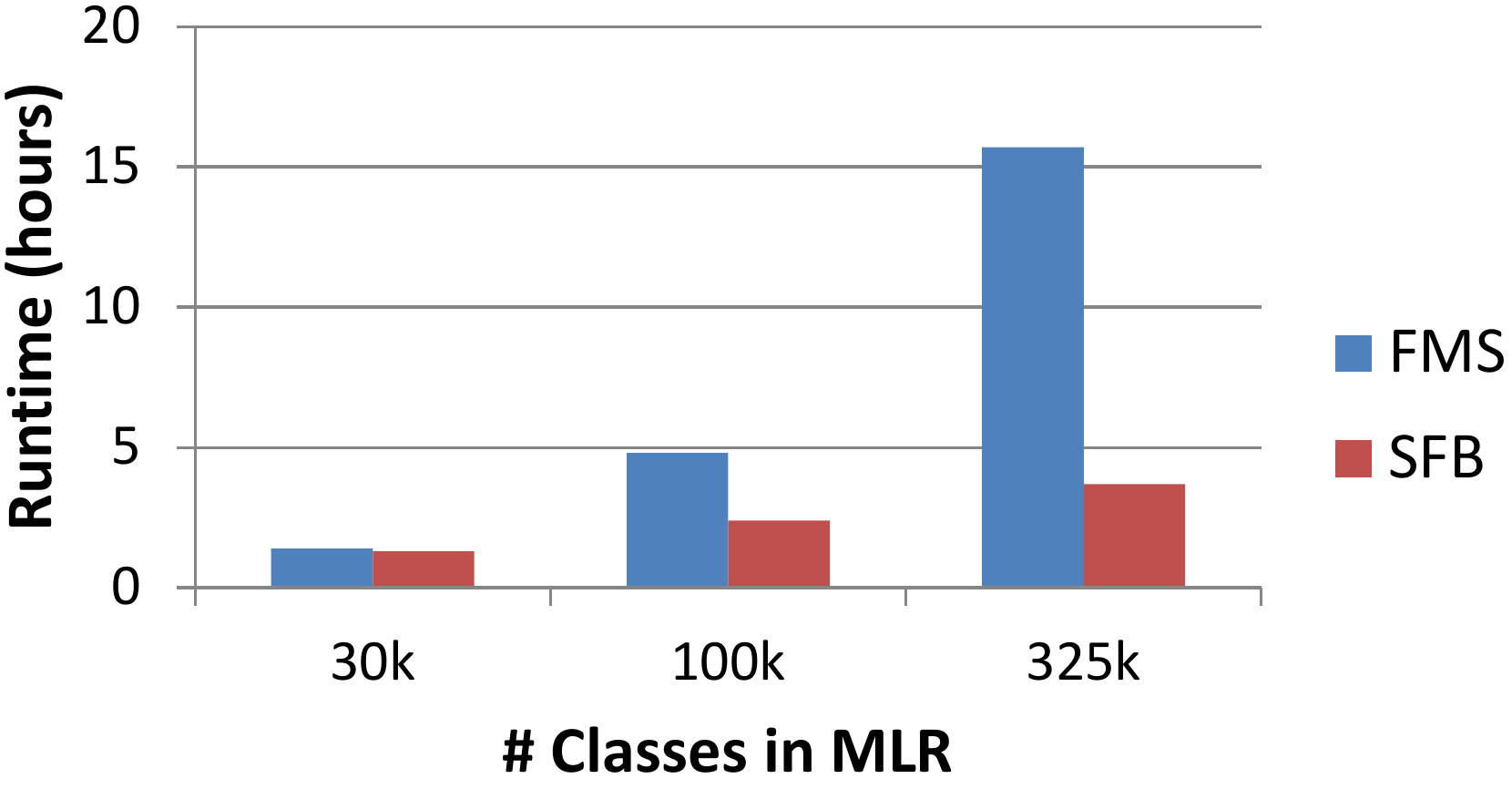}
\includegraphics[width=0.5\columnwidth]{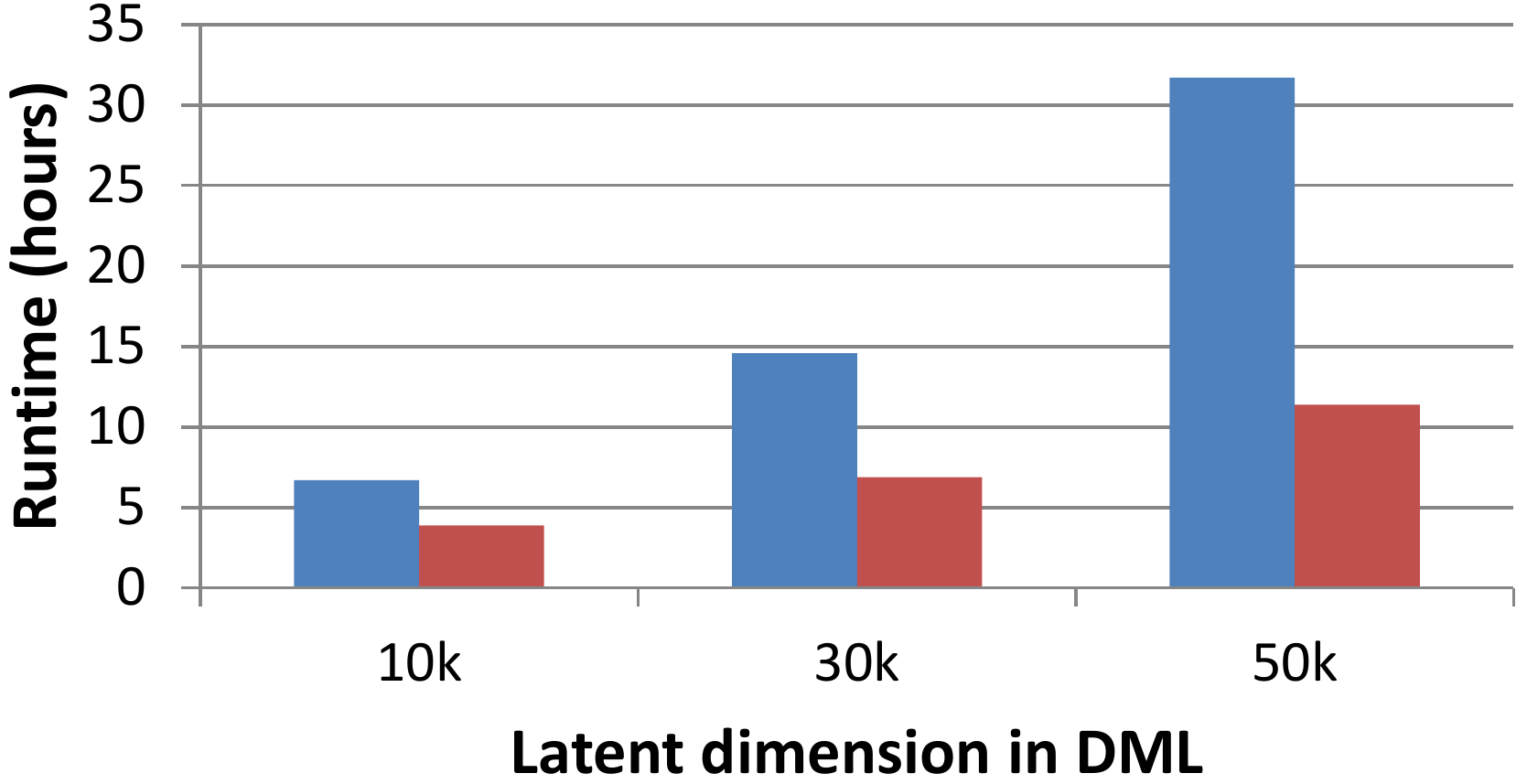}
\includegraphics[width=0.5\columnwidth]{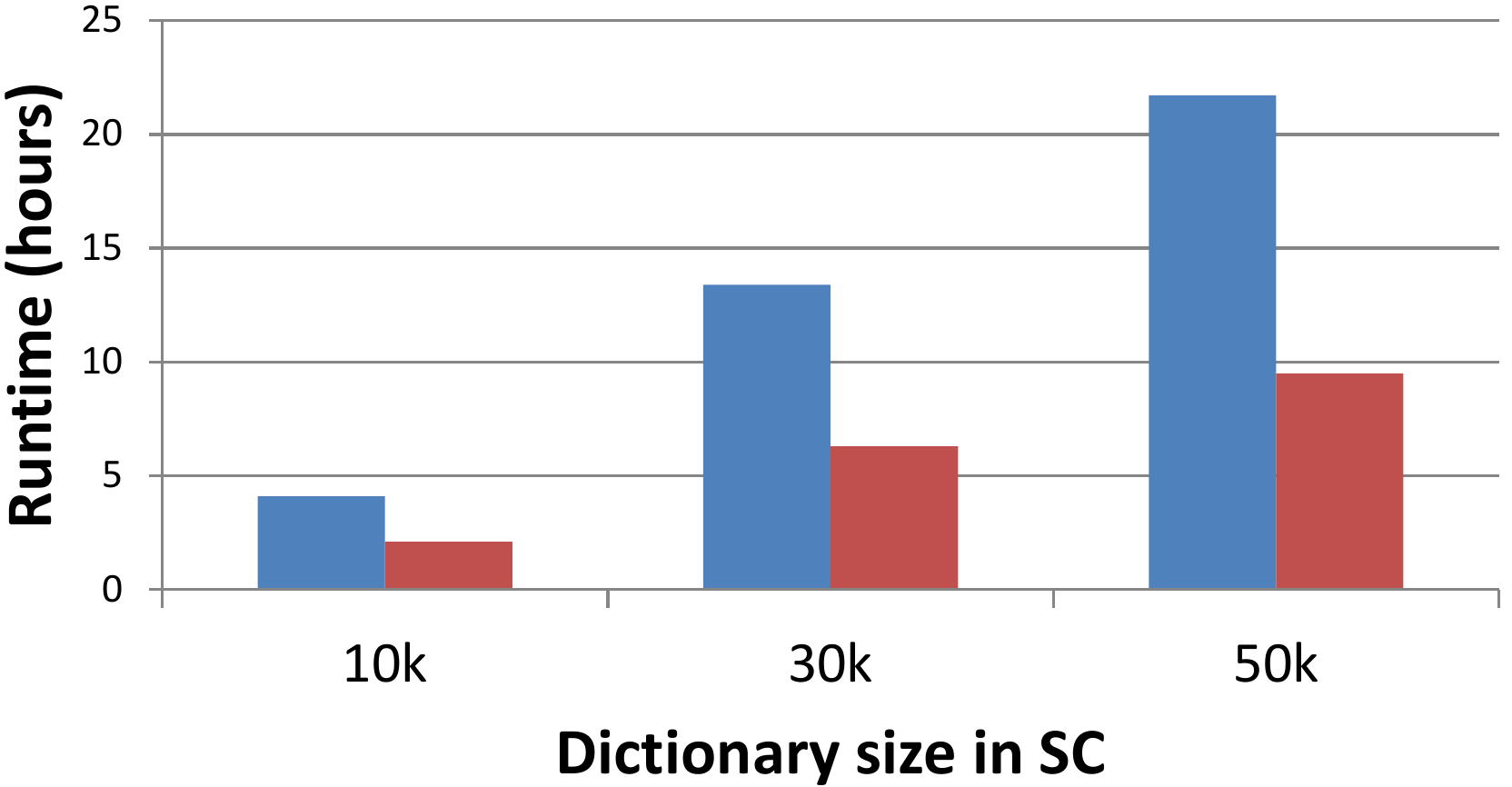}
\includegraphics[width=0.5\columnwidth]{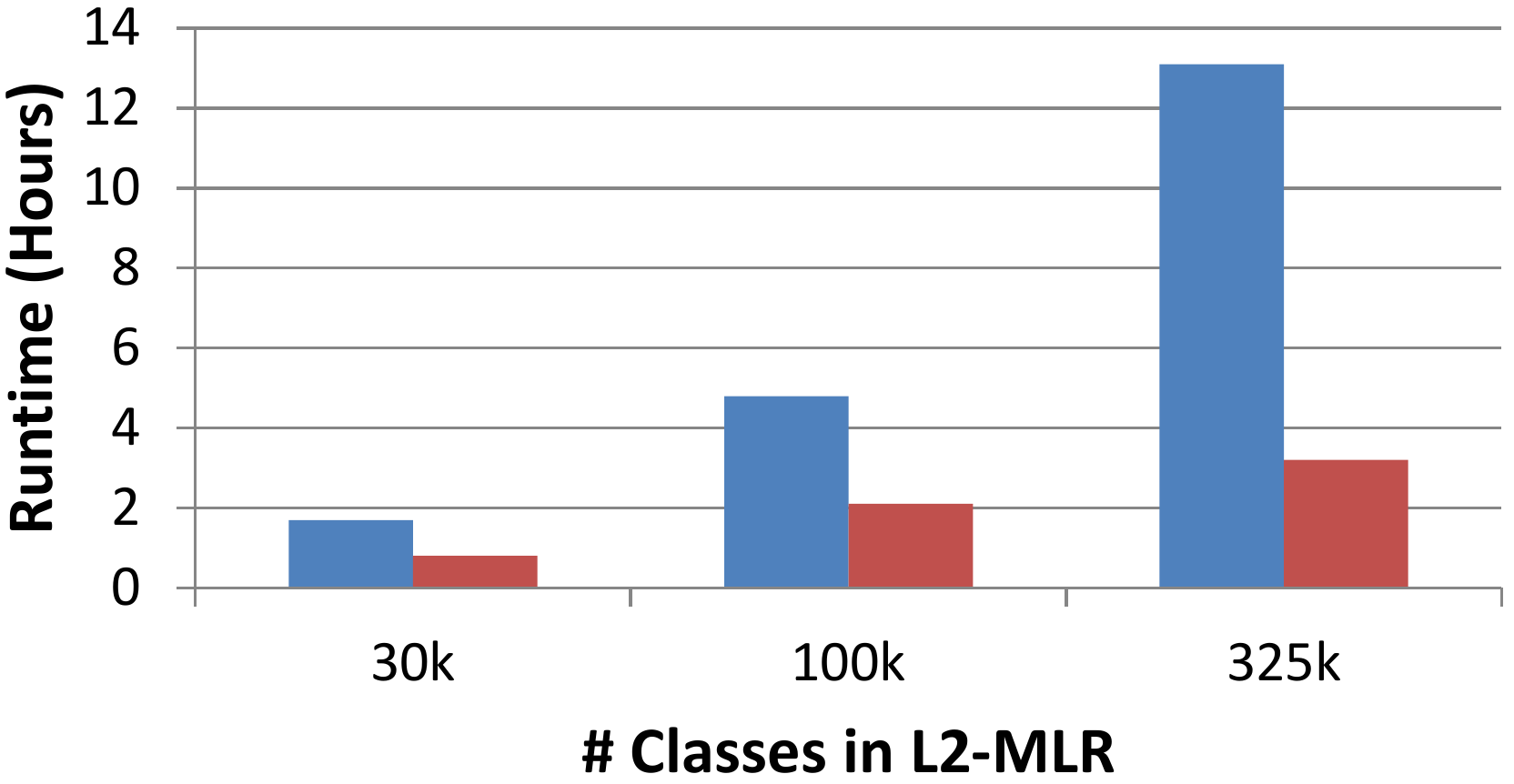}
\vspace{-0.1in}
\caption{Convergence time versus model size for MLR, DML, SC, L2-MLR (left to right), under SSP with staleness=20.}
\vspace{-0.2in}
\label{fig:exp_runtime_ssp}
\end{center}

\end{figure*}

\begin{figure*}[t]
\begin{center}
\includegraphics[width=0.6\columnwidth]{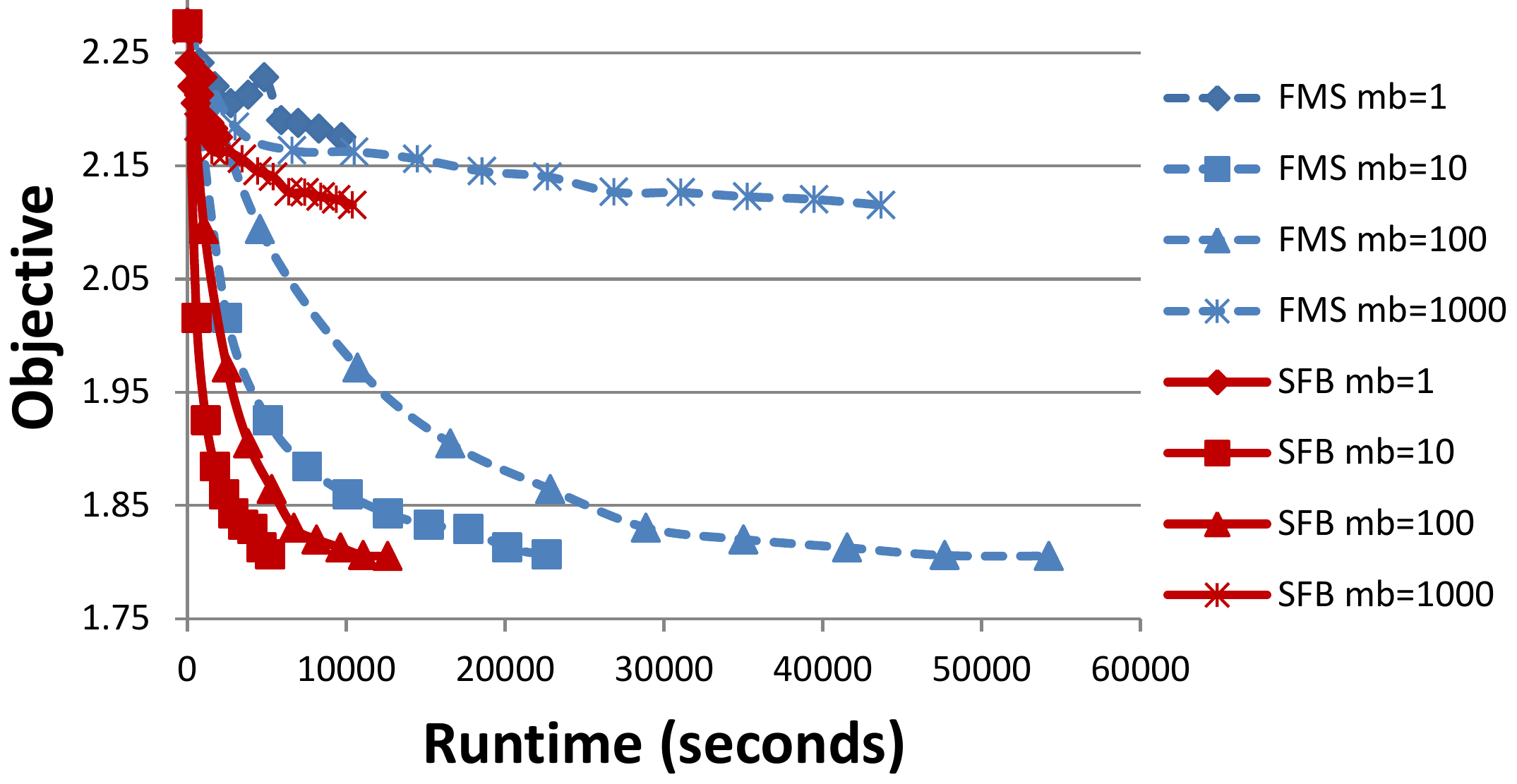}
\includegraphics[width=0.6\columnwidth]{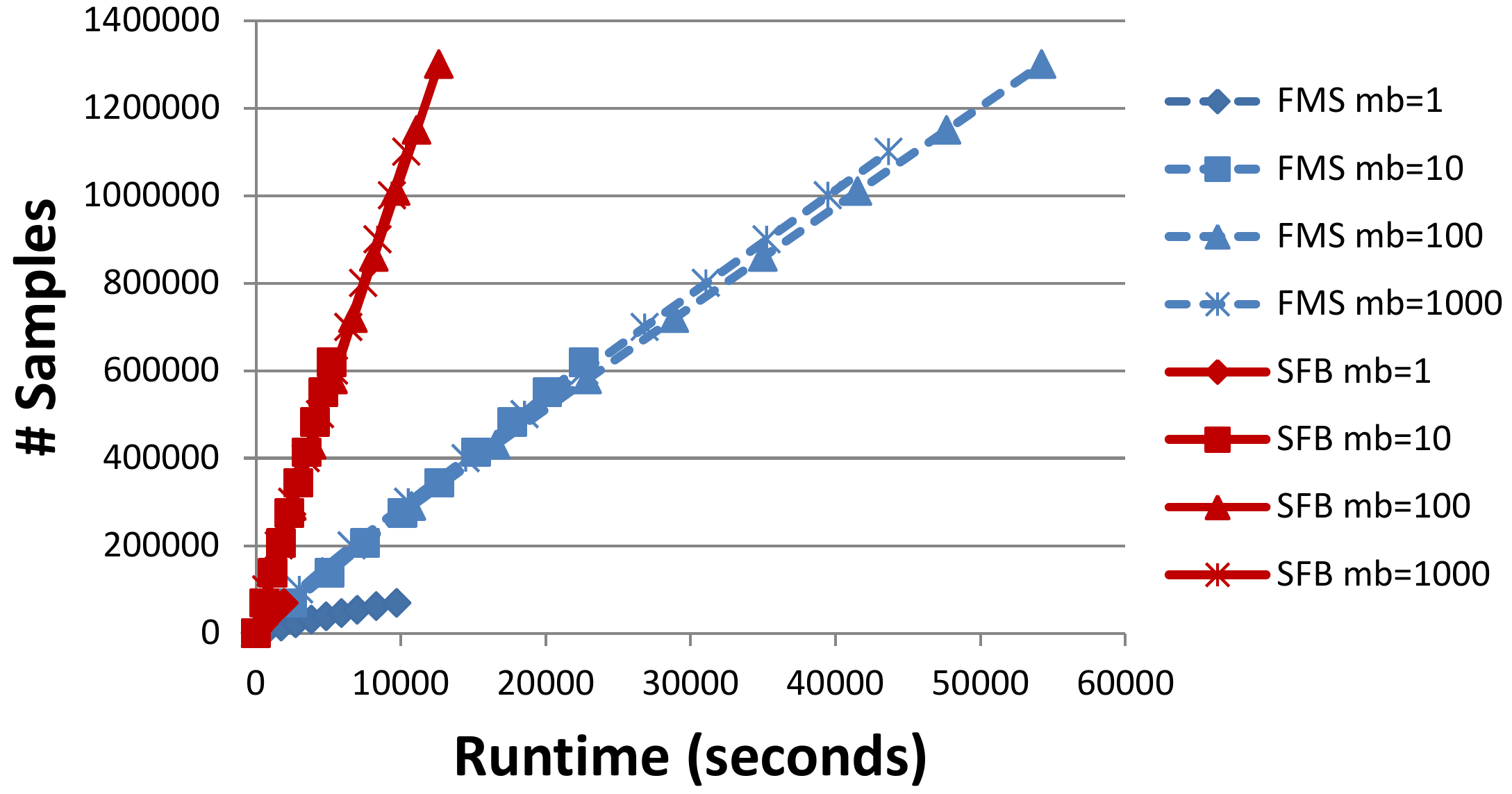}
\includegraphics[width=0.6\columnwidth]{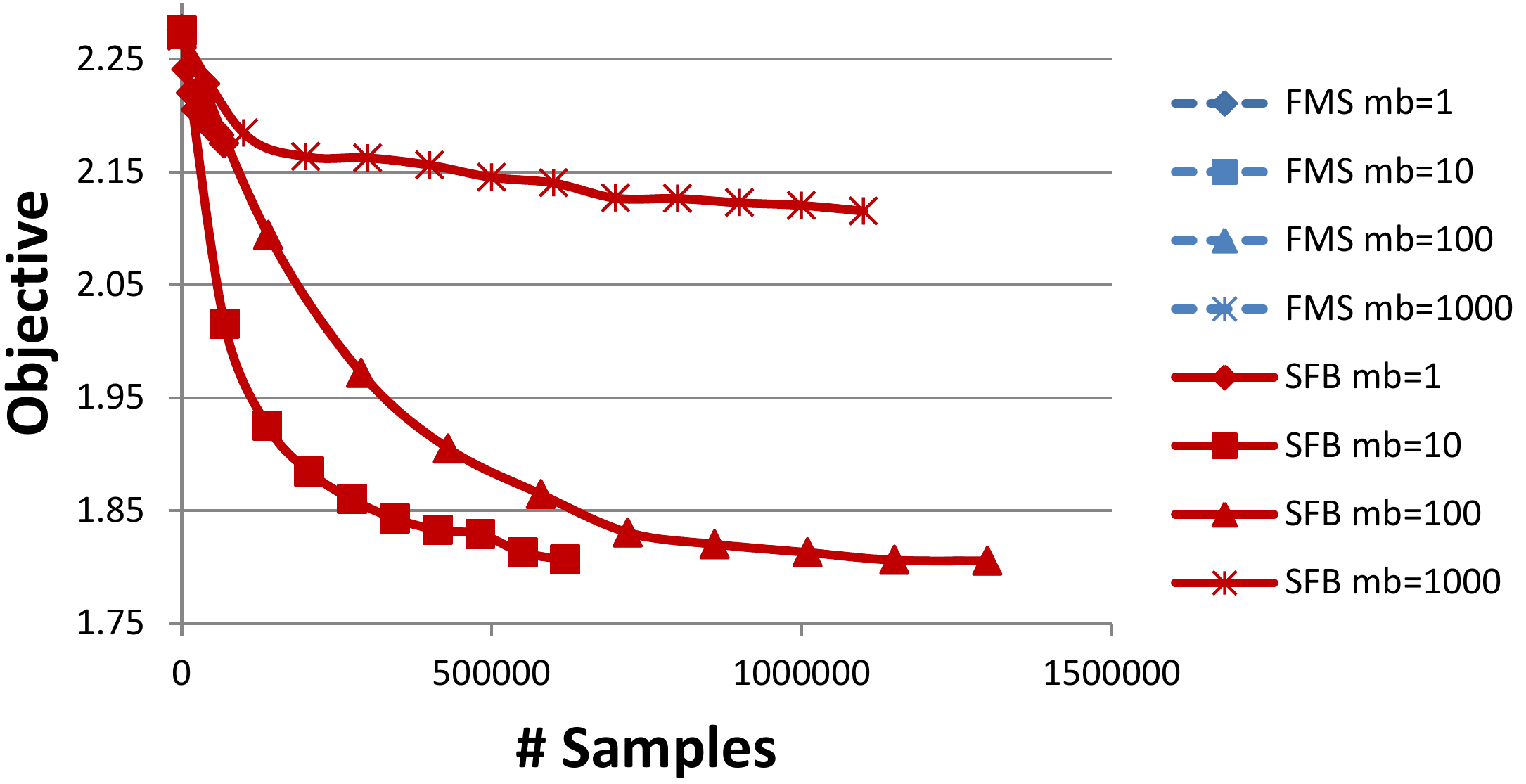}
\vspace{-0.1in}
\caption{MLR objective vs runtime (left), samples vs runtime (middle),  objective vs samples (right).}
%\qirong{We really need different line patterns and markers for each trend; the reviewers will definitely complain. Also need a way to indicate that many of the FMS and SFB trends overlap. Recommend that the samples vs runtime graph be plotted as a bar chart, for a start.}
\label{fig:iter_qtt_qlt}
\vspace{-0.2in}
\end{center}

\end{figure*}

Our analysis is based on the following auxiliary update
\begin{flalign}
\textstyle\mathbf{W}^c &= \textstyle\mathbf{W}^0 + \sum_{q=1}^{P}\sum_{t=0}^{c-1} U_q(\mb{W}_q^t, I_q^t),
\end{flalign}
%\pengtao{again, should it be $U_q(\mb{W}_q^t, I_q^t)$?}
Compare to the local update \eqref{eq:local} on machine $p$, essentially this auxiliary update accumulates all $c-1$ updates generated by all machines, instead of the $\tau_p^q(c)$ updates that machine $p$ has access to.
%This auxiliary sequence is easier to analyze since it contains no delay in the update rule.
We show that all local machine parameter sequences are asymptotically consistent with this auxiliary sequence:

\begin{thm}\label{thm:model_2}
Let $\{\mathbf{W}_p^c\}$, $p=1, \ldots, P$, and $\{\mathbf{W}^c\}$ be the local sequences and the auxiliary sequence generated by SFB for problem $\textbf{(P)}$ (with $h \equiv 0$), respectively.
Under \Cref{ass:model_3} and set the learning rate $\eta_c^{-1} = \frac{L_F}{2}+2sL +\sqrt{c}$, then we have
\begin{itemize}[leftmargin=*,topsep=0pt,noitemsep]
%\small
\item $\liminf\limits_{c\to\infty} \mathbb{E}\|\nabla F(\mathbf{W}^c)\|=0$, hence there exists a subsequence of $\nabla F(\mb{W}^c)$ that almost surely vanishes; %i.e. the limit points of $\{\mathbf{W^c}\}$ are stationary points of $F$ in expectation.
\item $\lim\limits_{c\to\infty} \max_p \|\mathbf{W}^c - \mathbf{W}_p^c\| = 0$, \ie the maximal disagreement between all local sequences and the auxiliary sequence converges to 0 (almost surely);
%all local sequences are asymptotically consistent with $\{\mathbf{W^c}\}$.
\item There exists a common subsequence of $\{\mb{W}_p^c\}$ and $\{\mb{W}^c\}$ that converges almost surely to a stationary point of $F$, with the rate
$\underset{c\le C}{\mathrm{min}}~\mathbb{E}\|\sum_{p=1}^{P}\nabla F_{p}(\mathbf{W}_p^c)\|_2^2 \le O\left(\frac{(L+L_F)\sigma^2 P s\log C}{\sqrt{C}}\right)
$
%\qirong{I'm confused by this; why is it valuable for us to look at the minimum over all $c\le C$, rather than the maximum? Is this some type of existence proof?}
%\todoy{It is impossible to look at the maximum, for instance, the initialization may be too bad, and taking maximum always includes it. In subgradient-type analysis, usually we take the minimum or the average.}
%\qirong{I see, but could you then explain the intuition? I can appreciate the average being bounded, but bounding the minimum seems rather weak --- I can't really see why it is useful to know this fact. If we can't give a good explanation, I feel it would be better to leave it in the appendix, than risk the reviewers questioning why we included such a result.}
%\todoy{Here is something we didn't try: the average parameter matrix of $P$ machines converges to 0? For convex $F_p$ we can derive a $O(\log(C)/\sqrt{C})$ rate after averaging. This may be easy to interpret, but of course we do not have the average matrix.}
\end{itemize}
\end{thm}
Intuitively, Theorem~\ref{thm:model_2} says that, given a properly-chosen learning rate, all local worker parameters $\{\mb{W}^c_p\}$ eventually converge to stationary points (i.e. local minima) of the objective function $F$, despite the fact that SF transmission can be delayed by up to $s$ iterations. Thus, SFB learning is robust even under bounded-asynchronous communication (such as SSP). 
Our analysis differs from \cite{BertsekasTsitsiklis89} in two ways:  (1) \cite{BertsekasTsitsiklis89} explicitly maintains a consensus model which would require transmitting the parameter matrix among worker machines --- a communication bottleneck that we were able to avoid;
(2) we allow subsampling in each worker machine. Accordingly, our theoretical guarantee is probabilistic, instead of the deterministic one in \cite{BertsekasTsitsiklis89}. 
%In large scale problems, this is hardly an issue.

%When the sufficient vectors are sparse, the communication cost of both SFB and CPS can be reduced. Let $S$ denote the average percentage of nonzeros entries in the SFs, then the cost of SFB and CPS becomes $O(P\log PS(J+D))$ and $O(PS^2JD)$ respectively. Note that when the sufficient vectors are highly sparse (equivalently $S$ is very small), the communication advantage of SFB over CPS diminishes. 
%In terms of computation, in each iteration of SGD, SFB bears a gradient reconstruction cost which is $O(P\log PKJD)$, as opposed to CPS where the cost is $O(PKJD)$. Compared with network communication, computation is much faster. Thereby, in SFB, the increased computation cost is moderate compared with the reduced communication cost.

\section{Experiments}
\label{sec:exp}

We demonstrate how four popular models can be efficiently learnt using SFB: (1) multiclass logistic regression (MLR) and distance metric learning (DML)\footnote{For DML, we use the parametrization proposed in \cite{weinberger2005distance}, which learns a linear projection matrix $\mb{L}\in \mathrm{R}^{d\times k}$, where $d$ is the feature dimension and $k$ is the latent dimension.} based on SGD; (2) sparse coding (SC) based on proximal SGD; (3) $\ell_2$ regularized multiclass logistic regression (L2-MLR) based on SDCA. For baselines, we compare with (a) Spark \cite{zaharia2012resilient} for MLR and L2-MLR, and (b) full matrix synchronization (FMS) implemented on open-source parameter servers \cite{ho2013more,li2014scaling} for all four models. In FMS, workers send update matrices to the central server, which then sends up-to-date parameter matrices to workers\footnote{This has the same communication complexity as \cite{chilimbi2014project}, which sends SFs from clients to servers, but sends full matrices from servers to clients (which dominates the overall cost).}. 
Due to data sparsity, both the update matrices and sufficient factors are sparse; we use this fact to reduce communication and computation costs.
%We did not compare with the server-client architecture \cite{chilimbi2014project} hybridizing vector communication (from workers to server) and matrix communication (from server to workers) since (1) its communication complexity is in the same order as FMS, both of which are quadratic to matrix dimensions; (2) their implementation is not publicly available. 
%\qirong{Reviewers don't really worry about using the same minibatch. I'm leaving it out to save space.}
%In each iteration of SGD/SDCA, all methods use the same minibatch for a deterministic comparison.
%\todoy{sorry for being a trouble-maker, but I just wish to exhaust all possible questions from the reviewers side that we might have to address: due to the high communication cost of sending the update matrix, this alternative approach should use a much larger mini-batch size, to average down the total time. Namely, if communication costs $T$ time, then it is optimal to use a mini-batch where each machine will spend roughly $T$ time on it. Otherwise one of communication and computation will be the bottleneck. Thus for fair comparison, this alternative should use a larger mini-batch size than SFB.}
%\qirong{No, these are very good questions. It is a good thing that Pengtao has new minibatch experiments that will help answer these questions... will edit the text after the graphs are in.}
Our experiments used a 12-machine cluster; each machine has 64 2.1GHz AMD cores, 128G memory, and a 10Gbps network interface.

\subsection{Datasets and Experimental Setup}

\begin{figure}[t]
\begin{center}
\includegraphics[width=0.49\columnwidth]{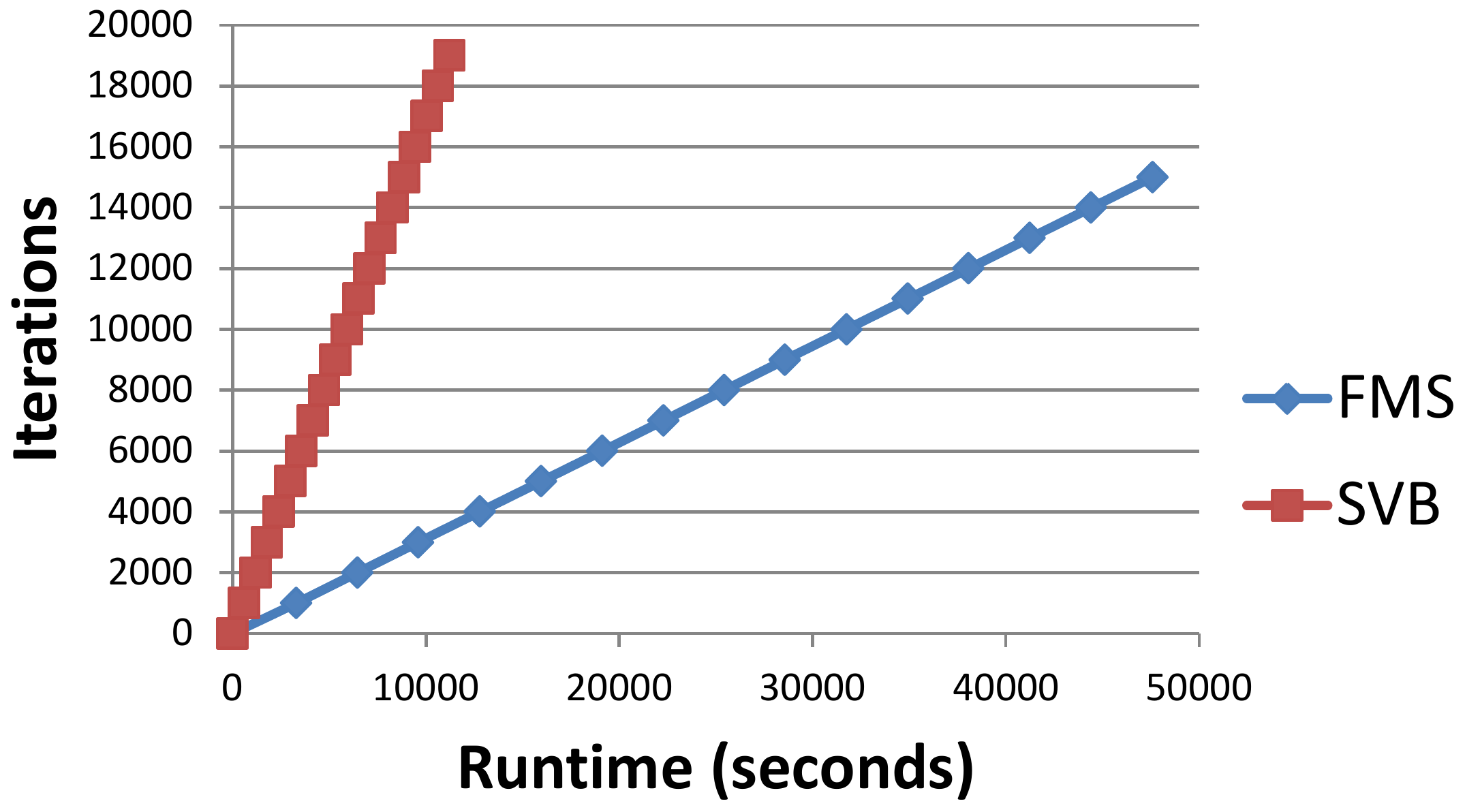}
\includegraphics[width=0.49\columnwidth]{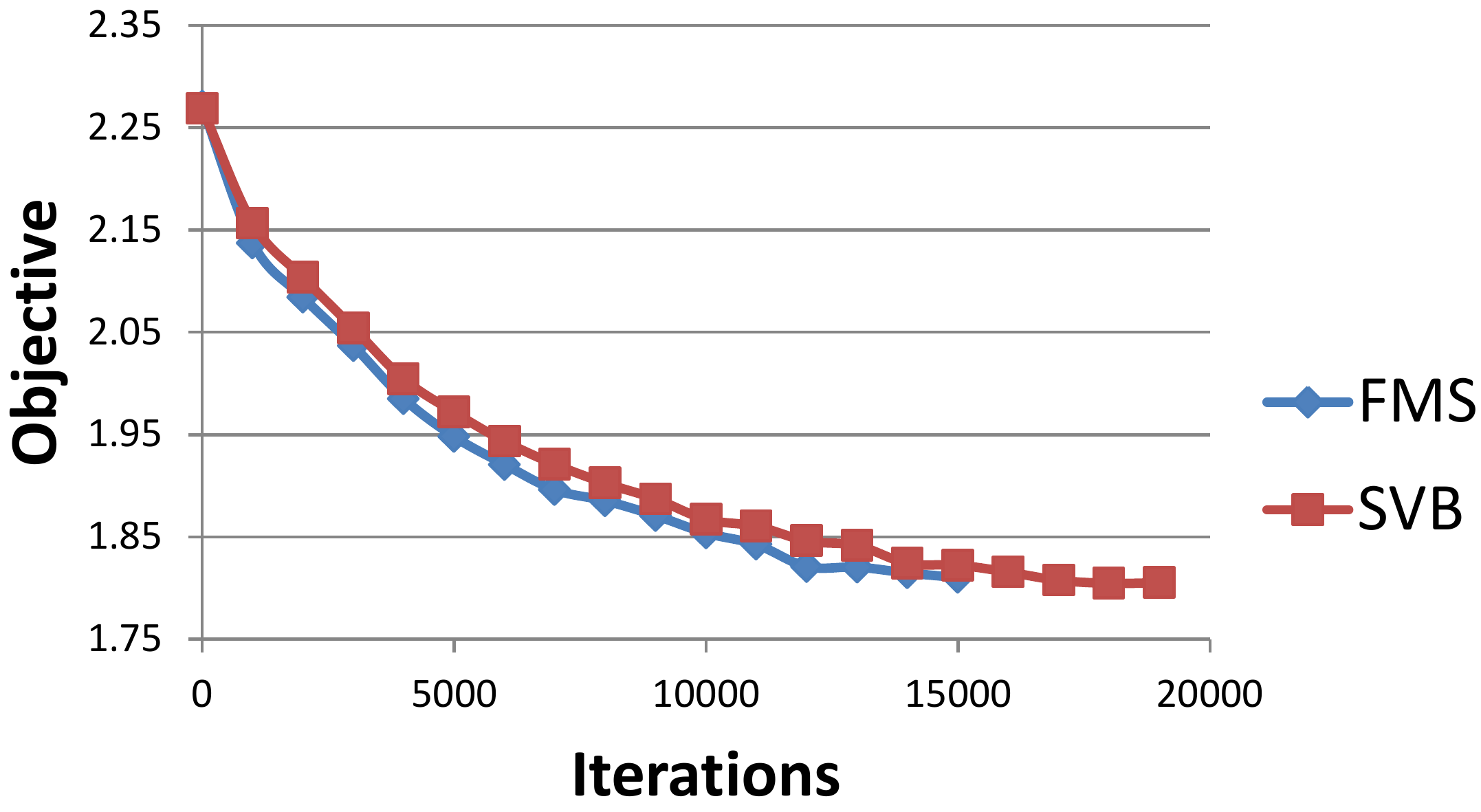}
\vspace{-0.1in}
\caption{MLR iteration throughput (left) and iteration quality (right) under SSP (staleness=20).}
\label{fig:iter_qtt_qlt_bsp_ssp}
\end{center}
\vspace{-0.2in}
\end{figure}

\begin{figure*}[t]
\begin{center}
\includegraphics[width=0.5\columnwidth]{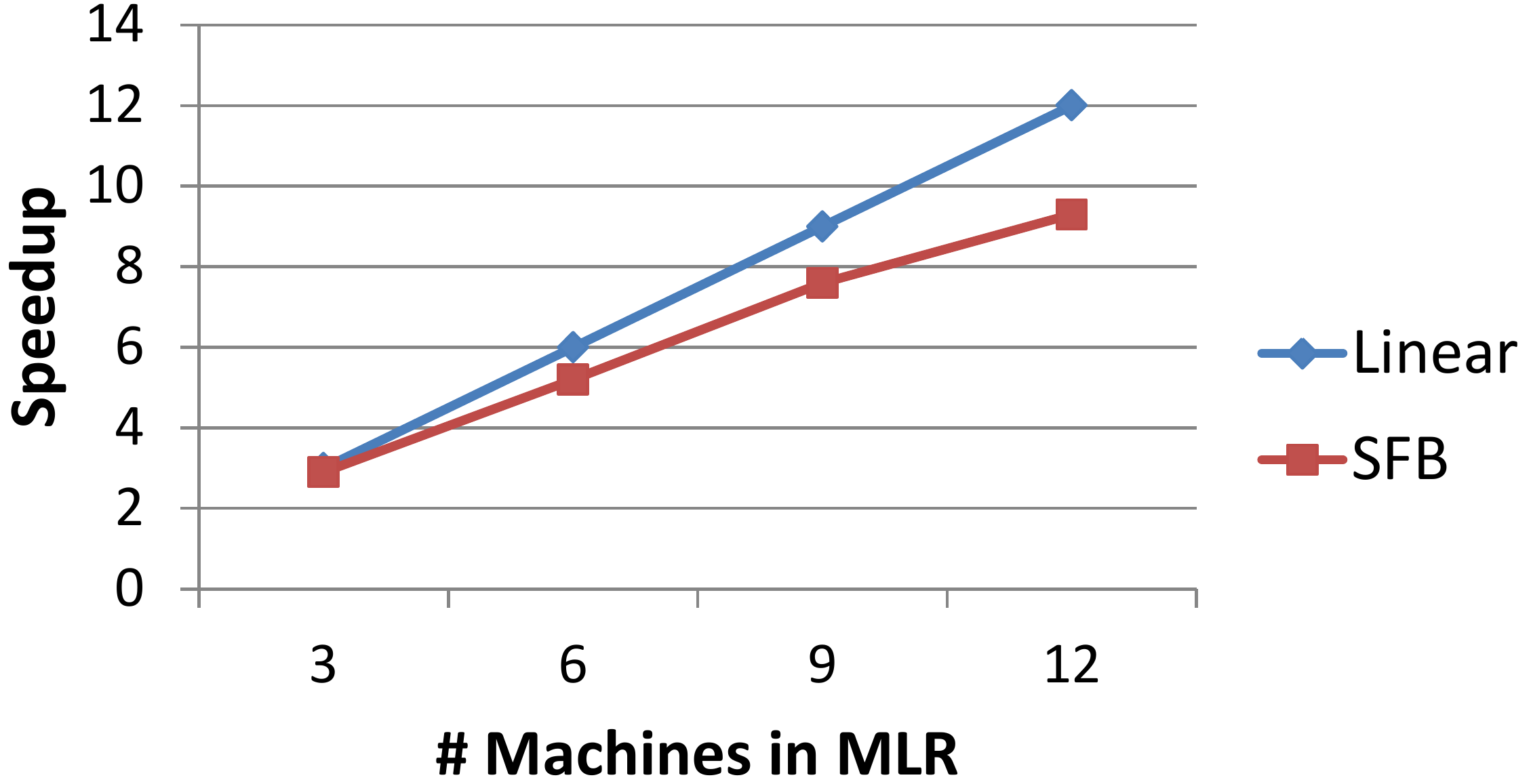}
\includegraphics[width=0.5\columnwidth]{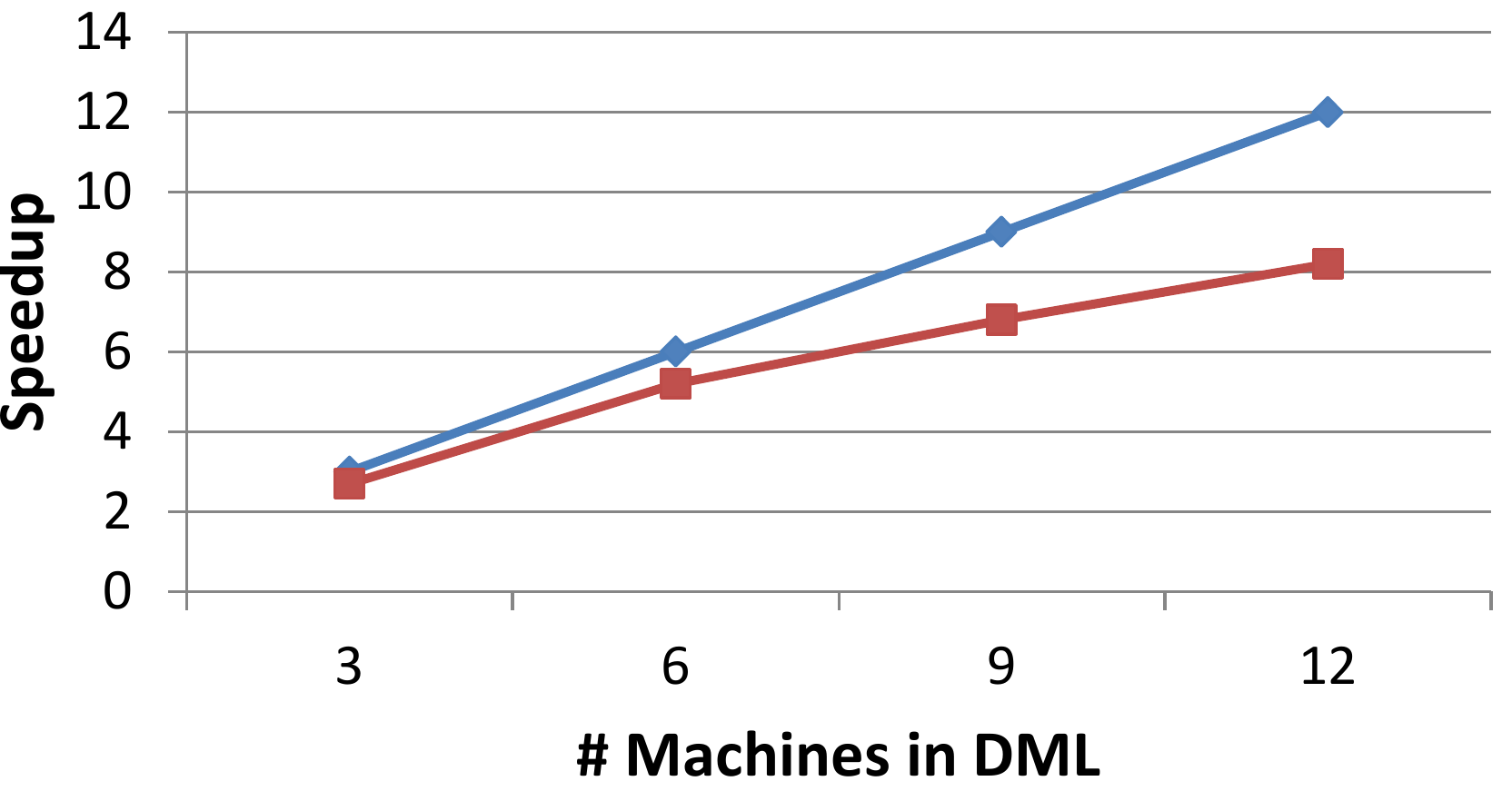}
\includegraphics[width=0.5\columnwidth]{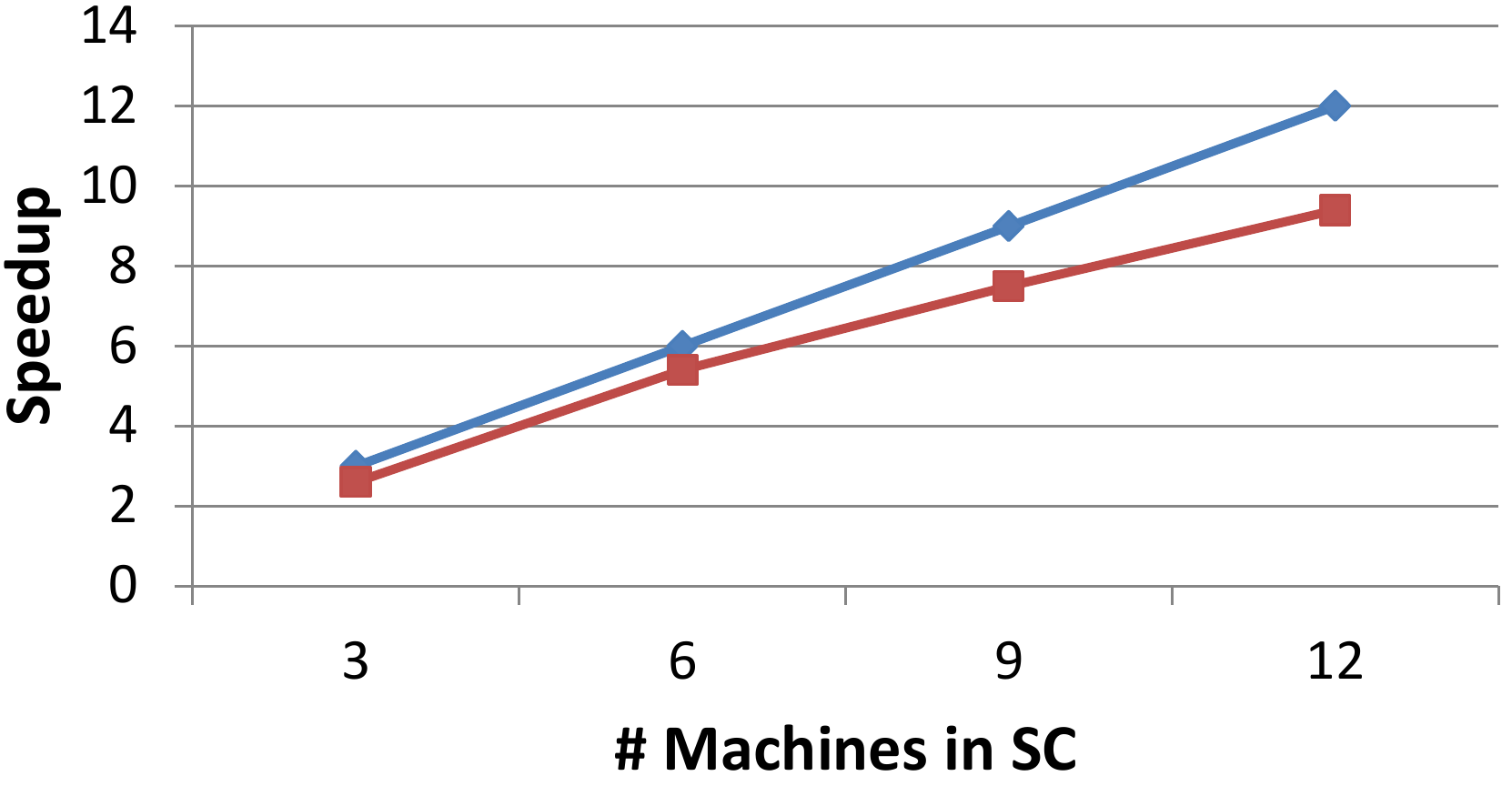}
\includegraphics[width=0.5\columnwidth]{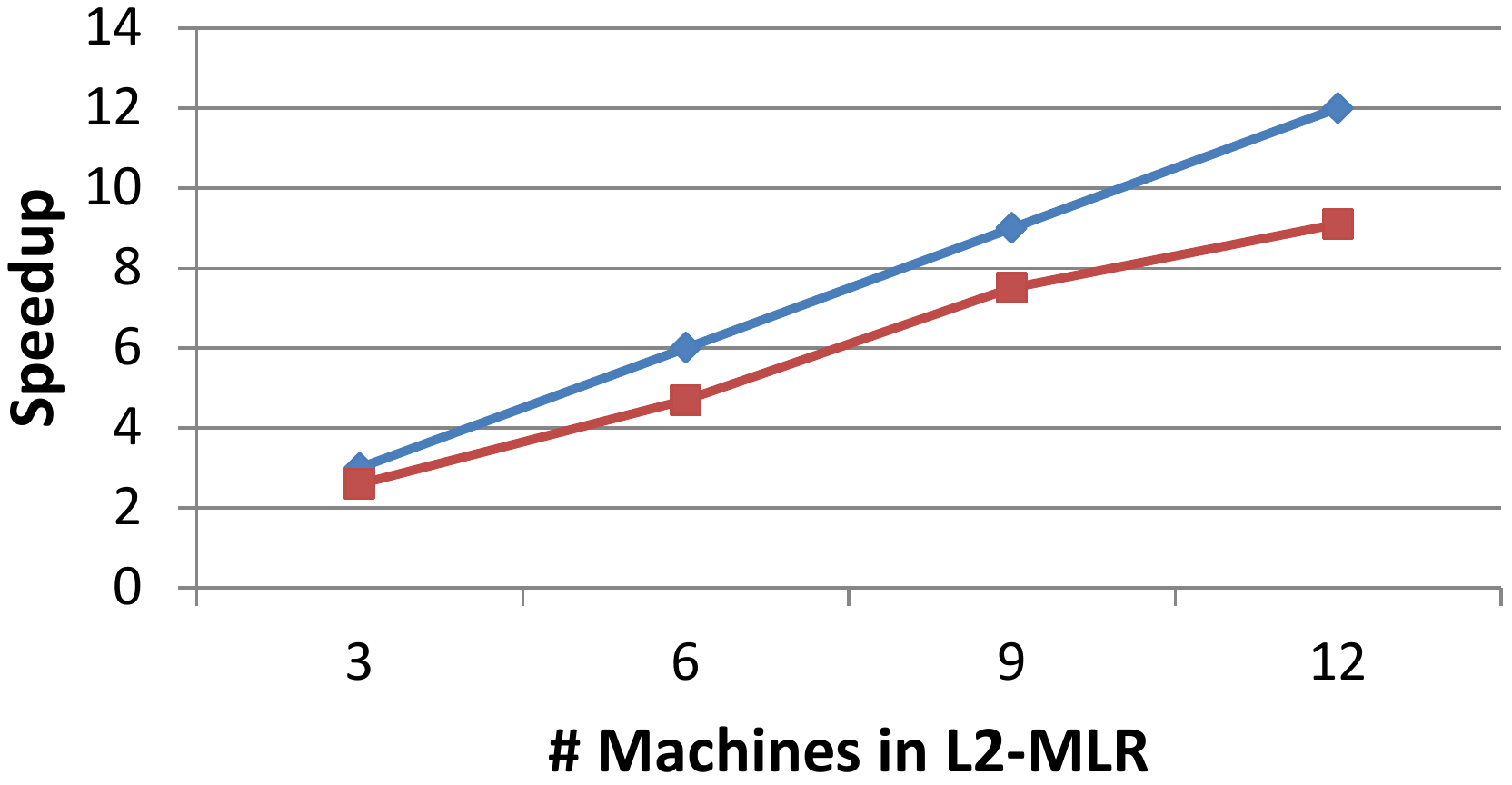}
\vspace{-0.1in}
\caption{SFB scalability with varying machines under BSP, for MLR, DML, SC, L2-MLR (left to right).}
\label{fig:exp_scalability}
\end{center}
\vspace{-0.2in}
\end{figure*}

\begin{figure*}[t]
\begin{center}
\includegraphics[width=0.5\columnwidth]{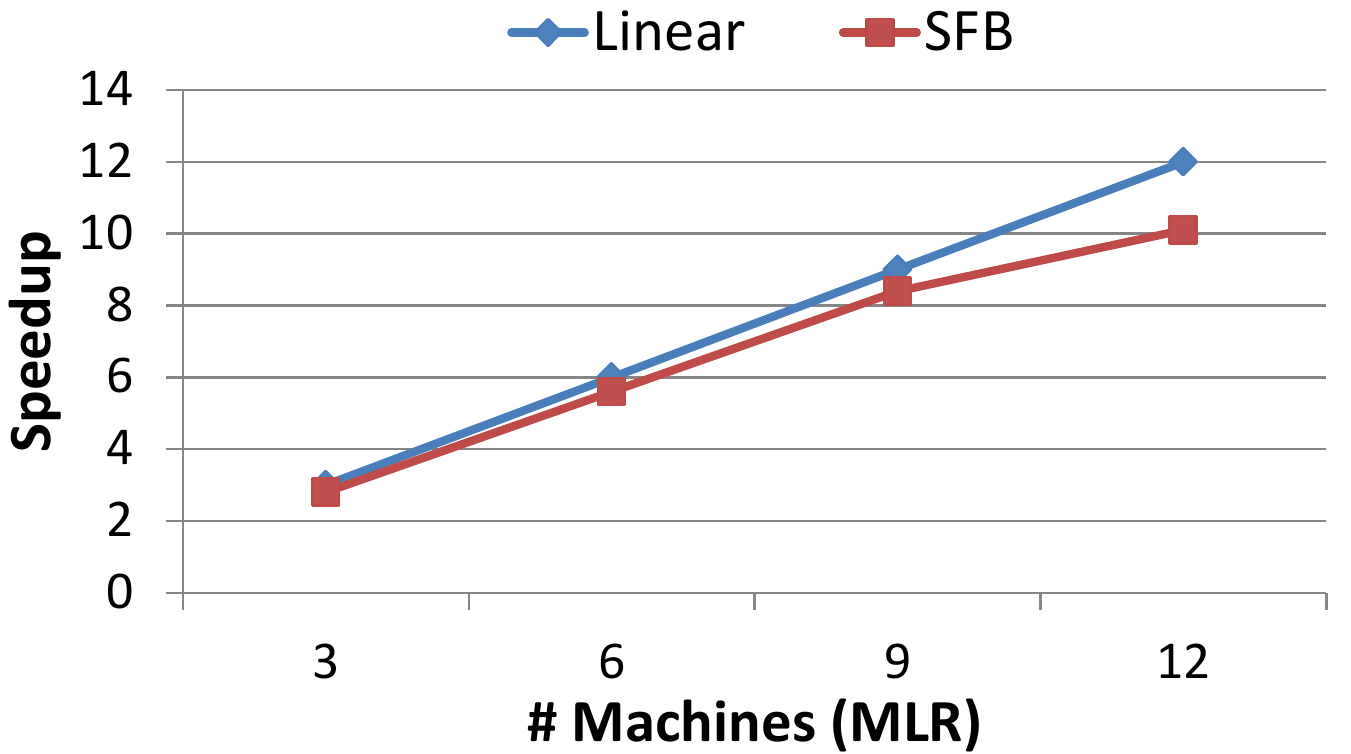}
\includegraphics[width=0.5\columnwidth]{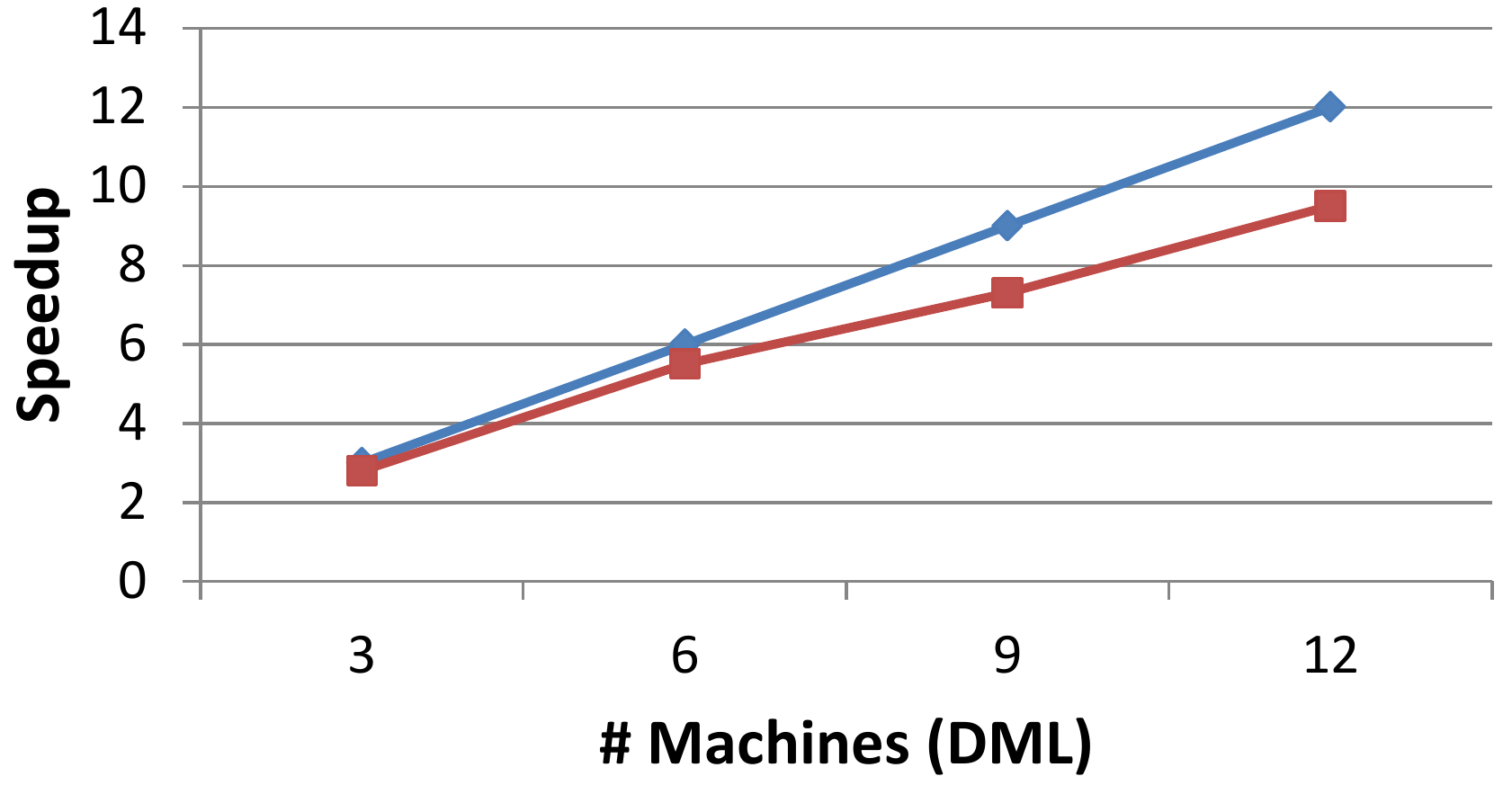}
\includegraphics[width=0.5\columnwidth]{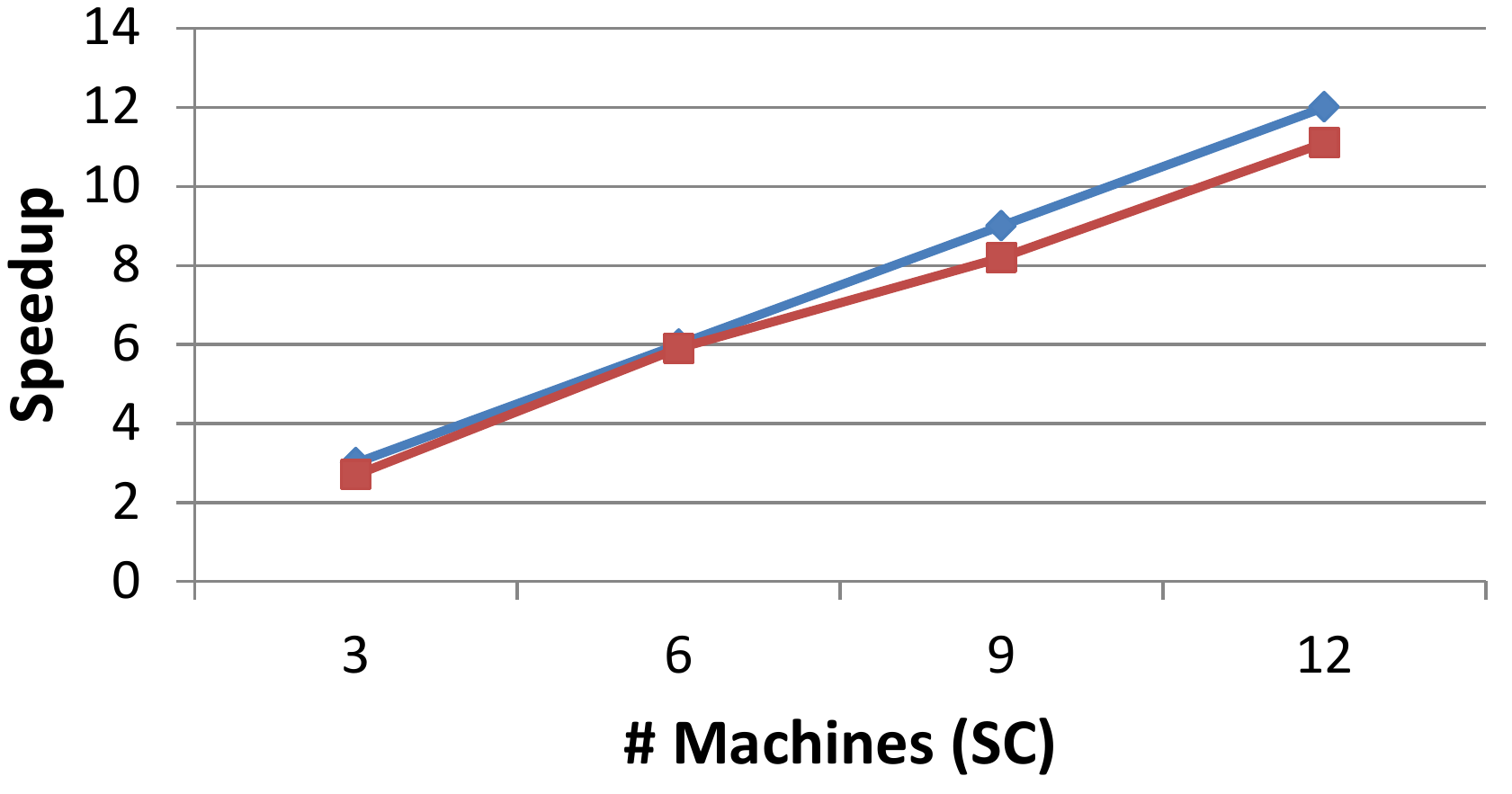}
\includegraphics[width=0.5\columnwidth]{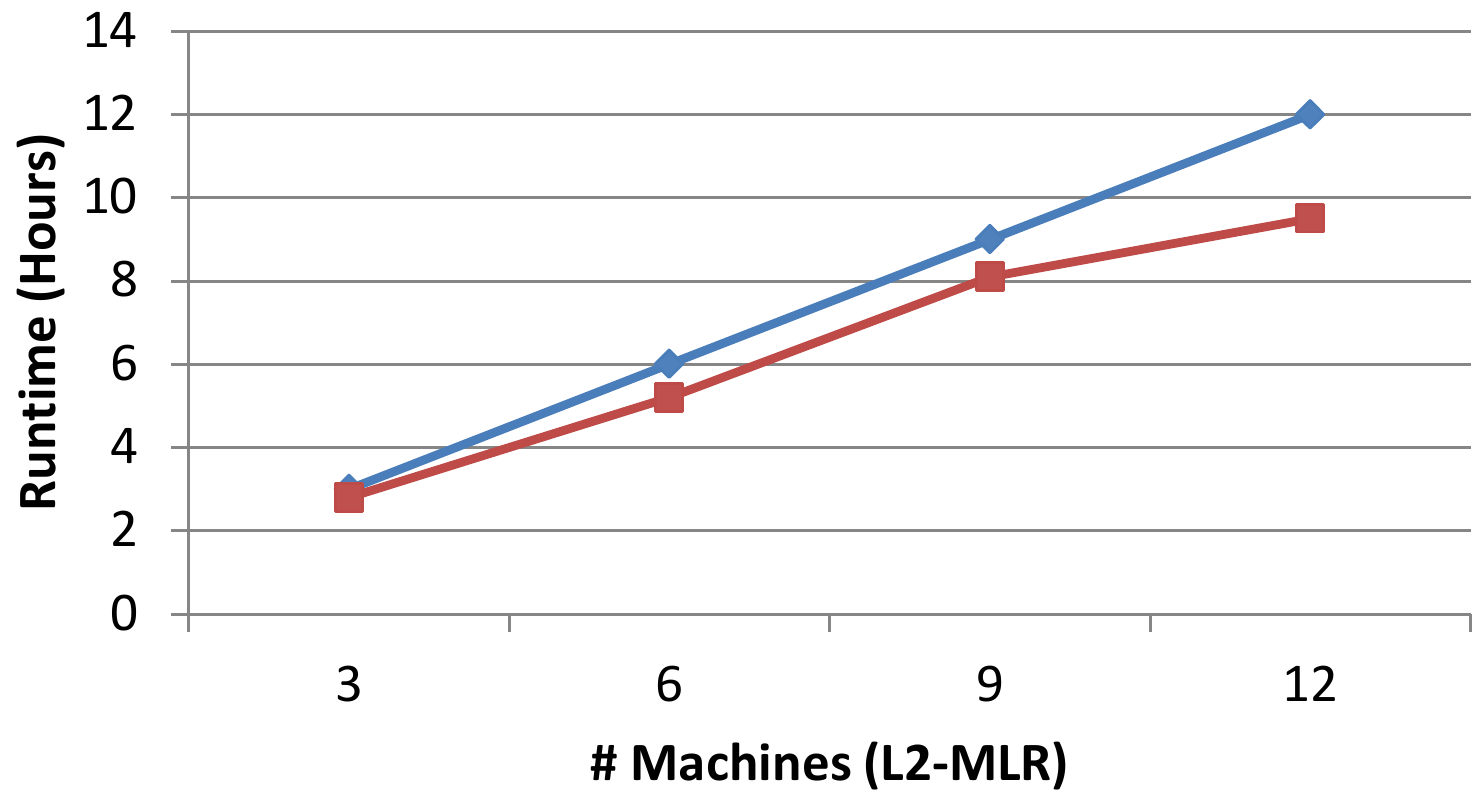}
%\vspace{-0.1in}
\caption{SFB scalability with varying machines, for MLR, DML, SC, L2-MLR (left to right), under SSP (staleness=20).}
\label{fig:exp_scalability_ssp}
\end{center}
\vspace{-0.2in}
\end{figure*}

\begin{figure*}[t]
\begin{center}
\includegraphics[width=0.5\columnwidth]{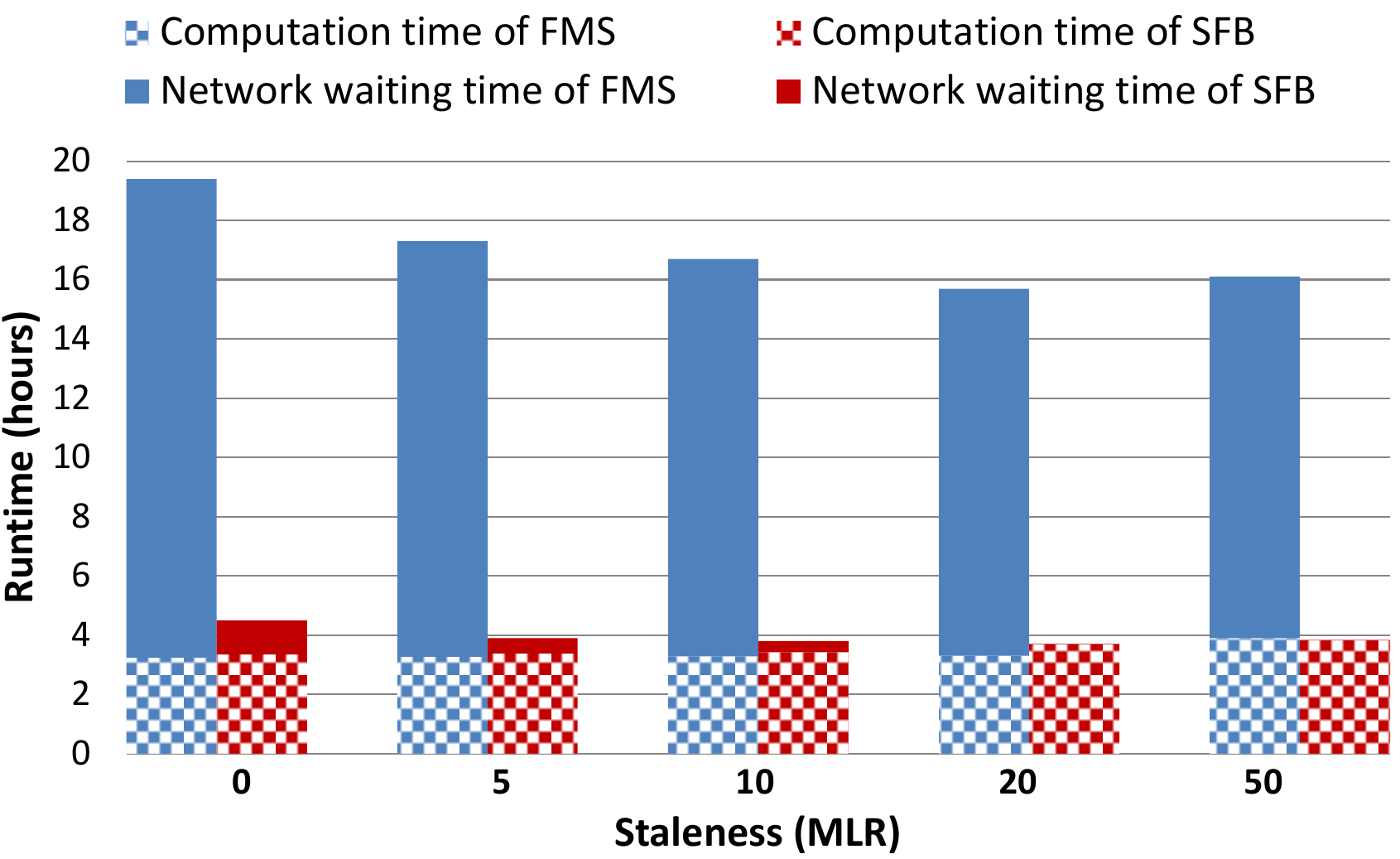}
\includegraphics[width=0.5\columnwidth]{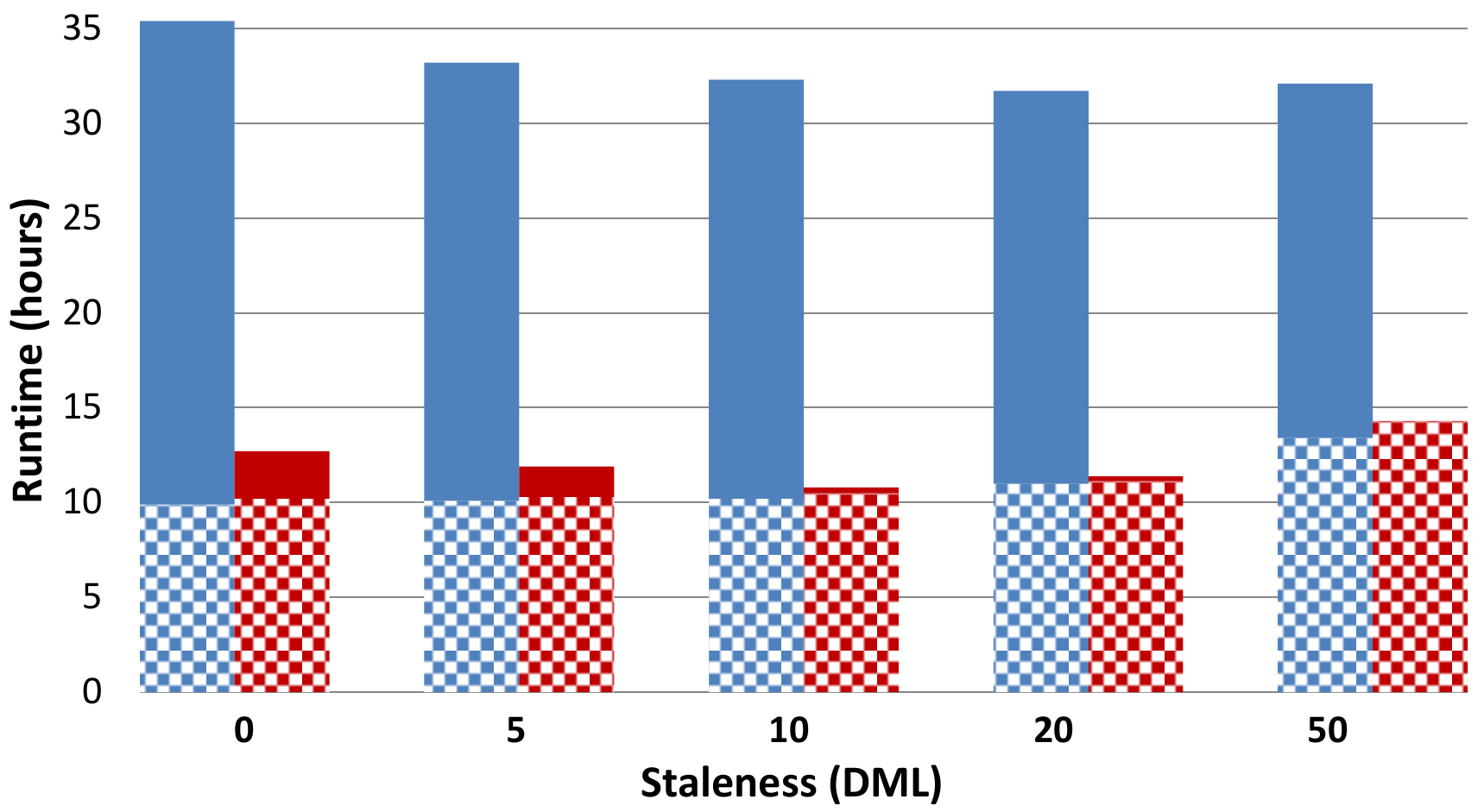}
\includegraphics[width=0.5\columnwidth]{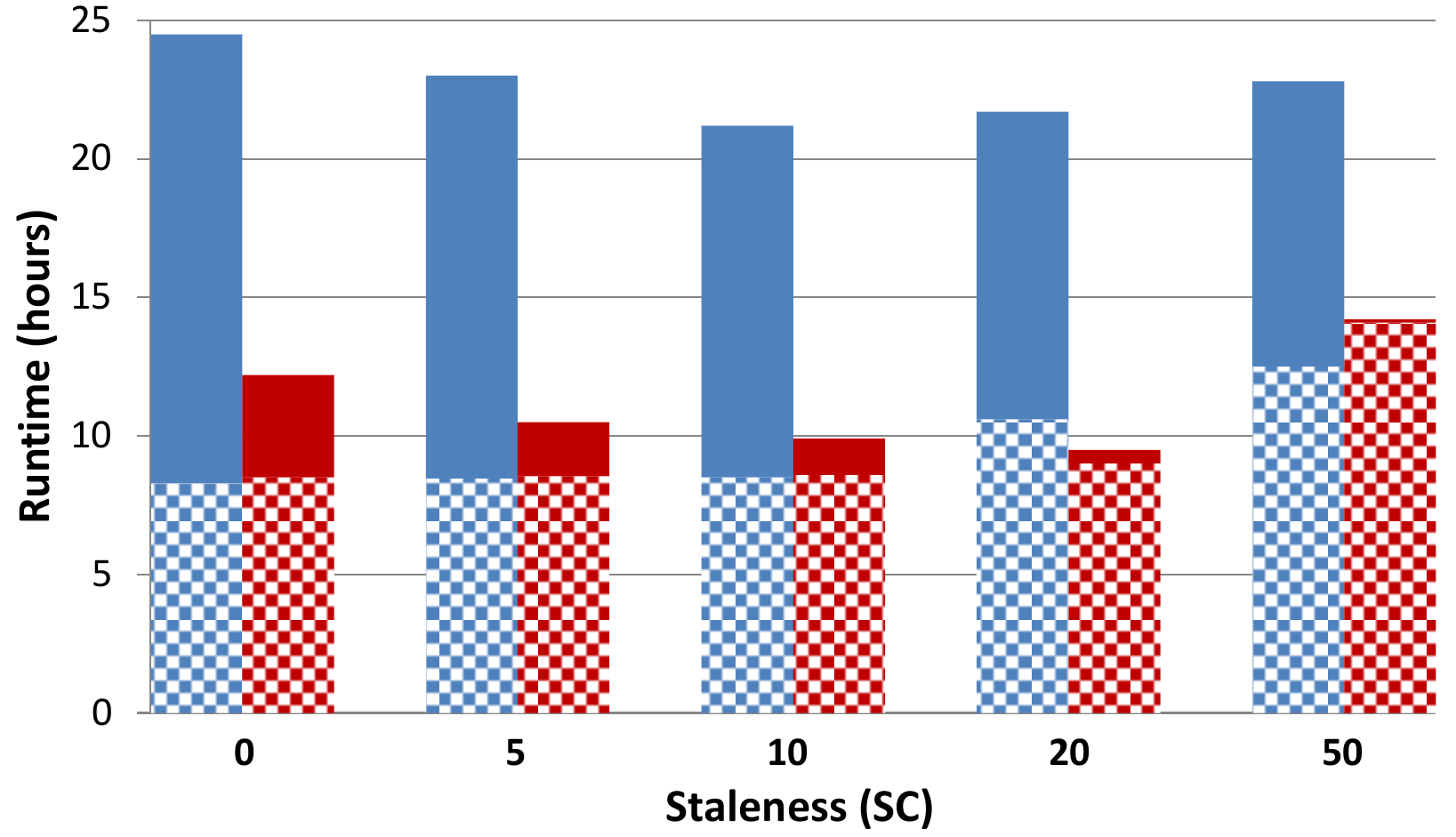}
\includegraphics[width=0.5\columnwidth]{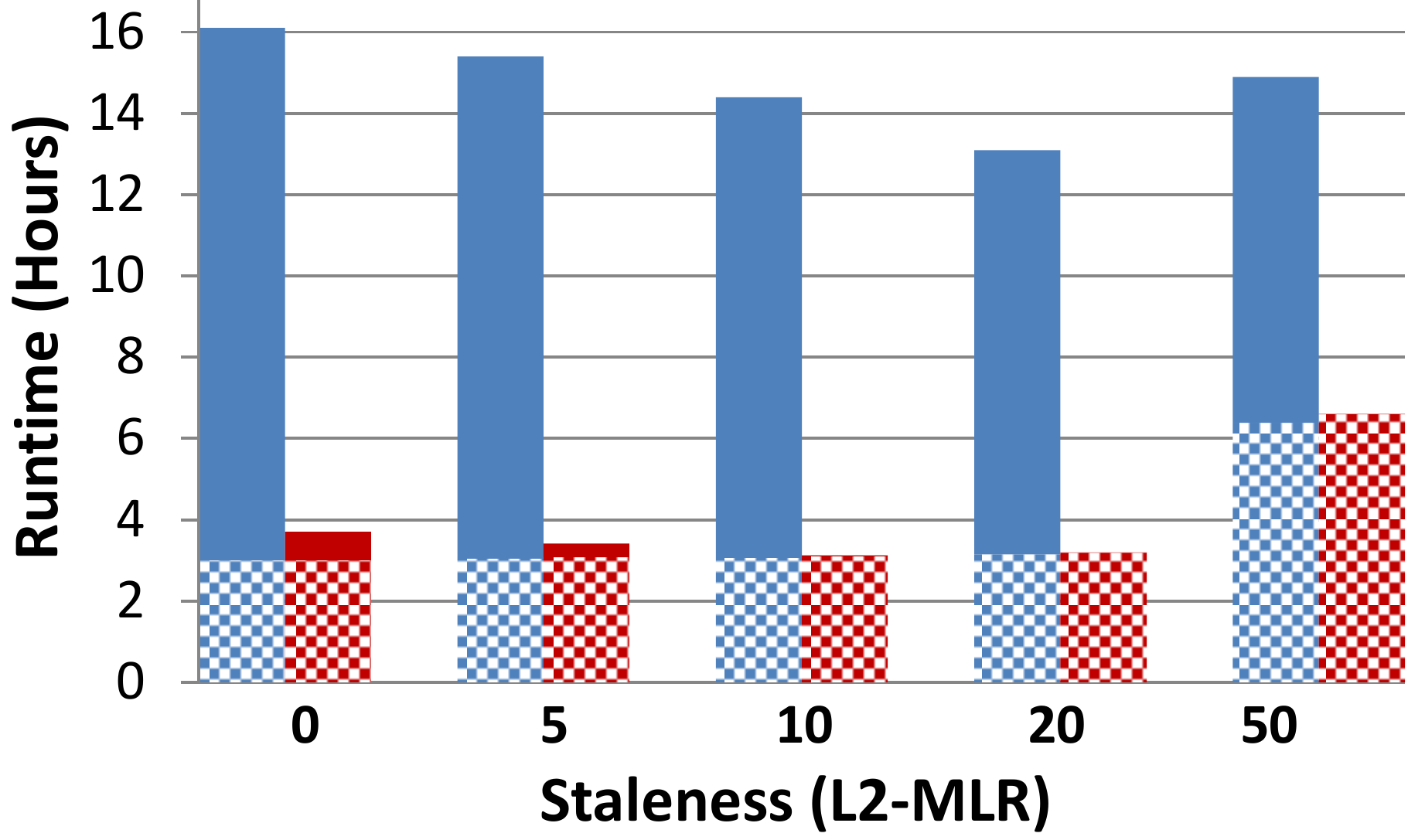}
%\vspace{-0.1in}
\caption{Computation vs network waiting time for MLR, DML, SC, L2-MLR (left to right).}
\label{fig:exp_timebreak}
\end{center}
\vspace{-0.2in}
\end{figure*}

We used two datasets for our experiments: (1) ImageNet \cite{deng2009imagenet} ILSFRC2012 dataset, which contains 1.2 million images from 1000 categories; the images are represented with LLC features \cite{wang2010locality}, whose dimensionality is 172k. (2) Wikipedia \cite{partalas2015lshtc} dataset, which contains 2.4 million documents from 325k categories; documents are represented with tf-idf, with a dimensionality of 20k.
We ran MLR, DML, SC, L2-MLR on the Wikipedia, ImageNet, ImageNet, Wikipedia datasets respectively, and the parameter matrices contained up to 6.5b, 8.6b, 8.6b, 6.5b entries respectively (the largest latent dimension for DML and largest dictionary size for SC were both 50k).
The tradeoff parameters in SC and L2-MLR were set to 0.001 and 0.1.
%For all methods, we tuned minibatch size in the range $K\in \{1,10,100,1000\}$; the best value was usually $K=10$ or 100.
We tuned the minibatch size, and found that $K=100$ was near-ideal for all experiments.
All experiments used the same constant learning rate (tuned in the range $[10^{-5},1]$). In all experiments except those in Section \ref{sec:exp:hsb}, we adopted a full broadcasting scheme. %$\{10^{-5},10^{-4},10^{-3},10^{-2},10^{-1},10^{0}\}$).

\subsection{Convergence Speed and Quality}

Figure \ref{fig:exp_runtime} shows the time taken to reach a fixed objective value, for different model sizes, using BSP consistency. Figure \ref{fig:exp_runtime_ssp} shows the results under SSP with staleness=20. 
%When the model size is small (e.g., 30K classes in MLR), SFB and FMS are comparable. As model size increases, SFB converges much faster than FMS (e.g., 4.5 times faster on MLR and 2.9 times faster on DML). The efficiency of SFB is attributed to its low communication cost, which grows linearly with the matrix dimensions whereas FMS bears a communication cost which is quadratic to matrix dimensions.  
%As the model size increases, communication becomes the bottleneck in FMS and causes prolonged network waiting time and considerable parameter synchronization delays. 
SFB converges faster than FMS, as well as Spark v1.3.1\footnote{Spark is about 2x slower than PS \cite{ho2013more,li2014scaling} based C++ implementation of FMS, due to JVM and RDD overheads.}. This is because SFB has lower communication costs, hence a greater proportion of running time gets spent on computation rather than network waiting.
%note that SFB and FMS iterations generate the exact same parameter matrix (under BSP).
This is shown in Figure \ref{fig:iter_qtt_qlt}, which plots data samples processed per second\footnote{We use samples per second instead of iterations, so different minibatch sizes can be compared.} (throughput) and algorithm progress per sample for MLR, under BSP consistency and varying minibatch sizes. The middle graph shows that SFB processes far more samples per second than FMS,
%(because SFs are much smaller than full matrices),
while the rightmost graph shows that SFB and FMS produce exactly the same algorithm progress per sample under BSP. For this experiment, minibatch sizes between $K=10$ and 100 performed the best as indicated by the leftmost graph.
%(with SFB being a little worse under SSP, due to the peer-to-peer architecture). Similar results were observed under SSP staleness 5, 10, 50 (graphs not shown).
%We attribute this to SFB's communication efficiency. SFB performs parameter synchronization by transmitting vectors, which can be done rapidly. In contrast, FMS transmits matrices, which takes much longer time. This demonstrates that SFB can greatly improve system efficiency without (or with minimal) sacrifice of convergence quality. 
%The other applications also exhibit similar behavior (graphs not shown). 
%Since having exactly the same mathematical update and using the BSP model and the same minibatch in each iteration, SFB and FMS have the same parameter matrix at the end of each iteration. However, due to the low communication cost, SFB can run more iterations than FMS over a fixed time interval, hence converges faster.
We point out that larger model sizes should further improve SFB's advantage over FMS, because SFB has linear communications cost in the matrix dimensions, whereas FMS has quadratic costs.
%As model size increases, the speedup gain of SFB over FMS becomes larger. This is because the communication cost of SFB is linear to matrix dimensions whereas FMS bears a cost quadratic to matrix dimensions. 
Under a large model size (e.g., 325K classes in MLR), the communication cost becomes the bottleneck in FMS and causes prolonged network waiting time and considerable parameter synchronization delays, while the cost is moderate in SFB.

We also compared the iteration throughput and quality of SFB and FMS under the SSP consistency model. Figure \ref{fig:iter_qtt_qlt_bsp_ssp} shows the iteration throughput (left) and iteration quality (right) for MLR, under SSP (staleness=20). The minibatch size was set to 100 for both SFB and FMS. As can be seen from the right graph, SFB has a slightly worse iteration quality than FMS. The reason we conjecture is the centralized architecture of FMS is more robust and stable than the decentralized architecture of SFB. On the other hand, the iteration throughput of SFB is much higher than FMS as shown in the left graph. Overall, SFB outperforms FMS in total convergence time.

\subsection{Scalability}
In all experiments that follow, we set the number of (L2)-MLR classes, DML latent dimension, SC dictionary size to 325k, 50k, 50k respectively.
Figure \ref{fig:exp_scalability} shows SFB scalability with varying machines under BSP, for MLR, DML, SC, L2-MLR.
Figure \ref{fig:exp_scalability_ssp} shows how SFB scales with machine count, under SSP with staleness=20. 
%These settings of model size are used throughout the following experiments. 
%and fixed model sizes (325K classes in MLR and L2-MLR, 50K latent dimension in DML, 50K dictionary size in SC). 
In general, we observed close to linear (ideal) speedup, with a slight drop at 12 machines. %Future work will focus on reducing peer-to-peer broadcast costs (e.g. via Halton sequences), in order to maintain linear scalability with more machines. 

\subsection{Computation Time vs Network Waiting Time}
Figure \ref{fig:exp_timebreak} shows the {\it total} computation and network time required for SFB and FMS to converge, across a range of SSP staleness values\footnote{The Spark implementation does not easily permit this time breakdown, so we omit it.} --- in general,
%On each worker, network waiting is incurred when the difference in iteration number between this worker and a remote worker is greater than the staleness value.
higher communication cost and lower staleness induce more network waiting.
In the figure, the horizontal axis corresponds to different staleness values. The blue solid bar, blue checker board, red solid bar and red checker board correspond to the network waiting time of FMS, computation time of FMS, network waiting time of SFB and computation time of SFB respectively.
For all staleness values, SFB requires far less network waiting (because SFs are much smaller than full matrices in FMS).
%Due to the fast communication of SFB, each worker can see other workers' updates (SFs) timely without too much network waiting. In contrast, the transmission of matrices in FMS is inefficient, which incurs much longer network waiting.
Computation time for SFB is slightly longer than FMS because (1) update matrices must be reconstructed on each SFB worker, and (2) SFB requires a few more iterations for convergence, because peer-to-peer communication causes a slightly more parameter inconsistency under staleness.
%(2) the slightly degraded iteration quality due to the decentralized architecture.
Overall, the SFB reduction in network waiting time remains far greater than the added computation time, and outperforms FMS in total time. 
%For both FMS and SFB, the shortest convergence times are achieved at moderate staleness values, confirming the importance of bounded-asynchronous communication.
%Because the reduction of network waiting time is much more significant than the increase of computation time, SFB reduces the total convergence time compared with FMS.

Another observation is that increasing staleness value can reduce the network waiting time for both FMS and SFB, thus allows more iterations to be performed in a given time interval. This is because a larger staleness gives each worker more flexibility to proceed at its own pace without waiting for others. On the other hand, as staleness increases, the computation time grows. If the staleness is too large (e.g., 50), the total convergence time is prolonged due to the rapid increasing of computation time. This is due to that the parameter copies on different workers bear a higher risk to be out of synchronization (thus inconsistent) under a larger staleness, which hurts the convergence quality of each iteration and requires more iterations to achieve the same objective value. The best tradeoff happens when staleness is around 10-20 in our experiments.

\subsection{Communication Cost}
\begin{figure*}[t]
\begin{center}
\includegraphics[width=0.5\columnwidth]{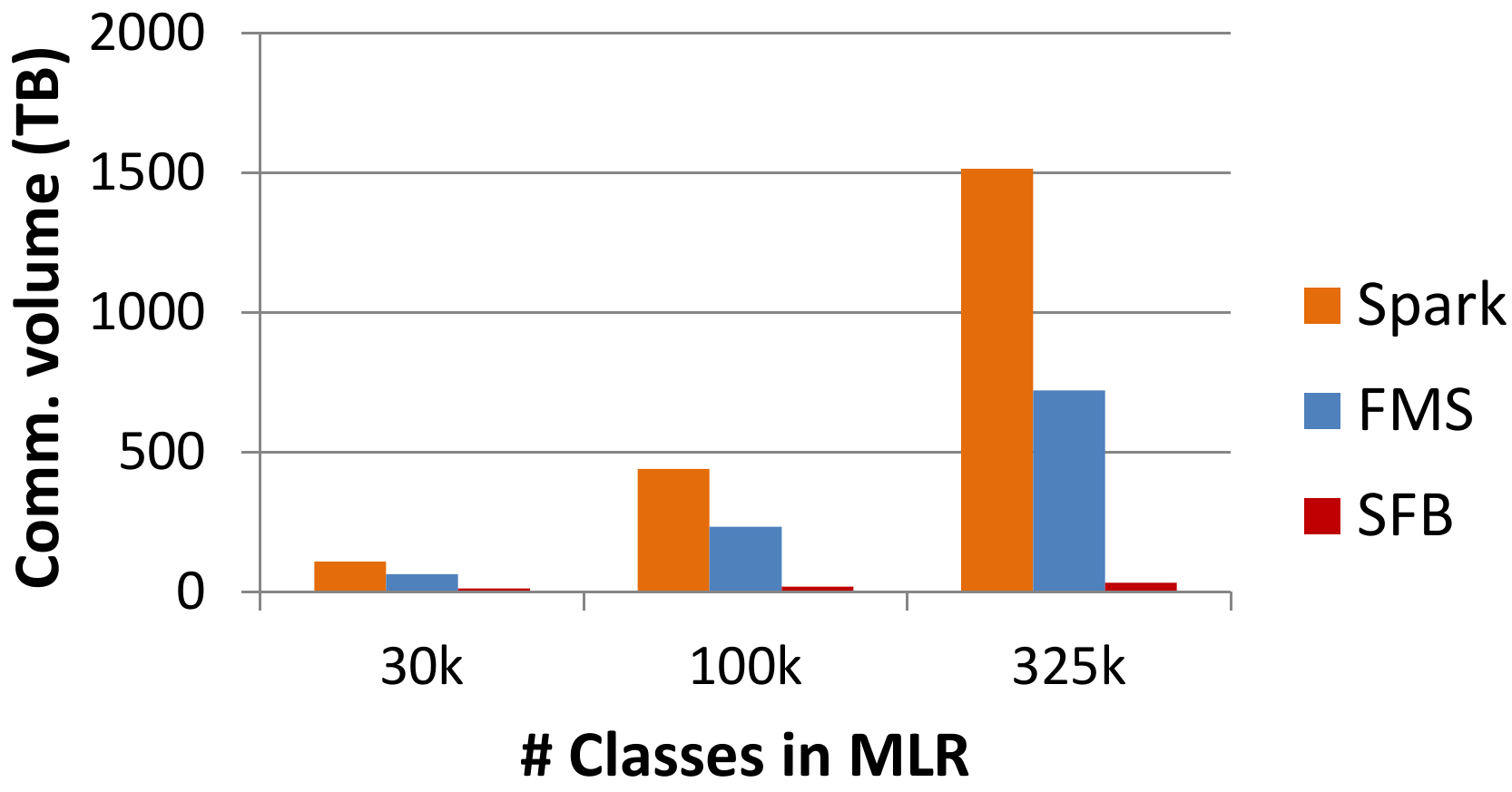}
\includegraphics[width=0.5\columnwidth]{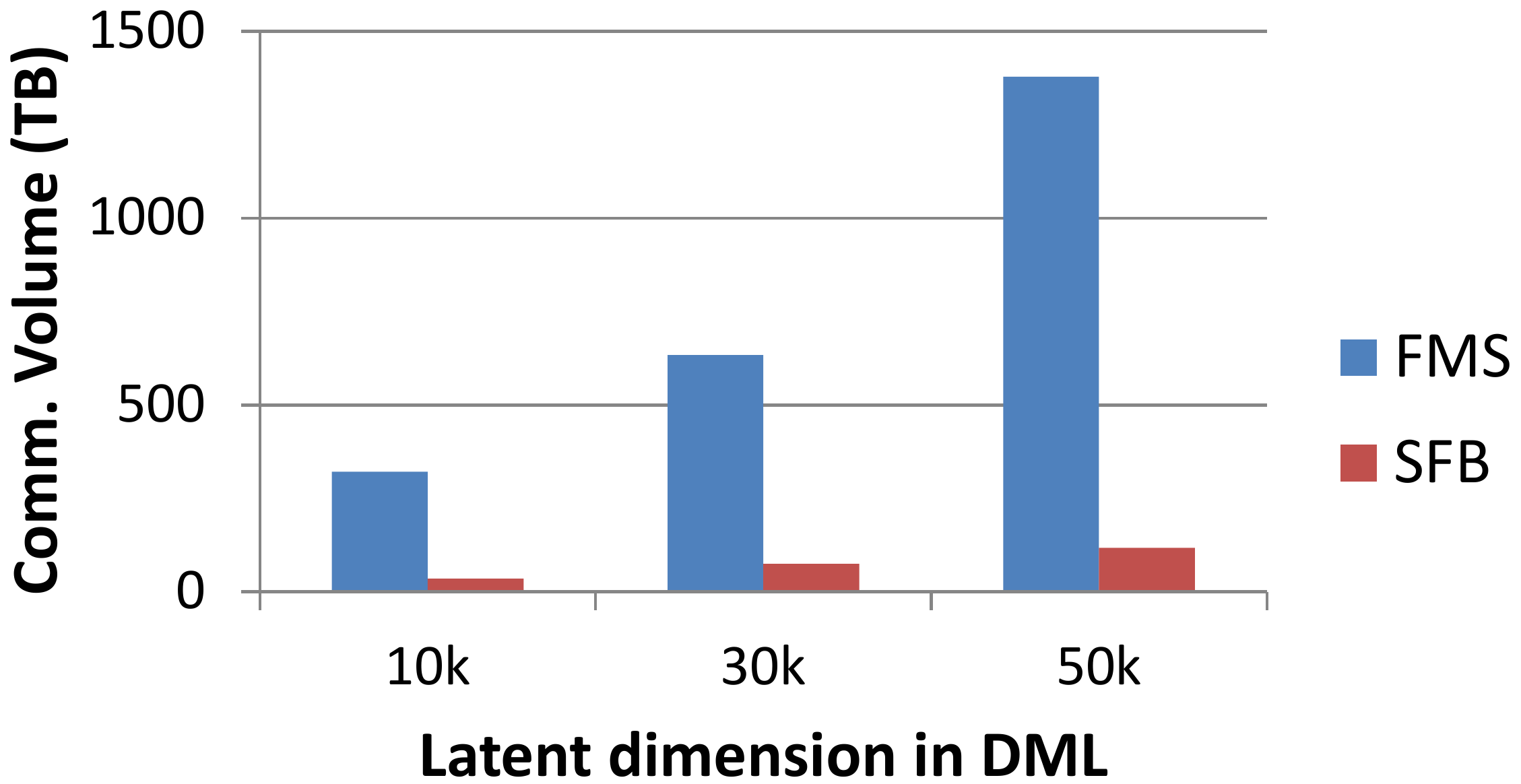}
\includegraphics[width=0.5\columnwidth]{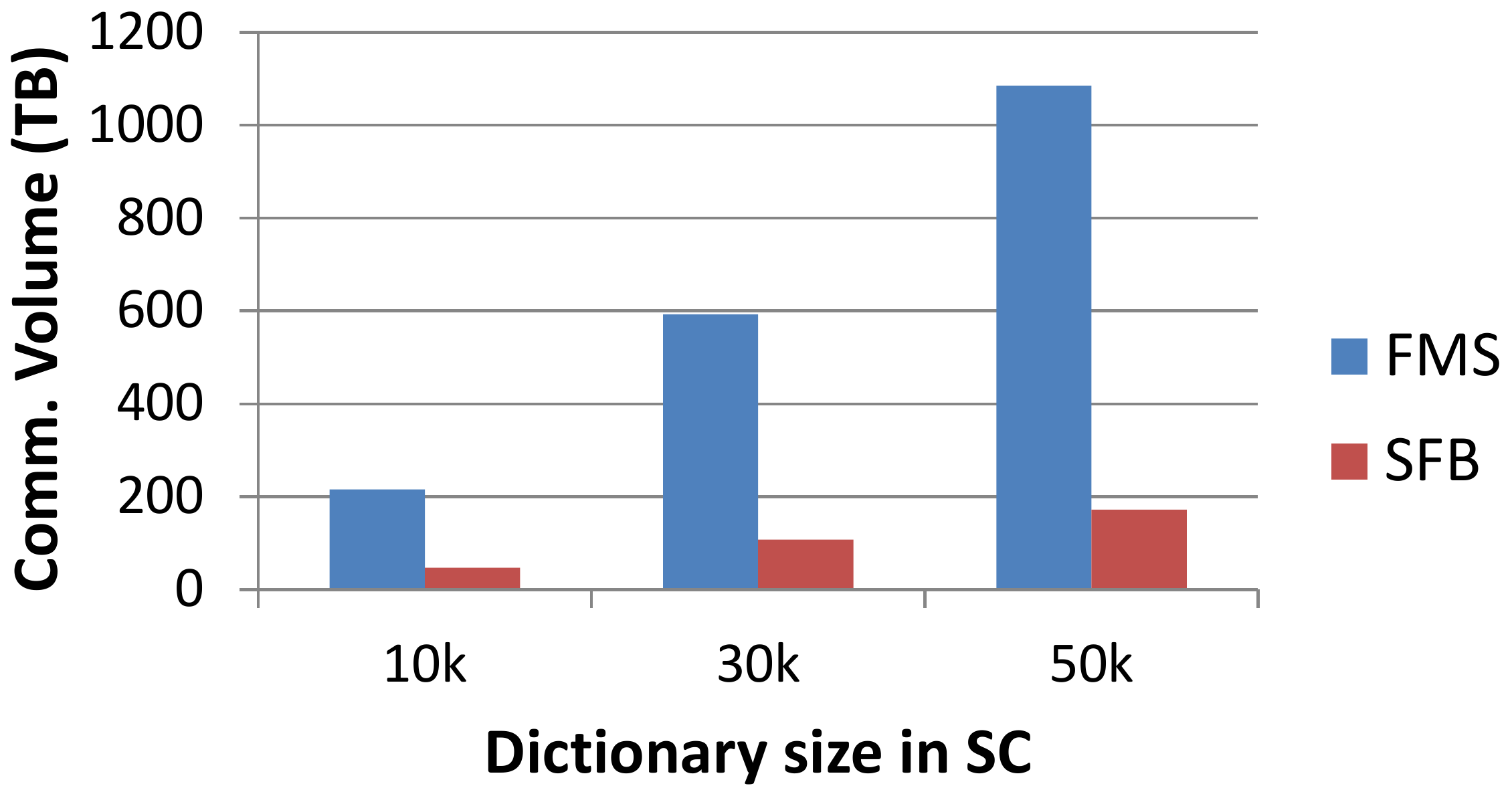}
\includegraphics[width=0.5\columnwidth]{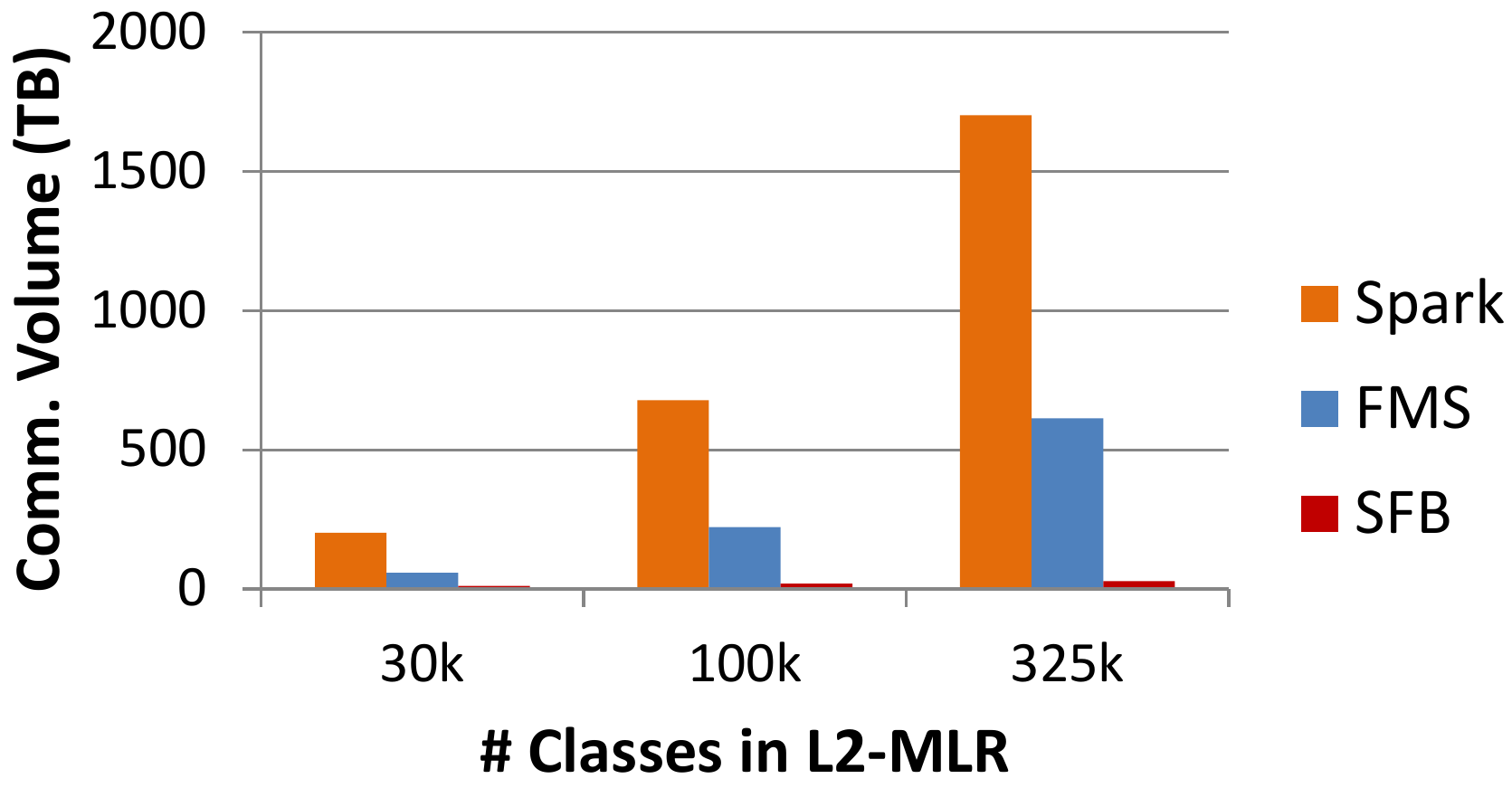}
\vspace{-0.1in}
\caption{Communication volume for MLR, DML, SC, L2-MLR (left to right) under BSP.}
\label{fig:exp_comm}
\end{center}
\vspace{-0.2in}
\end{figure*}

Figure \ref{fig:exp_comm} shows the communication volume of four models under BSP. As shown in the figure, the communication volume of SFB is significantly lower than FMS and Spark. Under the BSP consistency model, SFB and FMS share the same iteration quality, hence need the same number of iterations to converge. Within each iteration, SFB communicates vectors while FMS transmits matrices. As a result, the communication volume of SFB is much lower than FMS.

\begin{figure}[t]
\begin{center}
\includegraphics[width=0.49\columnwidth]{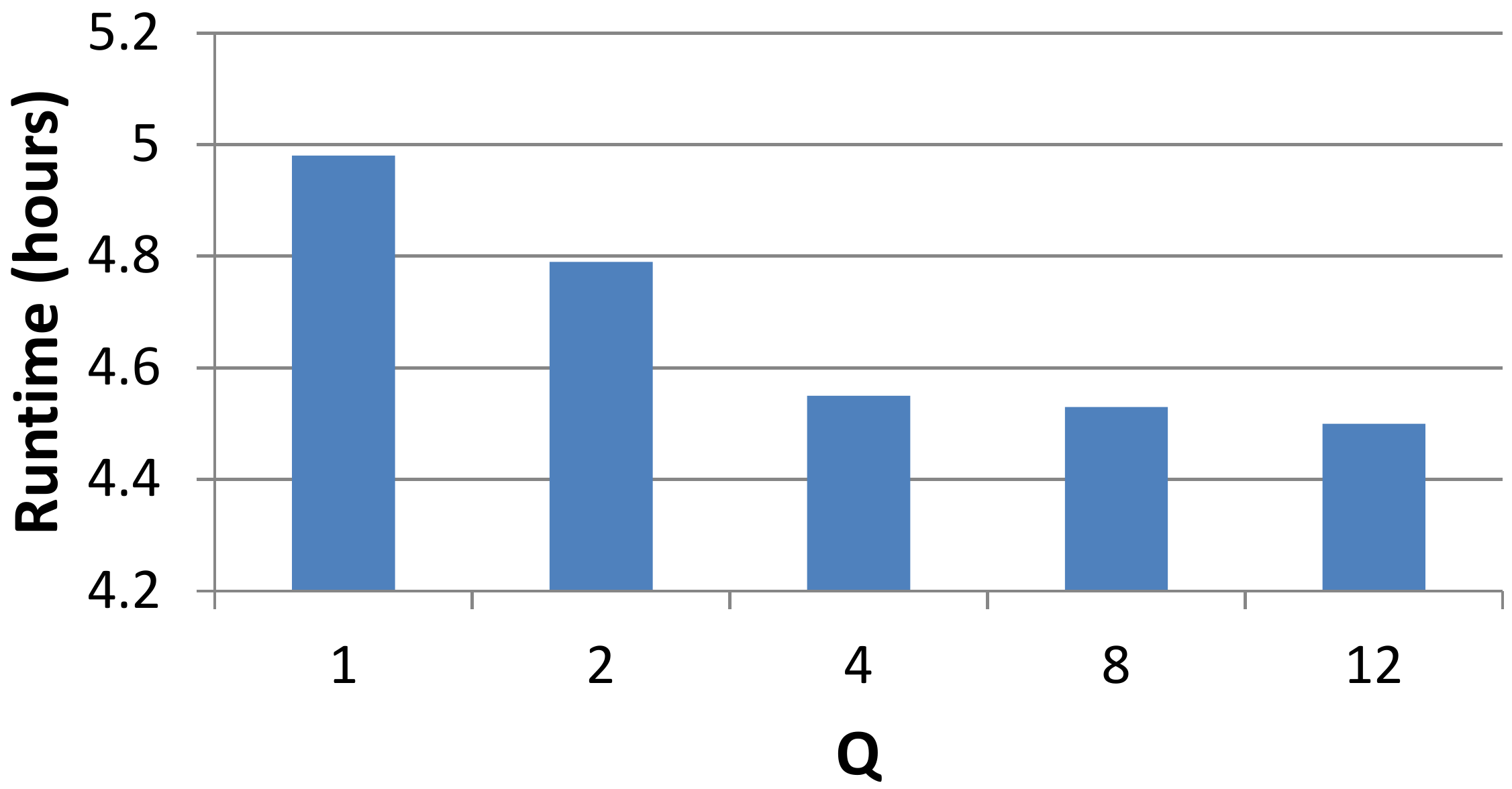}
\includegraphics[width=0.49\columnwidth]{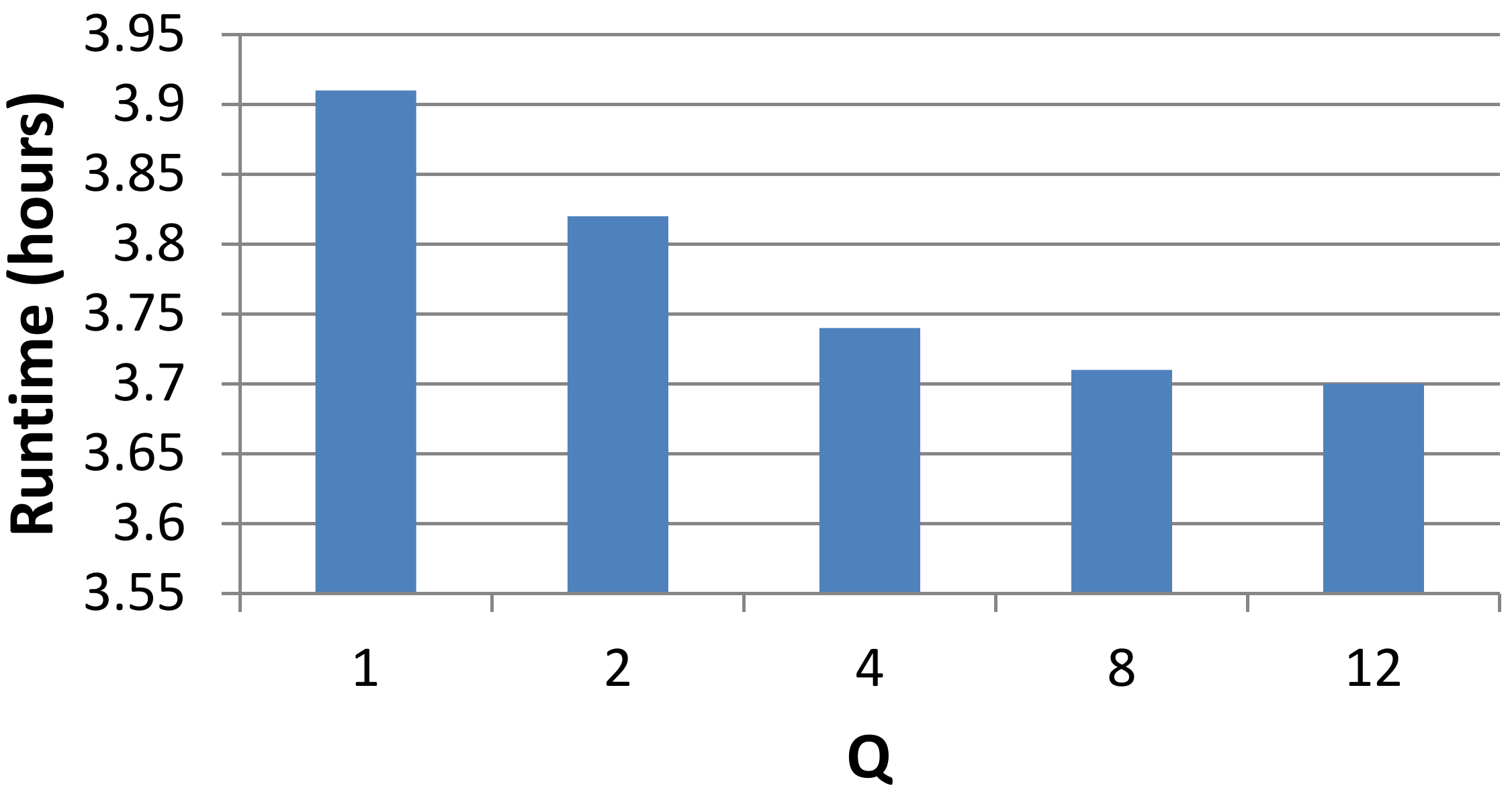}
\vspace{-0.1in}
\caption{Convergence time versus Q in Halton sequence broadcasting for MLR (left) and L2-MLR (right), under BSP.}
\label{fig:halton_bsp}
\end{center}
\vspace{-0.2in}
\end{figure}

\begin{figure}[t]
\begin{center}
\includegraphics[width=0.49\columnwidth]{svb/halton_l2mlr}
\includegraphics[width=0.49\columnwidth]{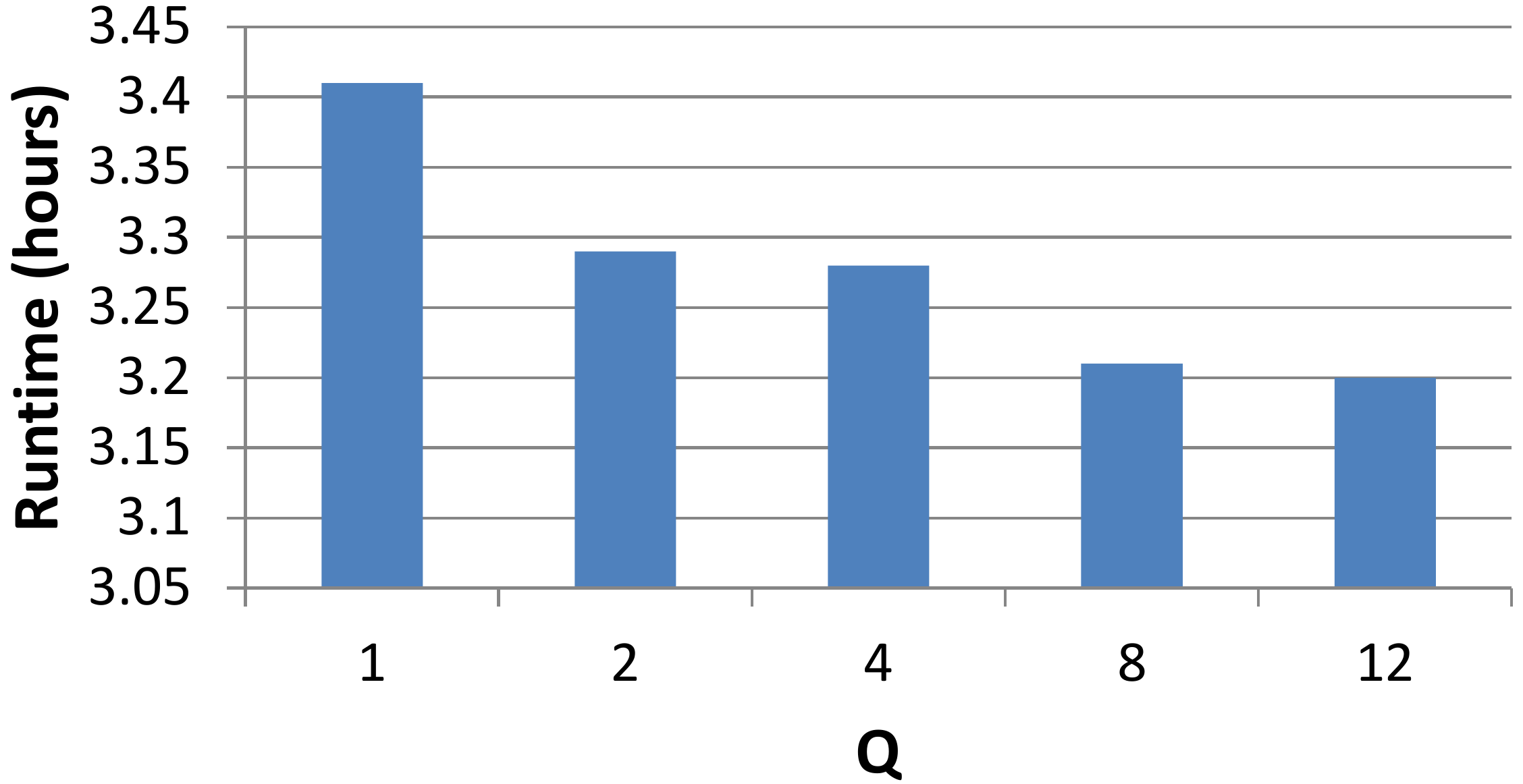}
\vspace{-0.1in}
\caption{Convergence time versus Q in Halton sequence broadcasting for MLR (left) and L2-MLR (right), under SSP (staleness=20).}
\label{fig:halton_ssp}
\end{center}
\vspace{-0.2in}
\end{figure}

\subsection{Halton Sequence Broadcasting}
\label{sec:exp:hsb}
We studied how the parameter $Q$ in Halton sequence broadcasting (HSB) affects the convergence speed of SFB. Figure \ref{fig:halton_bsp} and \ref{fig:halton_ssp} show the convergence time of MLR and L2-MLR versus varying Q, under BSP and SSP (staleness=20) respectively. As observed in these two figures, a smaller $Q$ incurs longer convergence time. This is because a smaller $Q$ is more likely to cause the parameter copies on different workers to be out of synchronization and degrade iteration quality. However, as long as $Q$ is not too small, the convergence speed of HSB is comparable with a full broadcasting scheme. As shown in the figures, for $Q\geq 4\approx \log_2 12$, the convergence time of HSB is very close to full broadcasting (where $Q=12$). This demonstrates that using HSB, we can reduce the communication cost from $O(P^2)$ to $O(PQ)\approx O(P\log P)$ with slight sacrifice of the convergence speed.
%\vspace{-0.1in}
%\paragraph{Scalability}
\iffalse
\paragraph{Iteration Throughput vs Quality}
Figure \ref{fig:iter_qtt_qlt_bsp_ssp} shows the iteration throughput (left) and iteration quality (right) for MLR, under SSP (staleness=20). The minibatch size was set to 100 for both SFB and FMS. As can be seen from the rightmost graph, SFB has a slightly worse iteration quality than FMS. The reason we conjecture is the centralized architecture of FMS is more robust and stable than the decentralized architecture of SFB. On the other hand, the iteration throughput of SFB is much higher than FMS as shown in the leftmost graph. 

\begin{figure}[t]
\begin{center}
%\includegraphics[width=0.24\columnwidth]{svb/iqtt_mlr_bsp}
%\includegraphics[width=0.24\columnwidth]{svb/iq_mlr_bsp}
\includegraphics[width=0.3\columnwidth]{svb/iqtt_mlr_ssp20}
\includegraphics[width=0.3\columnwidth]{svb/iq_mlr_ssp20}
\caption{MLR iteration throughput (left) and iteration quality (right) for MLR under SSP (staleness=20).}
\label{fig:iter_qtt_qlt_bsp_ssp}
\end{center}
\end{figure}
\fi

\section{Related Works}

A number of system and algorithmic solutions have been proposed to reduce communication cost in distributed ML. On the system side, 
\cite{dean2012large} proposed to reduce communication overhead by reducing the frequency of parameter/gradient exchanges between workers and the central server. 
%This may cause considerable staleness of the parameters and hence slow down the convergence of the algorithm. 
%Project Adam \cite{chilimbi2014project} leverages the fact that the gradient of weight matrix in neural network can be computed as the outer product of two vectors and proposed to send the two vectors from workers to the center server, instead of sending the whole gradient matrix. However, from the server to workers, the weight matrices need to be sent, which still incurs substantial communication overhead. In SFB, we adopt a decentralized, peer to peer architecture, which can completely avoid transmitting matrices.
\cite{li2014scaling} used filters to select part of ``important'' parameters/updates for transmission to reduce the number of data entries to be communicated. 
%In practice, how to properly measure ``importance'' can be subtle and difficult.
%While effective, these filters normally are unable to achieve an one-order magnitude (from quadratic to linear) reduction of communication cost as SFB is. 
On the algorithm side, \cite{tsianos2012communication} and \cite{yang2013trading} studied the tradeoffs between communication and computation in distributed dual averaging and distributed stochastic dual coordinate ascent respectively.  \cite{shamir2013communication} proposed an approximate Newton-type method to achieve communication efficiency in distributed optimization. SFB is orthogonal to these existing approaches and be potentially combined with them to further reduce communication cost.

Peer-to-peer, decentralized architectures have been investigated in other distributed ML frameworks \cite{bhaduri2008distributed,das2010local,ormandi2013gossip,li2015malt}. Our SFBcaster system also adopt such an architecture, but with the specific purpose of supporting the SFB computation model, which is not explored by existing peer-to-peer ML frameworks.

\section{Conclusions and Future Works}
In this paper, we identify the sufficient factor property of a large set of matrix-parametrized models: when these models are optimized with stochastic gradient descent or stochastic dual coordinate ascent, the update matrices are of low-rank. Leveraging this property, we propose a sufficient factor broadcasting computation model to efficiently handle the learning of these models with low communication cost. We analyze the cost and convergence property of SFB and provide an efficient implementation and empirical evaluations.

For very large models, the size of the local parameter matrix $\mb{W}$ may exceed each machine's memory capacity --- to address this issue, we would like to investigate partitioning $\mb{W}$ over a small number of nearby machines, or using out-of-core (disk-based) storage to hold $\mb{W}$ in future work.
%\qirong{Need to address size of local matrices here --- what if there is not enough RAM per machine? Recommend proposing out-of-core solution as future work: store the full local matrix on disk, to be updated in an intelligent streaming fashion as SFs come in.}

Finally, a promising extension to the SF idea is to re-parameterize the model $\mb{W}$ completely in terms of SFs, rather than just the updates. For example, if we initialize the parameter matrix $\mb{W}$ to be of low rank $R$, i.e., $\mb{W}^0=\sum_{j=1}^{R}\mb{u}_j\mb{v}_j^\top$, after $I$ iterations (updates), $\mb{W}^{I}=\sum_{j=1}^{R+I}\mb{u}_j\mb{v}_j^\top$. Leveraging this fact, for SGD without proximal operation and SDCA where $h(\cdot)$ is a $\ell_2$ regularizer, we can re-parametrize $\mb{W}^{I}$ using a set of SFs $\{(\mb{u}_j,\mb{v}_j)\}_{j=1}^{R+I}$, rather than maintaining $\mb{W}$ explicitly. This re-parametrization can possibly reduce both computation and storage cost, which we will investigate in the future. 

{
%\small
\bibliographystyle{abbrv}
\bibliography{vldb_sample}
}
\begin{appendix}
\section{Proof of Theorem 1}
\begin{proof}

\newcommand{\EE}{\mathbb{E}}
\newcommand{\Fcal}{\mathcal{F}}

Let $\Fcal^c := \sigma\{I_p^\tau: p=1,\ldots, P, \tau = 1, \ldots, c \}$ be the filtration generated by the random samplings $I_{p}^\tau$ up to iteration counter $c$, \ie, the information up to iteration $c$. Note that for all $p$ and $c$, $\mb{W}_p^c$ and $\mb{W}^c$ are $\Fcal^{c-1}$ measurable (since $\tau_p^q(c) \leq c-1$ by assumption), and $I_p^c$ is independent of $\Fcal^{c-1}$. Recall that the partial update generated by machine $p$ at its $c$-th iteration is
$$U_p(\mb{W}_p^c, I_p^c) = -\eta_c |S_p| \sum_{j\in I_p^c}\nabla f_{j}(\mathbf{W}_p^c)$$
Then it holds that 
$$U_p(\mathbf{W}_p^c) = \EE[ U_p(\mb{W}_p^c, I_p^c)  | \Fcal^{c-1}] = -\eta_c\nabla F_p(\mathbf{W}_p^c)$$ 
(Note that we have suppressed the dependence of $U_p$ on the iteration counter $c$.)

Then, we have
\begin{equation}
\begin{array}{l}
\EE \left[\sum_{p=1}^{P} U_p(\mathbf{W}_p^c, I_p^c) ~|~ \Fcal^{c-1} \right] = \sum_{p=1}^{P} \EE [ U_p(\mathbf{W}_p^c, I_p^c) ~|~ \Fcal^{c-1} ]\\
 = \sum_{p=1}^{P} U_p(\mathbf{W}_p^c)
 \end{array}
\end{equation}
Similarly we have
\begin{equation}
\begin{array}{l}
\EE\left[\big\|\sum_{p=1}^{P} U_p(\mathbf{W}_p^c, I_p^c)\big\|_2^2 ~|~ \Fcal^{c-1} \right] \\= \sum_{p,q=1}^{P}\EE[\langle U_p(\mathbf{W}_p^c, I_p^c), U_q(\mathbf{W}_q^c, I_q^c) \rangle ~|~ \Fcal^{c-1} ] \\
= \sum_{p,q = 1}^{P} \langle U_p(\mathbf{W}_q^c), U_q(\mathbf{W}_q^c) \rangle \\
+ \sum_{p=1}^{P} \EE\left[ \|U_p(\mathbf{W}_p^c, I_p^c) - U_p(\mathbf{W}_p^c)\|_2^2 ~|~ \Fcal^{c-1} \right]
 \end{array}
\end{equation}
The variance term in the above equality can be bounded as
\begin{equation}
\begin{array}{l}
\sum_{p=1}^{P} \mathbb{E}\left[\|U_p(\mathbf{W}_p^c, I_p^c) - U_p(\mathbf{W}_p^c)\|_2^2 ~|~ \Fcal^{c-1} \right] \\
= \eta_c^2 \underbrace{\sum_{p=1}^{P} \mathbb{E}\left[\||S_p|\sum_{j\in I_p^c}\nabla f_j(\mathbf{W}_p^c) - \nabla F_p(\mathbf{W}_p^c)\|_2^2 ~|~ \Fcal^{c-1} \right]}_{\hat{\sigma}^2 P}\\
\le  \eta_c^2 \hat{\sigma}^2 P
 \end{array}
\end{equation}
\balance
Now use the update rule $\mathbf{W}_p^{c+1}  = \mathbf{W}_p^c + \sum_{p=1}^{P}U_p(\mathbf{W}_p^c, I_p^c)$ and  the descent lemma \cite{BertsekasTsitsiklis89}, we have
\begin{equation}
\begin{array}{l}
F(\mathbf{W}^{c+1}) -F(\mathbf{W}^c) \\
\le \langle \mathbf{W}^{c+1} - \mathbf{W}^c, \nabla F(\mathbf{W}^c) \rangle + \frac{L_F}{2}\|\mathbf{W}^{c+1} - \mathbf{W}^c\|_2^2 \\
= \langle\sum_{p=1}^{P}U_p(\mathbf{W}_p^c, I_p^c), \nabla F(\mathbf{W}^c) \rangle + \frac{L_F}{2}\|\sum_{p=1}^{P}U_p(\mathbf{W}_p^c, I_p^c)\|_2^2 
 \end{array}
\end{equation}
Then take expectation on both sides, we obtain
%Need to put \left. and \right. around the &. Otherwise it doesn't work.
\begin{equation}
\begin{array}{l}
\mathbb{E}\left[\right.  \left.F(\mathbf{W}^{c+1})  -F(\mathbf{W}^c)~|~ \Fcal^{c-1}\right] \\
\le \langle \sum_{p=1}^{P}U_p(\mathbf{W}_p^c), \nabla F(\mathbf{W}^c)\rangle + \frac{L_F \eta_c^2 \hat\sigma^2 P}{2} \\
+ \frac{L_F}{2}\|\sum_{p=1}^{P}U_p(\mathbf{W}_p^c)\|_2^2
\\
= (\frac{L_F}{2} - \eta_c^{-1})\|\sum_{p=1}^{P}U_p(\mathbf{W}_p^c)\|_2^2+ \frac{L_F \eta_c^2 \hat\sigma^2 P}{2} \\
- \eta_c^{-1}\langle \sum_{p=1}^{P}U_p(\mathbf{W}_p^c), \sum_{p=1}^{P}[U_p(\mathbf{W}^c)- U_p(\mathbf{W}_p^c)]\rangle \\
\le (\frac{L_F}{2} - \eta_c^{-1})\|\sum_{p=1}^{P}U_p(\mathbf{W}_p^c)\|_2^2+\frac{L_F \eta_c^2 \hat\sigma^2 P}{2} \\
+ \|\sum_{p=1}^{P}U_p(\mathbf{W}_p^c)\|\sum_{p=1}^{P} L_p \|\mathbf{W}^c - \mathbf{W}_p^c\| 
 \end{array}
\end{equation}
Now take expectation w.r.t all random variables, we obtain
\begin{equation}
\label{eq:tmp}
\begin{array}{l}
\EE\left[F(\mathbf{W}^{c+1}) -F(\mathbf{W}^c)\right] \le (\frac{L_F}{2} - \eta_c^{-1})\EE\left[\|\sum_{p=1}^{P}U_p(\mathbf{W}_p^c)\|_2^2\right] \\
+ \sum_{p=1}^{P} L_p\EE\left[\|\sum_{p=1}^{P}U_p(\mathbf{W}_p^c)\| \|\mathbf{W}^c - \mathbf{W}_p^c\|\right] +\frac{L_F \eta_c^2 \sigma^2 P}{2}
 \end{array}
\end{equation}

Next we proceed to bound the term $\mathbb{E}\|\sum_{p=1}^{P}U_p(\mathbf{W}_p^c)\| \|\mathbf{W}^c - \mathbf{W}_p^c\|$. We list the auxiliary update rule and the local update rule here for convenience.
\begin{equation}
\begin{array}{l}
\mathbf{W}^c = \mathbf{W}^0 + \sum_{q=1}^{P}\sum_{t=0}^{c-1}U_q(\mathbf{W}_q^t, I_q^t), \\
\mathbf{W}_p^c = \mathbf{W}^0 +  \sum_{q=1}^{P}\sum_{t=0}^{\tau_p^q(c)} U_q(\mathbf{W}_q^t, I_q^t).
 \end{array}
\end{equation}
Now subtract the above two and use the bounded delay assumption $0\le (c-1) - \tau_p^q(c) \le s$, we obtain
 \begin{equation}
\begin{array}{l}
\|\mathbf{W}^c - \mathbf{W}_p^c\| = \|\sum_{q=1}^{P}\sum_{t=\tau_p^q(c)+1}^{c-1}  U_q(\mathbf{W}_q^t, I_q^t) \|\\
\le \|\sum_{q=1}^{P}\sum_{t=c-s}^{c-1}  U_q(\mathbf{W}_q^t, I_q^t) \|+ \|\sum_{q=1}^{P} \sum_{t=c-s}^{\tau_p^q(c)} U_q(\mathbf{W}_q^t, I_q^t)\|\\
\label{eq:tmp2}
\le \sum_{t=c-s}^{c-1} \|\sum_{q=1}^{P}U_q(\mathbf{W}_q^t, I_q^t)\| + \eta_{c-s} G
 \end{array}
\end{equation}
where the last inequality follows from the facts that $\eta_c$ is strictly decreasing, and $\|\sum_{q=1}^{P} \sum_{t=c-s}^{\tau_p^q(c)} \nabla F_q(\mathbf{W}_q^t, I_q^t)\|$ is bounded by some constant $G$ since $\nabla F_q$ is continuous and all the sequences $\mathbf{W}_p^c$ are bounded. Thus by taking expectation, we obtain
\begin{equation}
\begin{array}{l}
\EE\left[\|\sum_{p=1}^{P}U_p(\mathbf{W}_p^c)\| \right.  \left. \|\mathbf{W}^c - \mathbf{W}_p^c\|\right] \\
\le \EE \left[\|\sum_{p=1}^{P}U_p(\mathbf{W}_p^c)\|\Big(\sum_{t=c-s}^{c-1} \|\sum_{q=1}^{P}U_q(\mathbf{W}_q^t, I_q^t)\|+\eta_{c-s} G\Big) \right]\\
=\sum_{t=c-s}^{c-1}\EE\left[\|\sum_{p=1}^{P}U_p(\mathbf{W}_p^c)\| \|\sum_{q=1}^{P}U_q(\mathbf{W}_q^t, I_q^t)\|\right] \\
+ \eta_{c-s} G \cdot \EE\left[\|\sum_{p=1}^{P}U_p(\mathbf{W}_p^c)\|\right] \\
\le \sum_{t=c-s}^{c-1}\EE\left[\|\sum_{p=1}^{P}U_p(\mathbf{W}_p^c)\|_2^2+ \|\sum_{q=1}^{P}U_q(\mathbf{W}_q^t, I_q^t)\|_2^2\right] \\
+ \EE\|\sum_{p=1}^{P}U_p(\mathbf{W}_p^c)\|_2^2 + \eta_{c-s}^2 G^2\\
\le (s+1)\EE\|\sum_{p=1}^{P}U_p(\mathbf{W}_p^c)\|_2^2 + \eta_{c-s}^2 G^2\\
+ \sum_{t=c-s}^{c-1}\left[\EE\|\sum_{q=1}^{P}U_q(\mathbf{W}_q^t)\|_2^2 +\eta_t^2 \sigma^2 P\right] 
 \end{array}
\end{equation}
Now plug this into the previous result in \eqref{eq:tmp}:
\begin{equation}
\begin{array}{l}
\EE F(\mathbf{W}^{c+1}) -\EE F(\mathbf{W}^c)\\
 \le (\frac{L_F}{2} - \eta_c^{-1})\EE \|\sum_{p=1}^{P}U_p(\mathbf{W}_p^c)\|_2^2 
\\+(s+1)L\EE \|\sum_{p=1}^{P}U_p(\mathbf{W}_p^c)\|_2^2 + \eta_{c-s}^2 G^2 L\\
+ \sum_{t=c-s}^{c-1}\left[L\EE \|\sum_{p=1}^{P}U_p(\mathbf{W}_p^c)\|_2^2 + \eta_t^2 L\sigma^2 P\right]  + \frac{L_F \eta_c^2 \sigma^2 P}{2}\\
=(\frac{L_F}{2}+(s+1)L - \eta_c^{-1})\EE \|\sum_{p=1}^{P}U_p(\mathbf{W}_p^c)\|_2^2+\eta_{c-s}^2 G^2 L\\
+ \sum_{t=c-s}^{c-1}\left[L\EE \|\sum_{p=1}^{P}U_p(\mathbf{W}_p^c)\|_2^2 + \eta_t^2 L\sigma^2 P\right] +\frac{L_F \eta_c^2 \sigma^2 P}{2}
 \end{array}
\end{equation}
Sum both sides over $c = 0,...,C$:
\begin{equation}
\begin{array}{l}
\EE F(\mathbf{W}^{C+1}) -\EE F(\mathbf{W}^0) \\
\le\sum_{c=0}^{C}\left[(\frac{L_F}{2}+(2s+1)L - \eta_c^{-1})\EE \|\sum_{p=1}^{P}U_p(\mathbf{W}_p^c)\|_2^2\right]\\
+(L\sigma^2 P s+\frac{L_F \sigma^2 P}{2}) \sum_{c=0}^{C}\eta_c^2 + G^2 L\sum_{c=0}^{C}\eta_{c-s}^2
 \end{array}
\end{equation}
After rearranging terms we finally obtain
\begin{equation}
\begin{array}{l}
\sum_{c=0}^{C}\left[\eta_c^2 (\eta_c^{-1} - \frac{L_F}{2}-2(s+1)L)\EE\|\sum_{p=1}^{P} \right.
 \left. \nabla F_p(\mathbf{W}_p^c)\|_2^2\right]\\
 \le  \EE F(\mathbf{W}^0) - \EE F(\mathbf{W}^{C+1}) + (L\sigma^2 P s+\frac{L_F \sigma^2 P}{2}) \sum_{c=0}^{C}\eta_c^2 \\
 + G^2 L\sum_{c=0}^{C}\eta_{c-s}^2
\end{array}
\end{equation}

Now set $\eta_c^{-1} = \frac{L_F}{2}+2sL +\sqrt{c}$. Then, the above inequality becomes (ignoring some universal constants):
\begin{equation}
\begin{array}{l}
\label{eq:rate}
\sum_{c=0}^{C}\left[\frac{1}{\sqrt{c}}\mathbb{E}\|\sum_{p=1}^{P}\nabla F_p(\mathbf{W}_p^c)\|_2^2\right] \\
\le O\left(\Big((L+L_F)\sigma^2 P s\Big)\sum_{c=0}^{C}\frac{1}{c}\right).
\end{array}
\end{equation}
Since $\sum_{c=0}^{C}\frac{1}{c} = o(\sum_{c=0}^{C}\frac{1}{\sqrt{c}})$, we must have 
\begin{equation}
\begin{array}{l}
\label{eq:claim1}
\liminf\limits_{c\to\infty}\mathbb{E}\|\sum_{p=1}^{P}\nabla F_p(\mathbf{W}_p^c)\| = 0
\end{array}
\end{equation}
proving the first claim.

On the other hand, the bound of $\|\mathbf{W}^c - \mathbf{W}_p^c\|$ in \eqref{eq:tmp2} gives
\begin{equation}
\begin{array}{l}
\|\mathbf{W}^c - \mathbf{W}_p^c\| \le \sum_{t=c-s}^{c-1} \eta_t \|\sum_{q=1}^{P} |S_q| \sum_{j\in I_q^t}\nabla f_{j}(\mathbf{W}_q^t)\| + \eta_{c-s} G
\end{array}
\end{equation}
By assumption the sequences $\{\mb{W}_p^c\}_{p,c}$ and $\{\mb{W}^c\}_c$ are bounded and the gradient of $f_j$ is continuous, thus $\nabla f_{j}(\mathbf{W}_q^t)$ is bounded. Now take $c\rightarrow \infty$ in the above inequality and notice that $\underset{c\rightarrow \infty}{\mathrm{lim}}\eta_c = 0$, we have $\underset{c\rightarrow \infty}{\mathrm{lim}}\|\mathbf{W}^c - \mathbf{W}_p^c\| = 0$ almost surely, proving the second claim.

Lastly, the Lipschitz continuity of $\nabla F_p$ further implies
\begin{equation}
\begin{array}{l}
0 = \liminf\limits_{c\to\infty}\mathbb{E}\|\sum_{p=1}^{P}\nabla F_p(\mathbf{W}_p^c)\| \geq \liminf\limits_{c\to\infty}\mathbb{E}\|\sum_{p=1}^{P}\nabla F_p(\mathbf{W}^c)\| \\
= \liminf\limits_{c\to\infty}\mathbb{E}\|\nabla F(\mathbf{W}^c)\|=0
\end{array}
\end{equation}
Thus there exists a common limit point of $\mathbf{W}^c, \mathbf{W}_p^c$ that is a stationary point almost surely. From \eqref{eq:rate} and use the estimate $\sum_{c=1}^C \tfrac{1}{c} \approx \log C$, we have 
\begin{equation}
\begin{array}{l}
\min_{c=1,\ldots, C} \EE \left[\|\sum_{p=1}^{P}\nabla F_p(\mathbf{W}_p^c)\|_2^2\right] \leq O\left(\frac{(L+L_F)\sigma^2 P s\log C}{\sqrt{C}}\right).
\end{array}
\end{equation}
The proof is now complete.
\end{proof}

\end{appendix}

%\section{Additional Experiment Results}

%Figure \ref{fig:exp_runtime_ssp} shows the communication volume (measured by TB) for MLR, DML, SC, L2-MLR under a fixed number of iterations and the BSP consistency model. The communication volume in SFB is much smaller than Spark and

%\begin{appendix}
%You can use an appendix for optional proofs or details of your evaluation which are not absolutely necessary to the core understanding of your paper. 
%
%\section{Final Thoughts on Good Layout}
%Please use readable font sizes in the figures and graphs. Avoid tempering with the correct border values, and the spacing (and format) of both text and captions of the PVLDB format (e.g. captions are bold).
%
%At the end, please check for an overall pleasant layout, e.g. by ensuring a readable and logical positioning of any floating figures and tables. Please also check for any line overflows, which are only allowed in extraordinary circumstances (such as wide formulas or URLs where a line wrap would be counterintuitive).
%
%Use the \texttt{balance} package together with a \texttt{\char'134 balance} command at the end of your document to ensure that the last page has balanced (i.e. same length) columns.
%
%\end{appendix}

\end{document}